\documentclass[journal]{IEEEtran}

\usepackage{amsmath,amssymb,amsthm,amsfonts,mathtools} 
\usepackage{dsfont,eucal,bbm,bm,nicefrac} 
\usepackage{graphicx,float,subcaption,booktabs} 
	\graphicspath{{./figures-arxiv/}}
\usepackage{algorithm,algorithmic} 
\usepackage{hyperref} 
\usepackage{tikz,xcolor} 
\usepackage{cite}
\usepackage{textcomp}

\newtheorem{theorem}{Theorem}
\newtheorem{corollary}{Corollary}[theorem]

\newtheorem{remark}{Remark}

\newcommand{\pbra}[1]{\ensuremath{\left( #1\right)}}
\newcommand{\sbra}[1]{\ensuremath{\left[ #1\right]}}
\newcommand{\cbra}[1]{\ensuremath{\left\{ #1\right\}}}

\newcommand{\E}[1]{\ensuremath{\mathbb{E}\left[ #1\right]}}

\begin{document}

\title{Towards the One Learning Algorithm Hypothesis: A System-theoretic Approach}

\author{Christos N. Mavridis, \IEEEmembership{Member, IEEE}, and 
John S. Baras, \IEEEmembership{Life Fellow, IEEE}
\thanks{
The authors are with the 
Department of Electrical and Computer Engineering and 
the Institute for Systems Research, 
University of Maryland, College Park, USA.
{\tt\small emails:\{mavridis, baras\}@umd.edu}.}%
\thanks{Research partially supported by the 
Defense Advanced Research Projects
Agency (DARPA) under Agreement No. HR00111990027, 
by ONR grant N00014-17-1-2622, 
and by a grant from Northrop Grumman Corporation.%
}%
}

\maketitle
 \thispagestyle{empty}
\pagestyle{empty}

\begin{abstract}
The existence of a universal learning architecture
in human cognition is 
a widely spread conjecture supported by 
experimental findings from neuroscience. 
While no low-level implementation can be specified yet, 
an abstract outline of human perception and learning is believed to 
entail three basic properties:
(a) hierarchical attention and processing, 
(b) memory-based knowledge representation, and
(c) progressive learning and knowledge compaction.
%
%
We approach the design of such a learning architecture 
from a system-theoretic viewpoint, developing a closed-loop system with three main components:
(i) a multi-resolution analysis pre-processor, 
(ii) a group-invariant feature extractor, and
(iii) a progressive knowledge-based learning module. 
Multi-resolution feedback loops are used for learning, i.e., 
for adapting the system parameters to online observations. 
To design (i) and (ii), 
we build upon the established theory of wavelet-based multi-resolution 
analysis and the properties of group convolution operators. 
%
Regarding (iii), we introduce a novel learning algorithm 
that constructs progressively growing knowledge representations in multiple resolutions.
The proposed algorithm is an extension of the 
Online Deterministic Annealing (ODA) algorithm which 
can be viewed as a competitive-learning neural network
where the neurons represent codevectors in the data space
and the training rule is based on annealing optimization, 
solved using gradient-free stochastic approximation.
ODA has inherent
robustness and regularization properties 
and provides a means 
to progressively increase the complexity of the learning model 
i.e. the number of the neurons, as needed, 
through an intuitive bifurcation phenomenon.
%
The proposed multi-resolution approach is
hierarchical, progressive, knowledge-based, and interpretable,
allowing the localization of the error-prone regions of the data space.
%
We illustrate the properties of the proposed architecture 
in the context of the state-of-the-art learning algorithms and deep learning methods. 
\end{abstract}


\section{Introduction}
\label{Sec:Introduction}

\IEEEPARstart{L}{earning} 
from data samples has become an important component 
in the advancement of numerous research fields including artificial intelligence,
computational physics, biology and control systems. 
While virtually all learning problems can be formulated as 
constrained stochastic optimization problems, 
the optimization methods can be intractable 
due to the nature of the problems and the constraints originating 
from the model assumptions.   
%
For this reason, some of the most impactful breakthroughs in machine learning research
have been a result of selecting appropriately structured learning models
that can be trained efficiently using existing optimization methods \cite{bruna2013learning}.

In search for appropriate learning models and inspired
by established 
experimental findings from neuroscience 
\cite{roe1992visual,sharma2000induction}, 
a widely spread conjecture has been formed regarding 
the existence of a universal learning model
in human cognition.
While no observations can yet lead to the specifics of the implementation
of such an architecture, at an abstract level it is believed 
that human perception, cognition, and learning, are based 
on three main properties:
(a) hierarchical attention and information processing 
\cite{taylor2015global,riesenhuber1999hierarchical,sharma2000induction,
ito2004representation,read2002functional,chi2005multiresolution}, 
(b) memory-based knowledge representation 
\cite{carpenter2010adaptive}, and
(c) progressive learning and knowledge compaction 
\cite{poggio2016visual,lecun2015deep}.

Deep learning methods, 
currently dominating the field of machine learning
due to their performance in numerous applications, 
have made progress towards this direction,
especially regarding the hierarchical properties (a). 
They attempt to learn feature representations from data, 
using biologically-inspired models in artificial neural networks 
that mimic the low-level computational architecture of the human brain units
\cite{lecun2015deep,hinton2006fast,	krizhevsky2012imagenet,lee2009convolutional}.
They provide task-specific end-to-end learning algorithms 
implemented by a non-von Neumann computational architecture 
that does not discriminate
between computational and memory units.
As a result, however, they
do not provide
a framework for memory representation and knowledge compaction 
as in (b) and (c). 
\begin{figure}[t]
\centering
\includegraphics[trim=0 0 0 0,clip,width=0.45\textwidth]{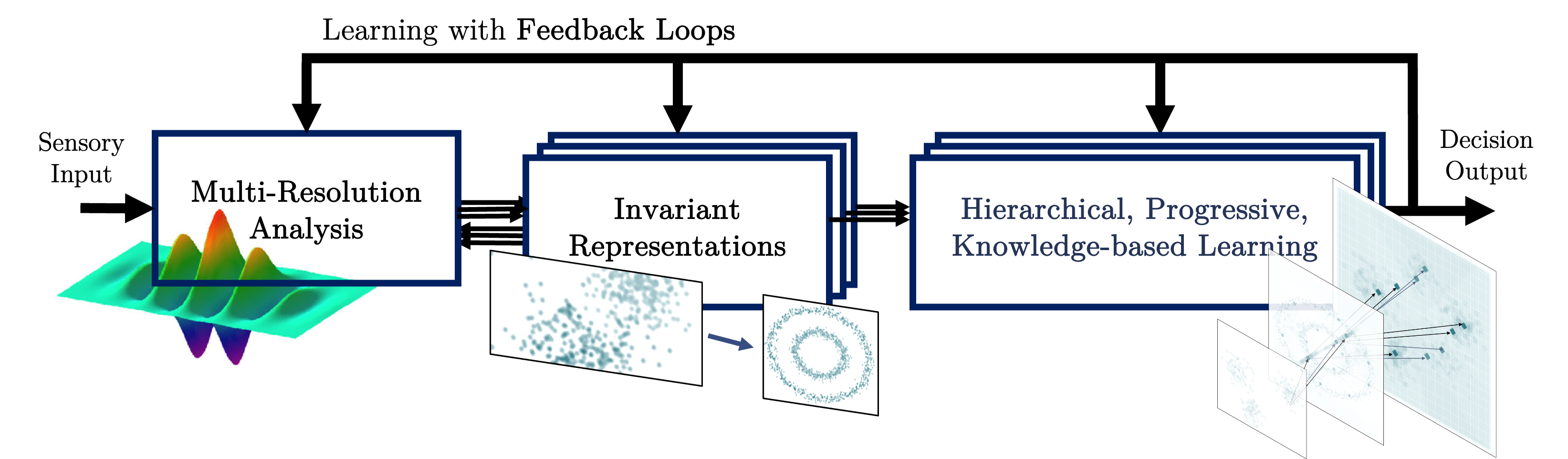}
\caption{Abstract system-theoretic block diagram of a ``universal learning architecture'' in human cognition.}
\label{fig:abstract_arch}
\end{figure}

On the other hand, it is understood that artificial intelligence systems
need not mimic the low-level architecture of the brain cells, 
but rather get inspirations from 
abstract properties of human intelligence \cite{jordan2019artificial}.
This becomes especially important when considering that adopting 
black-box deep neural network 
architectures results in
using overly complex models of a great many parameters 
in the expense of time, energy, data, memory and 
computational resources \cite{thompson2020computational,strubell2019energy},
Even in the applications when complexity is not an issue, 
the lack of
interpretability and mathematical understanding, 
and the vulnerability to small perturbations and adversarial attacks
\cite{szegedy2013intriguing,carlini2017towards,madry2017towards}, 
has led to an emerging hesitation in 
the use of deep learning models outside common benchmark datasets 
\cite{sehwag2019analyzing,northcutt2021pervasive}, 
and, especially, in security critical applications.
These models are hard to analyze with current mathematical tools, 
hard to train with current optimization methods, and their design relies
solely in experimental heuristics.
As a result, researchers have been trying to answer 
questions about 
their approximation power, 
the dynamics and convergence of optimization, 
the properties of the intrinsically 
generated representations, 
and the phenomena of 
overfitting, regularization, and overparameterization,
in an attempt to understand the conditions under which deep learning methods work as expected
\cite{poggio2020theoretical,mallat2016understanding,
vidal2017mathematics,poggio2016visual,liu2020toward}.

To this end, we approach the design of a
``universal learning architecture'' \cite{von2000visual,thaler2011neural}
from a system-theoretic viewpoint,
based on the aforementioned architectural abstractions of the 
processing system of the auditory and visual cortex of the human
\cite{taylor2015global,riesenhuber1999hierarchical,sharma2000induction,
ito2004representation,read2002functional,chi2005multiresolution}.
The architecture consists of a closed-loop system with 
three main components:
\begin{itemize}
\item[$(i)$] a hierarchical signal decomposition unit based on 
	multi-resolution wavelet analysis \cite{mallat2016understanding},
\item[$(ii)$] a feature extraction module that, 
	in combination with $(i)$, estimates group-invariant representations 
	in multiple resolutions, and
\item[$(iii)$] a hierarchical, progressive, robust, 
	and interpretable learning model
	for supervised and unsupervised learning that
	is able to represent and progressively
	compact accumulated knowledge.
\end{itemize}
Multi-resolution feedback loops are used for 
adaptive estimation of the optimal parameters for 
$(i)$, $(ii)$, and $(iii)$.
An abstract block diagram is depicted in 
Fig. \ref{fig:abstract_arch}. 
%

Regarding the subsystems $(i)$ and $(ii)$, we follow the established 
theory of wavelet decomposition to build a multi-resolution 
representation of the input. 
Wavelet decomposition is based on a cascade of
convolution and downsampling operators.
Convolutions create translation equivariant representations 
which are stable to deformations, and can be generalized to carry these
properties to any known compact Lie group \cite{mallat2012group}.
Downsampling, on the other hand, constitutes an averaging operation that
creates local invariance, and 
can be viewed as a pooling operation.
Finally, creating a hierarchical cascade 
of convolution operations followed by non-linear mappings 
(subsystem $(ii)$) and averaging 
functions, has been shown to construct multi-scale group-invariant 
representations \cite{mallat2016understanding}.

Regarding $(iii)$, 
we build upon 
the Online Deterministic Annealing (ODA) algorithm \cite{mavridis2021online}, 
a novel learning algorithm 
that constructs progressively growing knowledge representations. 
ODA can be formulated as a series of
soft-clustering optimization problems
%
%
%
\begin{align*}
\min_{M} ~ F_T(M) := D(M) - T H(M), 
\end{align*}
%
parameterized by a temperature coefficient $T$.
Here $D$ is the average distortion $D(M) := \E{\sum_i p(\mu_i|X) d(X,\mu_i)}$
measured by $d$, the proximity measure that defines the similarity between the 
random input $X$ and a codevector $\mu_i$ with appropriately defined association probabilities $p(\mu_i|X)$.
$H$ 
represents the Shannon entropy that 
characterizes the ``purity'' of the clusters induced by the codevectors, and
$T$ acts as a
Lagrange multiplier controlling the trade-off between minimizing 
the distortion $D$ and maximizing the entropy $H$.
The codevectors can be viewed as a set of neurons,
the weights of which live in the data space itself. 
The representation of the input 
in terms of memorized exemplars is an intuitive approach 
which parallels similar concepts from cognitive psychology
and neuroscience and induces interpretability. 
Moreover, it resembles vector quantization algorithms,  
which have shown impressive robustness 
against adversarial attacks,
suggesting suitability in security critical 
applications \cite{saralajew2019robustness}.
%
%
By successively solving the optimization problems $\min_{M} ~ F_T(M)$ 
for decreasing values of $T$, the model undergoes a series of 
phase transitions when the cardinality of the set of codevectors $M$ increases,
according to an intuitive bifurcation phenomenon.
This resembles an annealing process \cite{rose1998deterministic}
that introduces inherent robustness and regularization properties.
The optimization problems are solved using gradient-free stochastic approximation
\cite{borkar2009stochastic}, provided that the proximity measure 
$d$ belongs to the family of Bregman divergences:
information-theoretic dissimilarity measures (including the widely used 
Kullback-Leibler divergence)
that have been shown to play an important role in learning applications
\cite{banerjee2005clustering,villmann_onlineDLVQmath_2010}.
%
%
%

Finally, 
to construct a hierarchical multi-resolution learning approach
for $(iii)$, 
we develop a tree-structured ODA learning algorithm.
Traditional tree-structured vector quantization algorithms partition the data space 
by building a labeled binary tree using ad-hoc splitting criteria. 
This yields a faster, yet suboptimal solution to the original problem.
In contrast, we train each layer of the tree with the ODA algorithm that progressively 
adjusts the number of codevectors-children for each node according to the annealing optimization process.
Moreover, to make use of multiple-resolutions, 
we extend this approach having each layer trained 
using progressively more detailed representations
according to the multi-resolution analysis of $(i)$ and $(ii)$.
As will be shown, the combination of the convolution-based multi-resolution features with the 
proposed tree-structured multi-layer neural network resembles the structure of 
Deep Convolutional Networks \cite{lecun2015deep} and
Scattering Convolutional Networks \cite{bruna2013invariant}, 
and provides a general system-theoretic framework for progressive, hierarchical, 
interpretable, and knowledge-based learning.

\section{Hierarchical and Invariant Representations}
\label{Sec:Wavelets}

To understand 
how deep learning models work, 
it is important to mathematically characterize the properties of the internally 
learned representations \cite{zhao2016energy,lecun2016thenext,
anselmi2016unsupervised,krizhevsky2012imagenet}
and its inductive bias, i.e., 
the class of tasks for which they are predesigned to perform well.
As an example, in computer vision tasks, 
geometric stability has been shown to play an important role and can 
be measured 
in terms of invariance to translations and stability to local deformations, 
as will be explained later. 
Deep convolutional neural networks have been shown to 
provide this geometrical stability
\cite{vidal2017mathematics} by following a compositional 
approach with a hierarchically structured model
\cite{lecun2015deep,geman2007compositionality}.
The structure is mainly composed by successive operations of
a linear operator (or linear filter, e.g., convolutional layers), 
a nonlinear mapping (often a rectifying function, e.g., ReLu),
and a down-sampling step (e.g., max-pooling), which produces the 
input for the next stage of the architecture
\cite{bruna2013invariant,mallat2016understanding,anselmi2015deep}.
This approach  
seems to reflect neuro-scientific findings \cite{taylor2015global,geman2007compositionality,yuille2018deep},
and is also 
reminiscent of the multi-resolution representations produced by 
the extensively studied wavelet transform.
The wavelet transform \cite{mallat1989theory}, as well 
as extensions including the wavelet packets \cite{mallat1999wavelet},
offer a solid mathematical tool to understand the properties 
of the multi-resolution representation of a signal.
As a result, wavelets have been already used to study the mathematics of 
deep learning, with the scattering transform 
\cite{bruna2013invariant,mallat2016understanding}
having been shown to preserve 
the aforementioned geometric stability properties,
hinting that convolution
filters need not be learned from data, but can be defined a priori 
with respect
to the properties of the task at hand.

In this section, we briefly review the mathematics of the wavelet 
and wavelet packet transform and show how they can be used to 
build multi-resolution locally invariant representations 
with respect to compact Lie groups.

%

\subsection{Wavelet-based Multi-Resolution Analysis}

Consider a measurable signal $f(t)$ with finite energy, i.e., 
$f(t)\in L^2(\mathbb{R}^n)$. 
We will focus in one-dimensional 
signals ($n=1$) for simplicity of notation, 
although the analysis can be generalized for any $n>0$.
The continuous wavelet transform of $f$ corresponds to the 
set of coeeficients of the atoms of a redundant dictionary
of time-scale images (in $(u,s)$):
\begin{equation}
\cbra{\psi_{u,s}(t) = \frac{1}{\sqrt{s}}\psi(\frac{t-u}{s})}_{u\in\mathbb{R},s>0}
\end{equation}
where $\psi\in L^2(\mathbb{R})$ has zero average, 
i.e., $\int_{-\infty}^\infty \psi(t) dt = 0$,
and is called the mother wavelet.
In many useful cases, such as in directional wavelets for 2D images,
the mother wavelets are defined such that the redundant dictionaries 
constitute a 
complete and stable signal representation, called a frame 
\cite{mallat1999wavelet}.
For simplicity of notation, we will focus on 
orthonormal (Riesz) bases which remove all redundancy and 
define stable representations. 
A wavelet orthonormal basis is obtained by scaling a wavelet
$\psi$ with a dyadic scale $s=2^j$ and translating it by 
$u=2^jn$:
\begin{equation}
\cbra{\psi_{j,n}(t) = \frac{1}{\sqrt{2^j}}\psi(\frac{t-2^jn}{2^j})}_{(j,n)\in\mathbb{Z}^2}
\label{eq:psi}
\end{equation}
The simplest orthonormal basis can be constructed using the 
piecewise constant Haar wavelets, 
but regular wavelets of compact support can form 
an orthonormal basis as well \cite{mallat1999wavelet}.
The signal $f$ can be expressed in the wavelet bases as:
\begin{equation}
f = \sum_{j=-\infty}^{\infty} \sum_{n=-\infty}^{\infty} 
	c_{j,n} \psi_{j,n}
\end{equation}
where $c_{j,n} = \langle f,\psi_{j,n}\rangle$. 
Moreover, $f$ can be expressed as a direct sum:
\begin{align}
f &= \sum_{j>J} \sum_{n=-\infty}^{\infty} 
	c_{j,n} \psi_{j,n}
	+ \sum_{j\leq J} \sum_{n=-\infty}^{\infty} 
	c_{j,n} \psi_{j,n} \\
&= 	\sum_{n=-\infty}^{\infty} 
	a_{J,n} \phi_{J,n}
	+ \sum_{j\leq J}^{\infty} \sum_{n=-\infty}^{\infty} 
	c_{j,n} \psi_{j,n}
\end{align}
where $\phi_{j,n}$ is called the scaling function and is defined similar 
to (\ref{eq:psi}) according to a mother scaling function $\phi$.
The set $\cbra{\phi_{J,n}}$ is an orthonormal basis of
the subspace $V_J\subset L^2(\mathbb{R})$ which corresponds to the 
approximation of $L^2(\mathbb{R})$ at resolution $J$, and 
has the following properties:
\begin{itemize}
\item $V_j\subset V_{j-1}$, $\forall j\in \mathbb{Z}$,
\item $\cup_{j=-\infty}^{\infty} V_j$ is dense in $L^2(\mathbb{R})$
	with $\cap_{j=-\infty}^{\infty} V_j=\cbra{0}$,
\item $f(t) \in V_j \Leftrightarrow f(2t)\in V_{j-1}$, $\forall j\in \mathbb{Z}$, and
\item $f(t) \in V_j \implies f(t-2^jk)\in V_{j}$, $\forall k\in \mathbb{Z}$.
\end{itemize}
It is easy to see that the sum $\sum_{n=-\infty}^{\infty} 
	c_{j,n} \psi_{j,n}$
is an orthonormal basis for the subspace $W_j$ for which 
$V_j \oplus W_j = V_{j-1}$ and
$V_J = \oplus_{j>J} W_j$.
In other words, the additional information lost in the approximation 
$V_J$ at resolution $2^J$ is contained in the subspace $W_J$. 

In practice, a physical measurement device can only 
measure a signal at a finite resolution, denoted by $j=0$.
Then the coefficients $\cbra{c_{j,n}}_{j\in\mathbb{N}}$ and 
$\cbra{a_{j,n}}_{j\in\mathbb{N}}$ define the multi-resolution 
representation of the wavelet transform. 
Moreover, the coefficients
\begin{equation}
\begin{aligned}
c_{j,n} = \langle f,\psi_{j,n}\rangle 
	= \int_{-\infty}^\infty f(t) \psi_{j,n}(t) dt \\
a_{j,n} = \langle f,\phi_{j,n}\rangle 
	= \int_{-\infty}^\infty f(t) \psi_{j,n}(t) dt 
\end{aligned}
\end{equation}
can also be written as:
\begin{equation} 
\begin{aligned}
\langle f,\psi_{j,n}\rangle &= 
(f \ast \psi_{2^j})(t)\vert_{t=2^jn} := W_jf(t)\vert_{t=2^jn} \\
\langle f,\phi_{j,n}\rangle &= 
(f \ast \phi_{2^j})(t)\vert_{t=2^jn} 
\end{aligned}
\label{eq:psi_conv0}
\end{equation}
where $\psi_{2^j}(x) = 2^{-j/2} \psi(- 2^{-j} x)$, 
$\phi_{2^j}(x) = 2^{-j/2} \phi(- 2^{-j} x)$, and
$(f\ast g)(x) = \int_{-\infty}^\infty f(u) g(x-u) du$ is the convolution 
operation.
Taking into account that $\psi_{2^j}, \phi_{2^j} \in V_{j-1}$, the 
coefficients can be computed by:
\begin{equation} 
\begin{aligned}
\langle f,\psi_{j,n}\rangle &= (g \ast \langle f,\phi_{2^{j-1}}\rangle)(t)\vert_{t=2n} \\
\langle f,\phi_{j,n}\rangle &= (h \ast \langle f,\phi_{2^{j-1}}\rangle)(t)\vert_{t=2n} \\ 
\end{aligned}
\label{eq:psi_conv}
\end{equation}
Equation (\ref{eq:psi_conv}) reveals that the wavelet coefficients 
$\langle f,\psi_{j,n}\rangle$, $\langle f,\psi_{j,n}\rangle$ 
can be computed by hierarchical filtering with 
appropriately defined filters associated with the mother wavelet $\psi$, 
followed by a 
uniform downsampling by a factor of $2$.
In particular, $h$ is a low-pass filter and 
$g$ is a high-pass filter.
They depend on $\phi$ and $\psi$, 
and define a pair of quadrature mirror filters, 
an observation that has led to the fast
filter-bank implementation of the wavelet transform and its association 
with multi-resolution analysis \cite{mallat1989theory}.
%

\subsection{Wavelets, Translation Invariance, and Stability to Deformations}

As shown above, the computation of the multi-resolution wavelet representation of a signal
consists of successive operations of
a linear convolution operator,
followed by a downsampling step.
This structure is 
similar to deep convolutional neural networks, 
and has been very successful in multiple applications,
especially in sound and image recognition,
%
%
where translation invariance and stability to small 
deformations are important properties of the representation \cite{mallat2016understanding}.
The translation group is a Lie group of operators 
$\cbra{T_c}_{c\in\mathbb{R}}$
such that 
$T_c f(t) = f(t-c)$, while 
the group of deformations $D_\tau$ can be written as 
$D_\tau f(t) = f(t-\tau(t))$, where the local function 
$\tau$ depends on $t$ and thus deforms the image.
A representation $\Phi:L^2(\mathbb{R}\rightarrow H$, 
where $H$ is a Hilbert space, is said to be 
invariant with respect to $T_c$ if 
$\Phi(T_c f) = \Phi(f)$.
The wavelet transform (eq. (\ref{eq:psi_conv0}))
is based on the convolution operation,
and is therefore translation covariant (or equivariant), i.e., 
it commutes with the translation operator and 
satisfies the relation $W_j(T_c f) = T_c W_j(f)$.
At the same time, the wavelet transform, is stable to small deformations, 
which is a very important property that 
can be expressed as a Lipschitz continuity condition:
\begin{equation}
\| W_j(D_\tau f) - W_j(f) \|_2 \leq C \|f\|_2 \sup_t |\nabla\tau(t)|
\end{equation}
where $\|f\|_2 = \int |f(t)|^2 dt$ is the $L^2(\mathbb{R})$ norm, and 
$|\nabla \tau(t)|$ is a measure of the deformation amplitude
(it is assumed that $|\nabla \tau(t)|<1$, which makes the small deformation an invertible transformation \cite{bruna2013invariant}).

\begin{remark}
We note that, in the 
control theory and signal processing communities,
convolutions are associated with systems described by the term 
'linear time-invariant'. 
To avoid confusion, with the terminology used here, 
these systems are considered linear covariant (or equivariant) operators  
with respect to translation in time.
\end{remark}

To induce invariance with respect to translation, 
it is sufficient to integrate (average) a 
translation-covariant representation 
over the translation group.
That is to say that 
the integral $\int W_jf(c) dc$ is translation invariant.
However, it turns out to be 
the trivial invariant $\int_{-\infty}^\infty W_jf(t-c) dc = 0$, since,
by definition, $\int_{-\infty}^\infty \psi(t) dt = 0$. 
Therefore, it is necessary to insert a non-linear operation between the 
wavelet transform and the integration \cite{bruna2013invariant}. 
%
It has been shown that a point-wise and 
stable non-linear operator $\rho$ 
(for stability: $\|\rho f\|\leq \|f\|$ and $\|\rho f- \rho g\|\leq \|f-g\|$), 
that also commutes with the translation (and the deformation) group,
is the $L^1(\mathbb{R})$ norm: $\|f\|_1 = \int |f(t)| dt$, 
which acts as a rectifying function, similar to the ReLu function in 
deep convolutional neural networks.
The choice $\rho = \|\cdot\|_1$ also preserves the deformation stability
properties of the wavelet transform.

The cascade of the wavelet transform, the non-linear operation, 
and the averaging integral, results in the representation 
$\int \rho W_j(f)(t) dt$, in every resolution $j$.
However, global invariance is not always desired.
In contrast, it is often better to 
compute locally invariant representations for translations
up to a scale $2^J$ for some index $J$.
This can be obtained by substituting the integral 
by a convolution with a low-pass filter localized in a spatial 
window scaled at $2^J$. 
The result is a representation that belongs to $V_J$, and, 
in particular the projection of $\rho W_j(f)$ to $V_J$, for $J\geq j$, 
which can be computed by the pyramid of filters of the wavelet transform, 
as explained above.
Moreover, during the computation of the wavelet transform, the information
lost in averaging (projecting to $V_J$) is kept in the coefficients
$\cbra{c_{j,n}}_{j<J}$ of the wavelet transform.
This information can be used to build new invariant features by 
applying the wavelet transform again, creating a deep scattering network 
introduced by Bruna in \cite{bruna2013invariant}.
%
%
The implementation of the
scattering transform is based on a complex-valued convolutional neural network 
whose filters are fixed wavelets and $\rho$ is a complex modulus operator
as described above \cite{andreux2019kymatio}.
We note that the features of the scattering transform do not belong to the 
subspaces $\cbra{V_j, j\in \mathbb{N}}$ that define the classical wavelet hierarchy.
However, they define a multi-level 
hierarchy of successively more detailed information that can be used in a similar way.

%
%
%

\subsection{Building Invariance to compact Lie Groups}
\label{sSec:liegroups}

The invariance properties discussed above can be generalized to the action 
of arbitrary compact Lie groups, such as a rotation and translation group \cite{mallat2012group}.
Let $G$ be a compact Lie group and $L^2(G)$ be the space of measurable functions $f(r)$ such that 
$\|f\|^2 = \int_G |f(r)|^2 dr <\infty$, where $dr$ 
is the Haar measure of $G$. 
The left action of $g\in G$ on $f\in L^2(G)$ 
is defined by $L_g f(r) = f(g^{-1}r)$.
As a special case, the action of the translation group 
$T_c f(t) = f(t-c)$ translates the function $f$ to the right by $c$,
with $g^{-1} = -c$ translating the argument of $f$ to the left by $c$.
Similar to the usual convolution 
$(f\ast h)(x) = \int_{-\infty}^\infty f(u) h(x-u) du$
that defines a linear translation covariant operator,  
convolutions on a group appear naturally as 
linear operators covariant to the action of a group:
\begin{equation}
(f\ast h)(x) = \int_G f(g) h(g^{-1}x) dr
\end{equation}
where $dr$ is the Haar measure of $G$. 
As a result, an invariant representation 
relatively to the action of a compact Lie group, 
can be computed by averaging over covariant representations 
created by group convolution with appropriately defined wavelets, 
similar to the methodology explained above.

\subsection{Multi-resolution Feedback Loops}

The methodology explained in Section \ref{sSec:liegroups} can be used to 
build invariant features with respect to a known compact Lie Group.
In the case of image and sound recognition, it has been shown that local invariance  
with respect to the groups of spatial translation and rotation is sufficient \cite{mallat2016understanding}. 
In other words, the groups with respect to which we design the wavelet-based features described above,
are known a priori.
In general, such information may not be known, and may need to be estimated from the data.
This is the purpose of the multi-resolution feedback loops presented in Fig. \ref{fig:abstract_arch}, 
and specifically, the first set of feedback loops shown in Fig. \ref{fig:learning-arch-blue}(a).
The wavelet transform used can be generalized to a redundant hierarchical dictionary, as well,
the basis of which can be estimated again from data via the same feedback loops. 
%
In the case of a wavelet transform this reduces to estimating the mother wavelet \cite{zhuang1994optimal}.
To estimate different non-linear functions $\rho$, the second set of feedback loops 
can also be used.
Therefore, the multi-resolution feedback loops make the proposed approach 
a general framework for task-agnostic learning.
Estimating this non-linear mapping or the mother wavelet from data is beyond the scope of this paper.

\section{Online Deterministic Annealing for Unsupervised and Supervised Learning}
\label{Sec:ODA}

To build a progressive learning algorithm with the properties mentioned above, 
we will start our analysis with
unsupervised learning,
which can provide valuable insights into the nature of the
dataset at hand, and it plays an important role in the
context of visualization.
Central to unsupervised learning
is the representation of 
data in a vector space by typical representatives which is implemented with 
vector quantization variants \cite{mavridis2020convergence}. 
Given a random variable $X: \Omega \rightarrow S$ defined in the 
probability space $\pbra{\Omega, \mathcal{F}, \mathbb{P}}$, 
a quantizer $Q:S \rightarrow S$ is defined such that 
$Q(X) = \sum_{h=1}^K \mu_h \mathds{1}_{\sbra{X \in S_h}}$,
where $V := \cbra{S_h}_{h=1}^K$ forms a partition of $S$
and $M := \cbra{\mu_h}_{h = 1}^K$ represents a set of codevectors such that 
$\mu_h \in ri(S_h)$, $h\in \cbra{1, \ldots, K}$.
Given a dissimilarity measure 
$d:S \times ri(S) \rightarrow \left[0,\infty\right)$
one seeks the optimal $M, V$ in terms of minimum average distortion: 
\begin{equation}
\min_{M,V} ~ D(Q) := \E{d\pbra{X,Q(X)}} 
\label{eq:VQ}
\end{equation}	
Vector quantization algorithms assume that $Q$ is a deterministic function 
of $X$ and are proven to converge to locally optimal configurations
even when formulated as online learning algorithms
\cite{mavridis2020convergence}.
However, 
their convergence properties and final configuration 
depend heavily on two design parameters:
$(a)$ the number of clusters (neurons), and 
$(b)$ their initial configuration.
To deal with this phenomenon, the Online Deterministic Annealing approach 
\cite{mavridis2021online}
makes use of a probabilistic framework, where input vectors 
are assigned to clusters in probability, thus
dropping the assumption that $Q$ is a deterministic function of $X$.
For the randomized partition, the expected distortion becomes:
\begin{align*}
D = \E{d_\phi(X,Q)} 
  = \E{\E{d_\phi(X,Q)|X}} 
\end{align*}
The central idea of deterministic annealing 
is to seek the distribution that minimizes $D$ subject to a 
specified level of randomness, measured by the Shannon entropy
\begin{align*}
H(X,M) 
       = H(X) 
       - \E{\E{\log p(Q|X)|X}}
\end{align*}
with $p(\mu|x)$ representing 
the association probability relating the input vector $x$
with the codevector $\mu$. 
This is essentially a realization of the 
Jaynes's maximum entropy principle \cite{jaynes1957information}.
%
The resulting multi-objective optimization is conveniently formulated as the minimization of the Lagrangian
\begin{equation}
F = D-TH
\label{eq:F}
\end{equation}
where $T$ is the temperature parameter that acts as a Lagrange multiplier.
Clearly, (\ref{eq:F}) represents the scalarization method for trade-off 
analysis between two performance metrics. 
For large values of $T$ we maximize the entropy, 
and, as $T$ is lowered, we essentially transition 
from one Pareto point to another in a naturally occurring direction
that resembles an annealing process.
In this regard, the entropy $H$, which is closely related 
to the ``purity'' of the clusters,
acts as a regularization term which is given progressively less weight 
as $T$ decreases.
As is the case in vector quantization, 
one minimizes $F$ via a coordinate block optimization algorithm.
Minimizing $F$ with respect to the association probabilities $p(\mu|x)$ is 
straightforward and yields the Gibbs distribution 
\begin{equation}
p(\mu|x) = \frac{e^{-\frac{d(x,\mu)}{T}}}
			{\sum_\mu e^{-\frac{d(x,\mu)}{T}}}
\label{eq:E}
\end{equation}
while, in order to minimize $F$ with respect to the codevector locations $\mu$ 
we set the gradients to zero 
\begin{equation}
\begin{aligned}
\frac d {d\mu} D = 0 
& \implies 
\frac d {d\mu} \E{\E{d(X,\mu)|X}} = 0 
\end{aligned}
\label{eq:M}
\end{equation}
%

\subsection{Bifurcation Phenomena}

Adding to the physical analogy, it is significant that, 
as the temperature is lowered, the
system undergoes a sequence of ``phase transitions'', which
consists of natural cluster splits where the cardinality of the 
codebook (number of prototypes) increases. This is a bifurcation phenomenon 
that  provides a useful tool for controlling the size of the model
relating it to the scale of the solution.
At very high temperature ($T\rightarrow\infty$) the optimization yields
uniform association probabilities $p(\mu|x)=\nicefrac 1 K$,
and all the codevectors are located at the same point.
This is true regardless of the number of codevectors available.
We refer to the number of different codevectors 
resulting from the optimization process as
\textit{effective codevectors}.
These define the cardinality of the codebook, 
which changes as we lower the temperature.
The bifurcation occurs when the solution above a critical temperature $T_c$ is 
no longer the minimum of the free energy $F$ for $T<T_c$. 
A set of coincident codevectors then splits into separate subsets.
These critical temperatures $T_c$ can be traced when the Hessian of $F$ 
loses its positive definite property, and
are, in some cases, computable (see Theorem 1 in \cite{rose1998deterministic}).
In other words, an algorithmic implementation needs only
as many codevectors as the number of effective codevectors, which
depends only on the temperature parameter, i.e. the Lagrange multiplier
of the multi-objective minimization problem in (\ref{eq:F}).
As shown in \cite{mavridis2021online}, 
we can detect the bifurcation points 
by maintaining and perturbing pairs of codevectors at each 
effective cluster so that they separate only when a critical temperature
is reached. 

\subsection{Bregman Divergences as Dissimilarity Measures}

The proximity measure $d$ need not be a metric, and can be generalized 
to more general dissimilarity measures inspired by 
information theory and statistical analysis.
In particular, 
the family of Bregman divergences, 
which includes the widely used Kullback-Leibler divergence,  
can offer numerous advantages in learning applications
compared to the Euclidean distance alone \cite{banerjee2005clustering}. 
Notably, in the case of deterministic annealing, Bregman divergences play an even
more important role, since
we can show that, if $d$ is a Bregman divergence, 
the solution to the second optimization step 
(\ref{eq:M}) can be analytically computed in a convenient centroid form
\cite{mavridis2021online}:
\begin{equation}
\mu^* = \E{X|\mu} 
\label{eq:mu_star}
\end{equation}
%

\subsection{The online learning rule}

In an offline approach, the approximation of 
the conditional expectation $\E{X|\mu}$ 
is computed by the sample mean of the data points weighted by their association 
probabilities $p(\mu|x)$. 
To define an online training rule 
for the deterministic annealing framework,
we formulate a stochastic approximation algorithm 
to recursively estimate $\E{X|\mu}$ directly.
As a direct consequence of Theorem 4 in \cite{mavridis2021online}, the following corollary 
provides an online learning rule that solves the
optimization problem of the deterministic annealing algorithm.
\begin{corollary}
The online training rule 
\begin{equation}
\begin{cases}
\rho_i(n+1) &= \rho_i(n) + \beta(n) \sbra{ \hat p(\mu_i|x_n) - \rho_i(n)} \\
\sigma_i(n+1) &= \sigma_i(n) + \beta(n) \sbra{ x_n \hat p(\mu_i|x_n) - \sigma_i(n)}
\end{cases}
\label{eq:oda_learning1}
\end{equation}
where $\sum_n \beta(n) = \infty$, $\sum_n \beta^2(n) < \infty$,
and the quantities $\hat p(\mu_i|x_n)$ and $\mu_i(n)$ 
are recursively updated 
as follows:
\begin{equation}
\begin{aligned}
\mu_i(n) = \frac{\sigma_i(n)}{\rho_i(n)},\quad
\hat p(\mu_i|x_n) = \frac{\rho_i(n) e^{-\frac{d(x_n,\mu_i(n))}{T}}}
			{\sum_i \rho_i(n) e^{-\frac{d(x_n,\mu_i(n))}{T}}} 
\end{aligned}
\label{eq:oda_learning2}
\end{equation}
converges almost surely to a solution of the block optimization 
(\ref{eq:E}), (\ref{eq:mu_star}).
\label{thm:ODA}
\end{corollary}
The learning rule
(\ref{eq:oda_learning1}), (\ref{eq:oda_learning2}) 
is a stochastic approximation algorithm \cite{borkar2009stochastic}.

\subsection{Online Deterministic Annealing for Supervised Learning}

We can extend the proposed learning algorithm 
to be used for classification as well. For the classification problem, 
a pair of random variables 
$\cbra{X,c} \in S\times \cbra{0,1}$ defined in a probability space
	$\pbra{\Omega, \mathcal{F}, \mathbb{P}}$, is observed with
	$c$ representing the class of $X$ and $S\subseteq\mathbb{R}^d$.	
	Let 
	$M := \cbra{\mu_h}_{h = 1}^K$, where $\mu_h \in ri(S_h)$
	represent codevectors, and 
	define the set $C_\mu := \cbra{c_{\mu_h}}_{h = 1}^K$,
	such that $c_{\mu_h} \in \cbra{0,1}$ 
	represents the class of $\mu_h$ for all $h \in \cbra{1,\ldots,K}$.
	The quantizer $Q^c:S \rightarrow \cbra{0,1}$ is defined such that 
	$Q^c(X) = \sum_{h=1}^k c_{\mu_h} \mathds{1}_{\sbra{X \in S_h}}$.	
	Then, the minimum-error classification problem is formulated as
\begin{equation}
		\min_{\cbra{\mu_h,S_h}_{h = 1}^K} ~ J_B(Q^c) := 
		\pi_1 \sum_{H_0} \mathbb{P}_1\sbra{X\in S_h} +
			 \pi_0 \sum_{H_1} \mathbb{P}_0\sbra{X\in S_h} 
		%
		%
\label{eq:lvq}
\end{equation}	
	where $\pi_i := \mathbb{P}\sbra{c = i}, 
	\mathbb{P}_i\cbra{\cdot} := \mathbb{P}\cbra{\cdot | c = i}$,
	and $H_i$ is defined as $H_i := \cbra{h\in\cbra{1,\ldots,K}:Q^c=i}$, 
	$i\in\cbra{0,1}$.
%
In this case we can rewrite the expected distortion as
\begin{align*}
D = \E{d^b(c_X,Q^c)} 
\end{align*}
where $ d^b(c_x,c_\mu) = \mathds{1}_{\sbra{c_x\neq c_\mu}}$.
%
Because $d^b$ is not differentiable, 
using similar principles as in the case of LVQ, 
we can instead approximate the optimal solution by 
solving the minimization 
problem for the following distortion measure
\begin{equation}
d^c(x,c_x,\mu,c_\mu) = \begin{cases}
				d(x,\mu),~ c_x=c_\mu \\
				0,~ c_x\neq c_\mu			
				\end{cases}
\label{eq:dc}
\end{equation}
This particular choice for the distortion measure $d^c$ 
results in important regularization properties (see Section \ref{sSec:Algorithm} and \cite{mavridis2021online}). 
\subsection{The algorithm}
\label{sSec:Algorithm}

The Online Deterministic Annealing (ODA) algorithm %
for clustering and classification 
is depicted in Algorithm \ref{alg:ODA}.
In this section we briefly discuss the properties of the algorithm, as well as the effect of 
the parameters used.
A detailed discussion on the implementation of Alg. \ref{alg:ODA}
and the effect of its parameters
can be found in \cite{mavridis2021online}.

\textit{Temperature Schedule.}
The temperature schedule $T_i$ plays an important role in 
the behavior of the algorithm.
Starting at high temperature
$T_{max}$ ensures the correct operation of the algorithm.
The value of $T_{max}$ depends on the domain of the data. 
The stopping temperature $T_{min}$ can be set a priori or be decided
online depending on the performance of the model at each temperature level. 
It is common practice to use the geometric series $T_{i+1}=\gamma T_i$. 

\textit{Stochastic Approximation.}
Regarding the stochastic approximation stepsizes, 
simple time-based learning rates, 
e.g. of the form $\alpha_n = \nicefrac{1}{a+ bn}$, 
have been sufficient 
for fast convergence in all our experiments so far.
Convergence is checked with the condition
$d_\phi(\mu_i^n,\mu_i^{n-1})<\epsilon_c$
for a given threshold $\epsilon_c$ that can depend on the domain of $X$.

\textit{Bifurcation and Perturbations.}
To every temperature level $T_i$, corresponds a set of 
effective codevectors $\cbra{\mu_j}_{j=1}^{K_i}$, 
which consist of the different solutions 
of the optimization problem (\ref{eq:F}) at $T_i$.
Bifurcation, at $T_i$, is detected by maintaining a pair of perturbed 
codevectors $\cbra{\mu_j+\delta, \mu_j-\delta}$ 
for each effective codevector $\mu_j$ generated at $T_{i-1}$,
i.e. for $j=1\ldots,K_{i-1}$.
Using arguments from variational calculus \cite{rose1998deterministic},
it is easy to see that, upon convegence, 
the perturbed codevectors will merge if a critical 
temperature has not been reached, and will get separated otherwise. 
%
%
Therefore, the cardinality of the model is at most doubled at 
every temperature level.
For classification, a perturbed codevector for each class is generated.

\textit{Regularization.}
Merging is detected by the condition $d_\phi(\mu_j,\mu_i)<\epsilon_n$,
where $\epsilon_n$ is a design parameter
that acts as a regularization term for the model. 
Large values for $\epsilon_n$ (compared to the support of the data $X$)
lead to fewer effective codevectors, while small $\epsilon_n$ 
values lead to a fast growth in the model size, 
which is connected to overfitting.
%
%
An additional regularization mechanism that comes as a natural 
consequence of the stochastic approximation learning rule,
is the detection of idle codevectors.
To see that, notice that the sequence $\rho_i(n)$ resembles an 
approximation of the probability $p(\mu_i,c_{\mu_i})$, and 
whether or not they
become negligible ($\rho_i(n)<\epsilon_r$) 
is a natural criterion for removing the codevector $\mu_i$.
%
The threshold $\epsilon_r$ is a parameter that usually takes 
values near zero.

\textit{Complexity.}
The complexity of Alg. \ref{alg:ODA} for a fixed temperature coefficient $T_i$ is 
$O(N_{c_i} (2K_i)^2 d)$,
where $N_{c_i}$ is the number of stochastic approximation iterations needed for convergence 
which corresponds to the number of data samples observed, 
$K_i$ is the number of codevectors of the model at temperature $T_i$, and 
$d$ is the dimension of the input vectors, i.e., $x\in\mathbb{R}^d$.
Therefore, assuming a training dataset of $N$ samples and
a temperature schedule $\cbra{ T_1=T_{max}, T_2, \ldots, T_{N_T}=T_{min} }$, 
the worst case complexity of Algorithm \ref{alg:ODA} becomes:
\begin{align*}
O(N_{c} (2\bar K)^2 d)
\end{align*}
where $N_c=\max_i \cbra{N_{c_i}}$ is an upper bound on the number of data samples observed
until convergence at each temperature level, and
\begin{align*}
N_T \leq \bar K \leq \min\cbra{ \sum_{n=0}^{N_T-1} 2^n, \sum_{n=0}^{\log_2 K_{max}} 2^n}
< N_T K_{max}
\end{align*}
where the actual value of $\bar K$
depends on the bifurcations occurred as a result of reaching critical temperatures
and the effect of the regularization mechanisms described above.
Note that typically $N_c \ll N$ as a result of the stochastic approximation algorithm,
and $\bar K \ll N_T K_{max}$ as a result of the progressive nature of the ODA 
algorithm.

\textit{Fine-Tuning.}
In practice, 
because the convergence to the Bayes decision surface 
comes at the limit $(K,T)\rightarrow (\infty,0)$, 
a fine-tuning mechanism should be designed to run on top of 
the proposed algorithm after $T_{min}$.
This can be either an LVQ algorithm \cite{Kohonen1995}
or some other local model. 
%


\begin{algorithm}[hb!]
\caption{Online Deterministic Annealing}
\label{alg:ODA}
\begin{algorithmic}
\STATE Select Bregman divergence $d_\phi$
\STATE Set temperature schedule: $T_{max}$, $T_{min}$, $\gamma$
\STATE Decide maximum number of codevectors $K_{max}$ 
\STATE Set convergence parameters: $\cbra{\alpha_n}$, 
	$\epsilon_c$, $\epsilon_n$, $\epsilon_r$, $\delta$ 
\STATE Select initial configuration 
	$ \cbra{\mu^i}: c_{\mu^i} = c,~ \forall c \in\mathcal{C} $
\STATE Initialize: $K = 1$, $T = T_{max}$
\STATE Initialize:	$p(\mu^i) = 1$, $\sigma(\mu^i) = \mu^i p(\mu_i)$, 
	$\forall i$ 	
\WHILE{$K<K_{max}$ \textbf{and} $T>T_{min}$}
\STATE Perturb  
	$\mu^i \gets  
		\cbra{\mu^i+\delta, \mu^i-\delta}$, $\forall i$ 
\STATE Increment $K\gets 2K$
\STATE Update $p(\mu^i)$, $\sigma(\mu^i)\gets\mu^i p(\mu^i)$, $\forall i$ 
\STATE Set $n \gets 0$
\REPEAT 
%
\STATE Observe data point $x$ and class label $c$
\FOR{$i = 1,\ldots, K$} 
\STATE Compute membership $s^i = \mathds{1}_{\sbra{c_{\mu^i}=c}}$ 
\STATE Update: 
\vskip -0.3in
	\begin{align*}
	p(\mu^i|x) &\gets \frac{p(\mu^i) e^{-\frac{d_\phi(x,\mu^i)}{T}}}
			{\sum_i p(\mu^i) e^{-\frac{d_\phi(x,\mu^i)}{T}}} \\
	p(\mu^i) &\gets p(\mu^i) + \alpha_n \sbra{s^i p(\mu^i|x) - p(\mu^i)} \\
	\sigma(\mu^i) &\gets \sigma(\mu^i) + 
		\alpha_n \sbra{s^i x p(\mu^i|x) - \sigma(\mu^i)} \\
	\mu^i &\gets \frac{\sigma(\mu^i)}{p(\mu^i)}	
	\end{align*}
\vspace{-1.5em}
\STATE Increment $n\gets n+1$
\ENDFOR
\UNTIL $d_\phi(\mu^i_n,\mu^i_{n-1})<\epsilon_c$, $\forall i $
\STATE Keep effective codevectors: \\ 
	\quad\quad discard $\mu^i$ if $d_\phi(\mu^j,\mu^i)<\epsilon_n$, 
	$\forall i,j,i\neq j$
\STATE Remove idle codevectors: \\ 
	\quad\quad 	discard $\mu^i$ if $p(\mu^i)<\epsilon_r$, $\forall i$
\STATE Update $K$, $p(\mu^i)$, $\sigma(\mu^i)$, $\forall i$
\STATE Lower temperature $T \gets \gamma T$ 
\ENDWHILE
\end{algorithmic}
\end{algorithm}

\section{Progressive Learning in Multiple Resolutions}
\label{Sec:TS-ODA}

%
As opposed to linear models or simple artificial neural networks
where the forward pass is often computationally inexpensive
scaling linearly with respect to the number of neurons, 
in prototpe-based methods, every 
codevector has to be compared with all $K$ of the codevectors used, 
resulting in a computational complexity that scales quadratically with respect to $K$.
The progressive nature of the online deterministic annealing approach 
partially mitigates this problem, but as $K$ increases, the running time of the 
algorithm is greatly affected. 
To counteract that, we impose a tree structure in the growth of the ODA model,
which will also serve as a means for developing a multi-resolution online deterministic annealing
learning algorithm.

\subsection{Tree-Structured Online Deterministic Annealing}

Tree-structured vector quantizers (TSVQ) provide 
computationally efficient means of compressing
multivariate data. 
%
While
lacking the optimality properties of full-search techniques,
TSVQ are easier to implement and 
give rise to variable-rate codes that
frequently outperform fixed-rate, full-search techniques with
the same average number of bits per sample.
%
%
%
%

The progressive nature of a greedy growing 
tree structure aligns perfectly with the online deterministic annealing algorithm.
In fact, as will be shown, ODA constructs a tree of arbitrary many  
children per parent node 
thus dropping the dependency 
of the number of codevectors to the number of cell splits, 
which is associated with the depth of the tree \cite{riskin1991greedy,nobel1996termination}.
%
%
In other words, every parent cell is split into a progressively
growing number of children cells as the temperature decreases. 
We refer to this as a horizontal split.
Once the temperature coefficient $T$ drops bellow a predefined 
temperature threshold, then the partition is fixed and every children node
will serve as a parent node that will be divided into more cells by 
further application of the ODA algorithm. 
We refer to this as a vertical split.
We note that the simplicity of the splitting criterion 
roots in the annealing nature of ODA. The splitting criterion 
can even be generalized to involve additional terms such as 
the percentage of improvement in accuracy or distortion reduction 
for every temperature step.

As before, let $X: \Omega \rightarrow S$ be a random variable. 
A tree-structured quantizer $Q_\Delta:S \rightarrow S$ 
is defined by a set of codevectors 
$M := \cbra{\mu_h}_{h=1}^{K}$ arranged in a tree structure 
$\Delta$ with a single root node $\nu_0$.
A node $\nu_i$ represents a region $S_{\nu_i}\subseteq S$.  
It is associated with a set of codevectors $M_{\nu_i}:=\cbra{\mu_j}$, where $|M_{\nu_i}|=K_{\nu_i}$,
and a Voronoi partition $V_{\nu_i}:=\cbra{S_j}$ of $S_{\nu_i}$ with respect to $M_{\nu_i}$ and a divergence measure $d$
(see Section \ref{Sec:ODA}).
A tree-structure is a special case of a connected, acyclic directed 
graph, where each node has a single parent node (except for the root node)
and an arbitrary number of children nodes, that is, we do not 
restrict $\Delta$ to be a binary tree. 
The set $C(\nu_i)$ represents the nodes $\cbra{nu_j}$ that are the children of $\nu_i$, 
while the set $P(\nu_j)$ represents the node $\nu_i$ for which $\nu_j \in C(\nu_i)$.
The level $l\geq 0$ of a node $\nu_h \in \Delta$ 
is the length of the path $\cbra{\nu_0, \ldots, \nu_i, \nu_j, \ldots, \nu_h}$
leading from the root node $\nu_0$ to $\nu_h$ such that $\nu_j \in C(\nu_i)$.
To explicitly state the level $l$ of a node $\nu_i$ we will write $\nu_i^{(l)}$. 
The terminal nodes $\tilde N := \cbra{\nu: C(\nu)=\emptyset}$ are called leaves, 
and the union of their associated codevectors will be denoted
$\tilde M:=\cbra{\mu_j}$, where $|\tilde M|=\tilde K$ is the number of leaf codevectors, 
and $\tilde l:=\max \cbra{l:\mu_\nu^{(l)}\in\tilde M,\ \forall \nu} <\infty$ 
will denote the maximum depth of the tree.
The quantizer $Q_\Delta$
defines a hierarchical 
partitioning scheme for the domain $S$, such that
for every node $\nu_i^{(l)} \in \Delta$ associated with the region $S_i$, 
its children nodes $\cbra{\nu^{(l+1)}: \nu^{(l+1)} \in C(\nu_i^{(l)})}$ are associated with
the regions $\cbra{S_j: S_j \in V_{\nu_i}}$ which form a Voronoi partition of $S_{\nu_i}$ with respect to $M_{\nu_i}$.

After a finite number of vertical cell splits, 
the ODA tree $Q_\Delta$
will have a depth $\tilde l<\infty$ and a finite number 
$\tilde K<\infty$ of leaf cells.
A final application of the ODA algorithm (Alg. \ref{alg:ODA}) 
$\tilde l$ times, results, in the limit $K\rightarrow\infty$, 
in a consistent density and 
risk estimator, since all the assumptions needed 
(see \cite{mavridis2021online}) 
carry over to the leaf cells as subsets of the original domain. 
We note that due to the fact that the tree-structured $Q_\Delta$
defines a partition of $S$, the training of the nodes 
is done in parallel by doing one update of the ODA algorithm (Alg. \ref{alg:ODA}) 
in every region $S_i$ associated to $\nu_i$ asynchronously, depending on what region 
each new observation $X$ belongs to.
As a result, depending on the density of the random variable $X$, some cells 
will be visited more often than others, which will result in
some branches of the tree growing faster than others, 
inducing a variable-rate code that
frequently outperforms fixed-rate, full-search techniques with
the same average number of bits per sample.
Alternatively, when learning offline using a dataset, 
all nodes can be trained using parallel processes,
which can be utilized by multi-core computational units.
Algorithm \ref{alg:ts-oda}, 
presents the pseudocode for 
the greedy-growing Tree-Structured 
Online Deterministic Annealing algorithm using multiple-resolutions, 
as will be explained in the next section.

\begin{algorithm}[H]
\caption{Multi-Resolution ODA Algorithm}
\begin{algorithmic}
\STATE Set temperature schedule:
 $\overline T = \cbra{\overline T_{\tilde l}, \overline T_{\tilde l-1}, \ldots, \overline T_{0}}$,
$\underline T = \cbra{\underline T_{\tilde l}, \underline T_{\tilde l-1}, \ldots, \underline T_{0}}$
\STATE Initialize $\nu_0^{(0)}$, $M_{\nu_0}$, $V_{\nu_0}$.
\REPEAT 
\STATE Observe data point $(X,c)$
\STATE $w = \nu_0$, $l=\tilde l$, $x = X_{\tilde l}$
\WHILE {$C(w)\neq \emptyset$}
\STATE $w = v \in C(w)$ such that $x \in S_{v}$
\STATE $l=l-1$
\STATE $x=X_l$
\ENDWHILE 
\STATE Update $M_{w}$ using Alg. \ref{alg:ODA} in $S_{w}$ 
with ($T_{max}=\overline T_l$, $T_{min}=\underline T_l$)
\IF {ODA in $S_{w}$ converged \textbf{and} $l<\tilde l$}
\STATE Split $w$ to $C(w)$ with respect to $V_w$
\ENDIF
\UNTIL Convergence
\end{algorithmic}
\label{alg:ts-oda}
\end{algorithm}


\subsection{Hierarchical Learning with Multi-Resolution ODA}

The tree-growing mechanism detailed above aligns with our intuition to 
use more detailed information to train deeper layers of the tree, 
and can be used to develop a hierarchical multi-resolution learning scheme.  
In the following analysis, we assume that the multi-resolution representation 
is defined by the subspaces $\cbra{V_j, j\in \mathbb{N}}$ induced by the wavelet transform 
described in Section \ref{Sec:Wavelets}. This analysis can be directly generalized to 
other hierarchical dictionaries, including the scattering transform.
Consider the high-resolution representation of the input $X_0\in V_0=S$ 
and the multi-resolution 
approximation defined by $\cbra{ X_1,X_2,\ldots,X_{\tilde l} }$, the projections 
of $X$ to the subspaces $\cbra{ V_1, V_2, \ldots, V_{\tilde l} }$, 
as described in Section \ref{Sec:Wavelets}. 
%
%
The input representations $\cbra{ X_{\tilde l},\ldots, X_1, X_0 }$, 
can be used to train a sequence of vector quantizers 
with optimal sets of codevectors 
$\cbra{ M^{(\tilde 1)},\ldots, M^{(1)}, M^{(0)} }$, respectively, 
where $M^{(l)}:= \cbra{\mu_h^{(l)}}_{h=1}^{K_l}$, and 
$\mu_h^{(l)}\in V_l$.
Since $V_{\tilde l}\subset \ldots \subset V_1 \subset V_0$, 
it follows that $X_l\in V_{\lambda<l}$, 
and, as a result, 
$\mu_h^{(1)}\in V_{\lambda<l}$, for all $l=1,\ldots,\tilde l$.
In other words, since the multi-resolution representation 
defines a sequence of spaces $V_l\subset V_{l-1}$, both the low-resolution 
representations $X_l$ and the corresponding codevectors $M^{(1)}$
belong to a subspace of $V_{\lambda<l}$, and, of course, 
a subspace of the original input space $V_0$.
Because the codevectors $\mu_h^{(l)}\in V_l$ live in a subspace of $V_0$,
we can define the distortion $J(M^{(l)})$ (similar to eq. \ref{eq:VQ})
induced by the optimal codevectors 
in resolution $l$ and measured in $V_0$.
Similarly, we define the
classification error 
$L(M^{(l)})= \pi_1(l) \sum_{H_0(l)} \mathbb{P}_1\sbra{X_0\in S_h(l)} +
			 \pi_0(l) \sum_{H_1(l)} \mathbb{P}_0\sbra{X_0\in S_h(l)}$ 
(similar to \ref{eq:lvq}),
as the probability of error induced by the Voronoi partition of the
codevectors $M^{(l)}$ and measured in $V_0$.
Because of the fact that $V_l\subset V_0$, it is easy to see that
$J(M^{(\lambda)}) \geq J({M^{(l)}})$ and 
$L(M^{(\lambda)}) \geq L({M^{(l)}})$ for $\lambda\geq l$.

In other words, learning with progressively better approximations
yields progressively better classification results, as 
one would expect. 
This fact can be exploited to reduce the computational complexity 
of the learning algorithm following a hierarchical architecture
that mirrors the progressive attention mechanism observed 
in human cognition \cite{taylor2015global,riesenhuber1999hierarchical,sharma2000induction,
ito2004representation,read2002functional,chi2005multiresolution}.
At this point, the feature space that is created
through the wavelet transform (Section \ref{Sec:Wavelets}), 
has a multi-resolution structure,
while the learning module has a progressively growing tree structure.
We make a fusion of the two structures by applying the 
using different 
input representations in each level of the tree, 
by matching the resolution $(l)$ to the depth level $(\tilde l - l)$ 
of each node. 
This idea was first introduced in 
\cite{baras1993efficient} and \cite{baras1994wavelet}.
More specifically, we start with the representation 
corresponding to the coarsest resolution $X_{\tilde l}$ 
of the input (which is a projection in $V_{\tilde l}$),
and apply Alg. \ref{alg:ODA} on the region $S_{\nu_0}$ of the root node $\nu_0$,
until a threshold temperature is reached.
At this point, the level $l=1$ of the tree-structured quantizer 
$Q_\Delta$ has been fixed, and this qualifies as the first vertical split. 
The nodes $C(\nu_0)$ are horizontally split by the application of 
Alg. \ref{alg:ODA} using the input representations 
$X_{\tilde l-1}$, and this procedure continuous until the 
desired distortion/accuracy level is reached, or until 
a stopping criterion is met, 
depending on the cardinality of the neurons (system resources) and 
the temperature of the annealing process.
The multi-resolution online deterministic annealing algorithm is depicted in 
Alg. \ref{alg:ts-oda}.

\begin{figure}[ht]
\centering
\begin{subfigure}[b]{0.24\textwidth}
\centering
\includegraphics[trim=0 0 0 0,clip,width=1.0\textwidth]{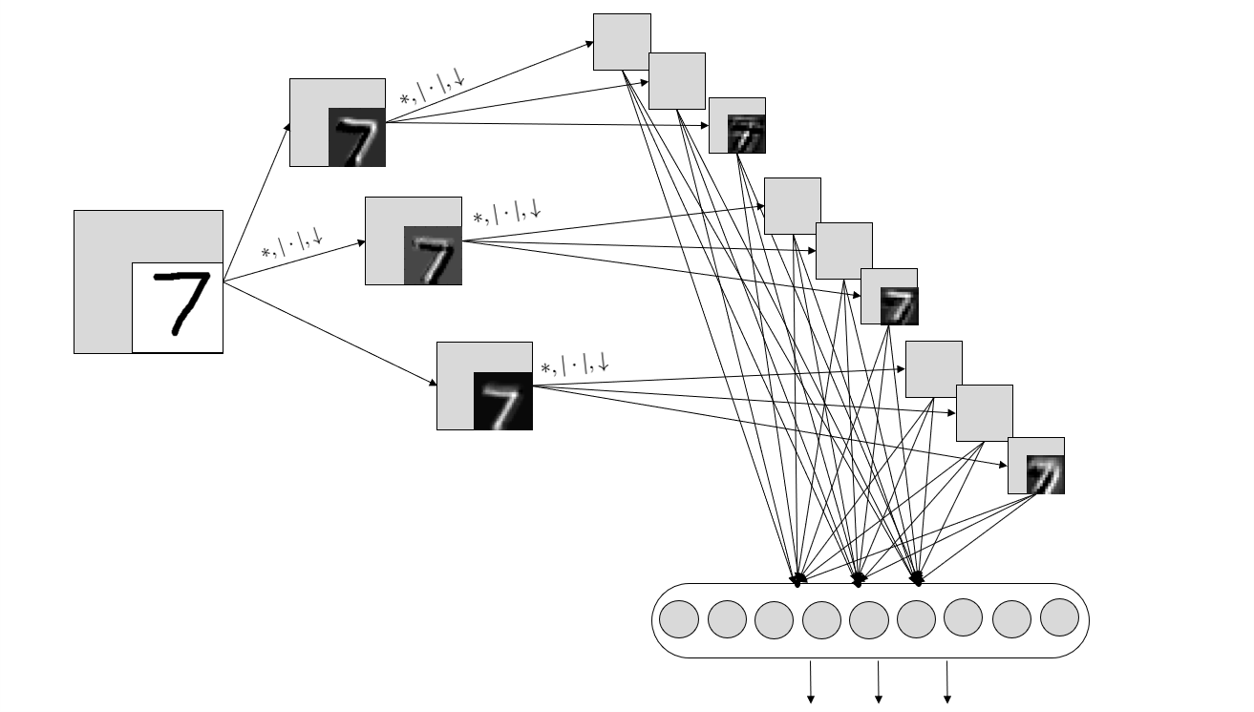}
\caption{DCNN.}
\end{subfigure}
\begin{subfigure}[b]{0.24\textwidth}
\centering
\includegraphics[trim=0 0 0 0,clip,width=1.0\textwidth]{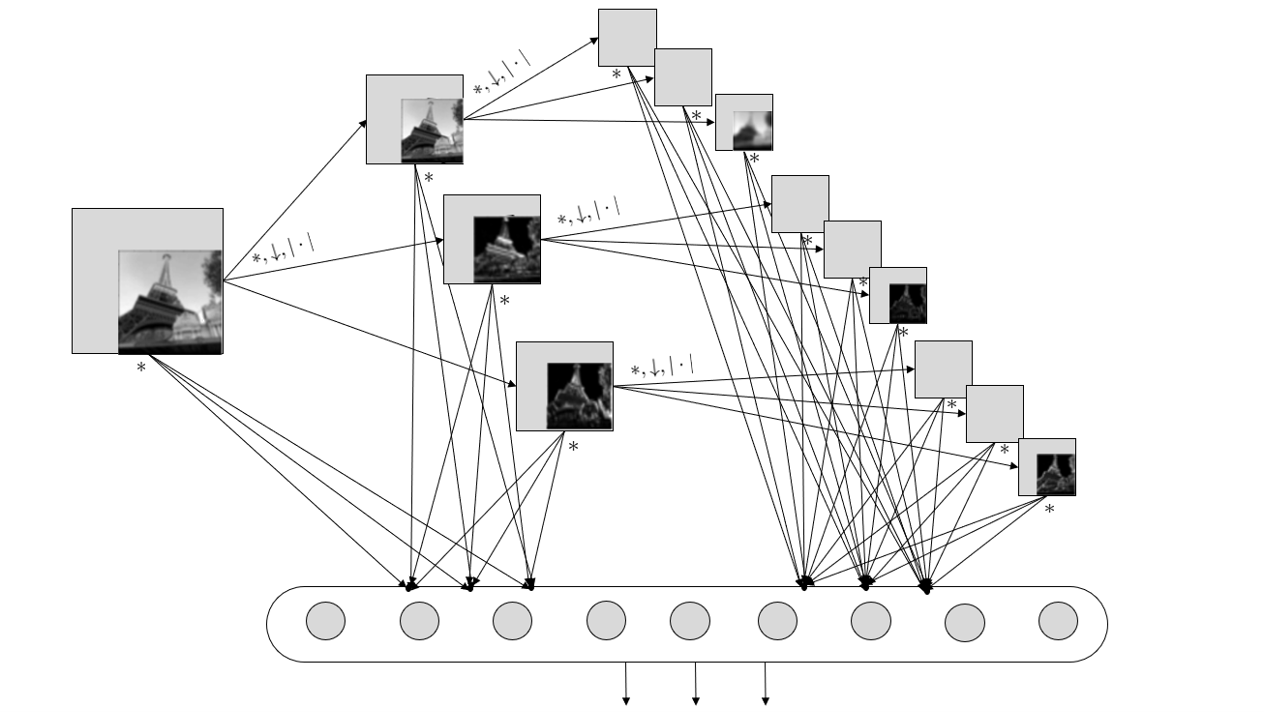}
\caption{SCN.}
\end{subfigure}
%
\caption{The architectures of 
	Deep Convolutional Neural Networks (DCNN) and 
	Scattering Convolutional Networks (SCN).
	The arrows represent a cascade of convolution, rectifying, and downsampling operations.}
\label{fig:TSODAvsDCNNvsSCN_intro}
\end{figure}

\begin{figure}[ht]
\centering
\begin{subfigure}[b]{0.4\textwidth}
\centering
\includegraphics[trim=0 0 0 0,clip,width=1.0\textwidth]{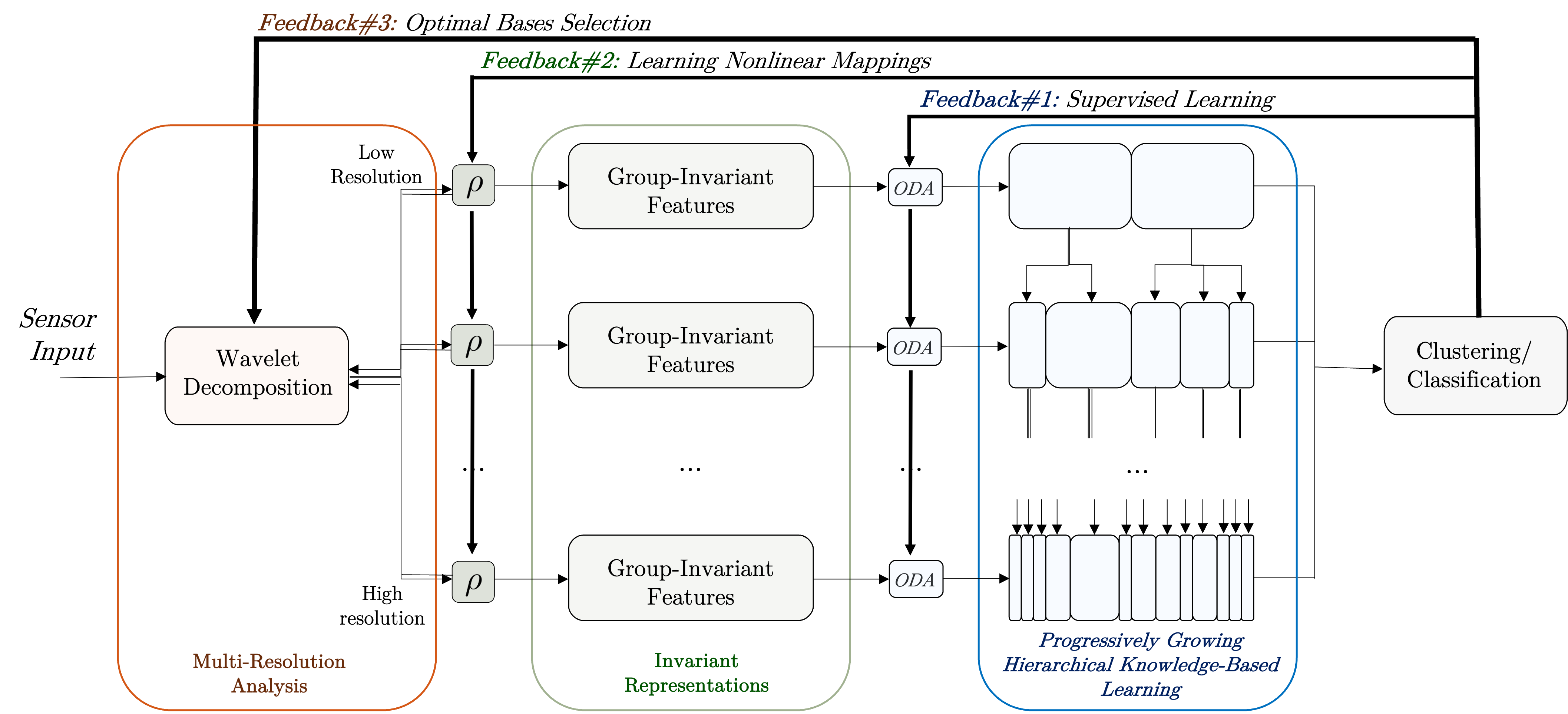}
\caption{System-theoretic block diagram.}
\end{subfigure}
\begin{subfigure}[b]{0.3\textwidth}
\centering
\includegraphics[trim=0 0 0 0,clip,width=1.0\textwidth]{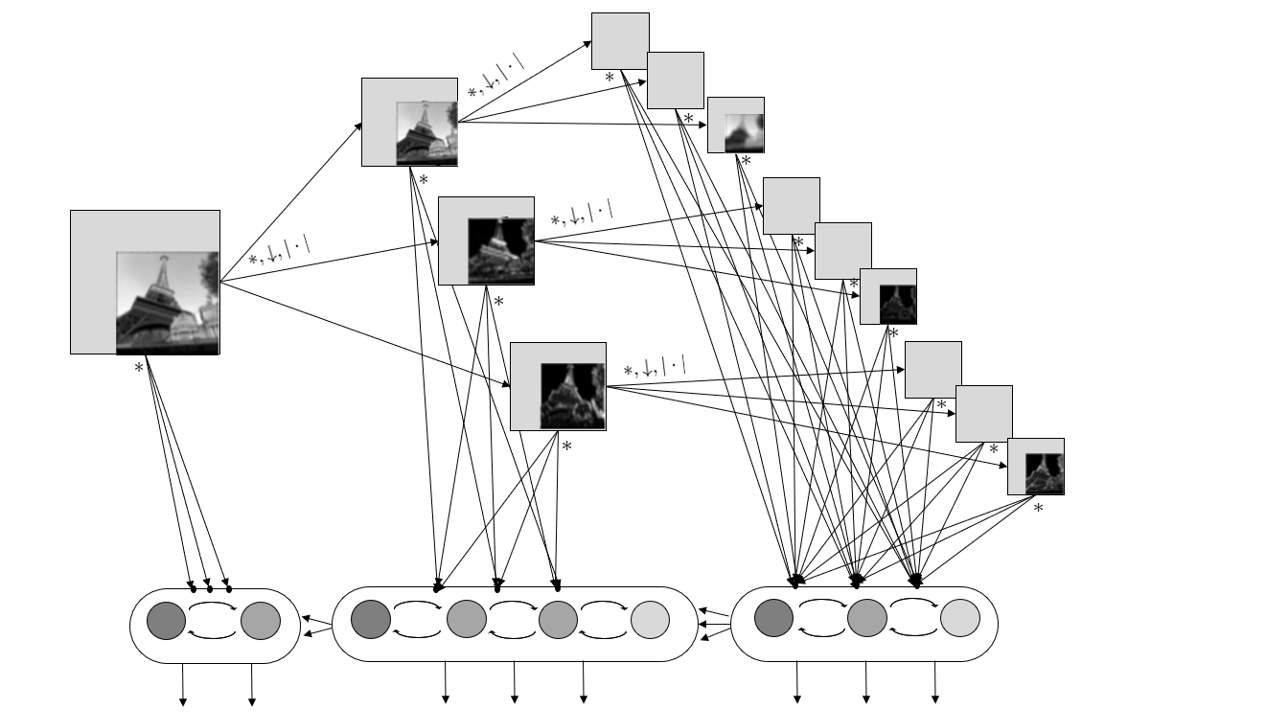}
\caption{Deep neural network architecture.}
\end{subfigure}
\caption{Block diagram of the proposed learning architecture as a closed-loop system (a) and 
as a deep neural network (b).
The arrows represent a cascade of convolution, rectifying, and downsampling operations.}
\label{fig:learning-arch-blue}
\end{figure}

The resulting architecture shares similar properties with 
Deep Convolutional Networks \cite{lecun2015deep} and the
Scattering Convolutional Networks \cite{bruna2013invariant}, 
as shown in Fig. \ref{fig:TSODAvsDCNNvsSCN_intro} and Fig. \ref{fig:learning-arch-blue}, 
and provides a general system-theoretic framework for progressive, hierarchical, 
and knowledge-based learning.
We illustrate this framework in the block diagram 
of Fig. \ref{fig:learning-arch-blue}.

%

\section{Experimental Evaluation and Discussion}
\label{Sec:Results}

We illustrate the properties and evaluate the performance 
of the proposed learning algorithm in widely used artificial and real datasets
for clustering and classification.

\subsection{ODA in One Resolution}

\begin{figure}[t]
\centering
%
%
%
\begin{subfigure}[b]{0.45\textwidth}
\centering
\includegraphics[trim=80 50 60 55,clip,width=0.24\textwidth]{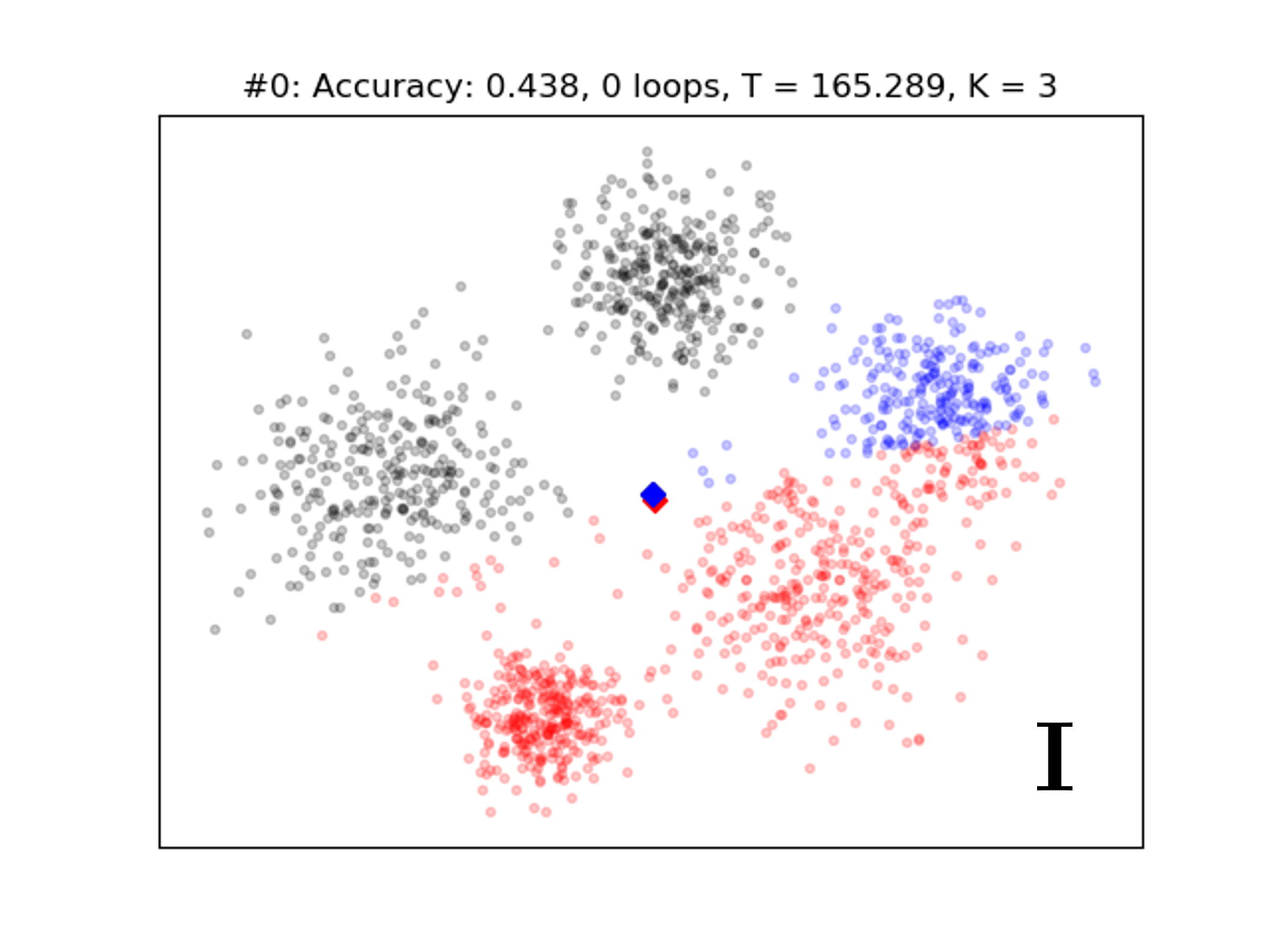}
\includegraphics[trim=80 50 60 55,clip,width=0.24\textwidth]{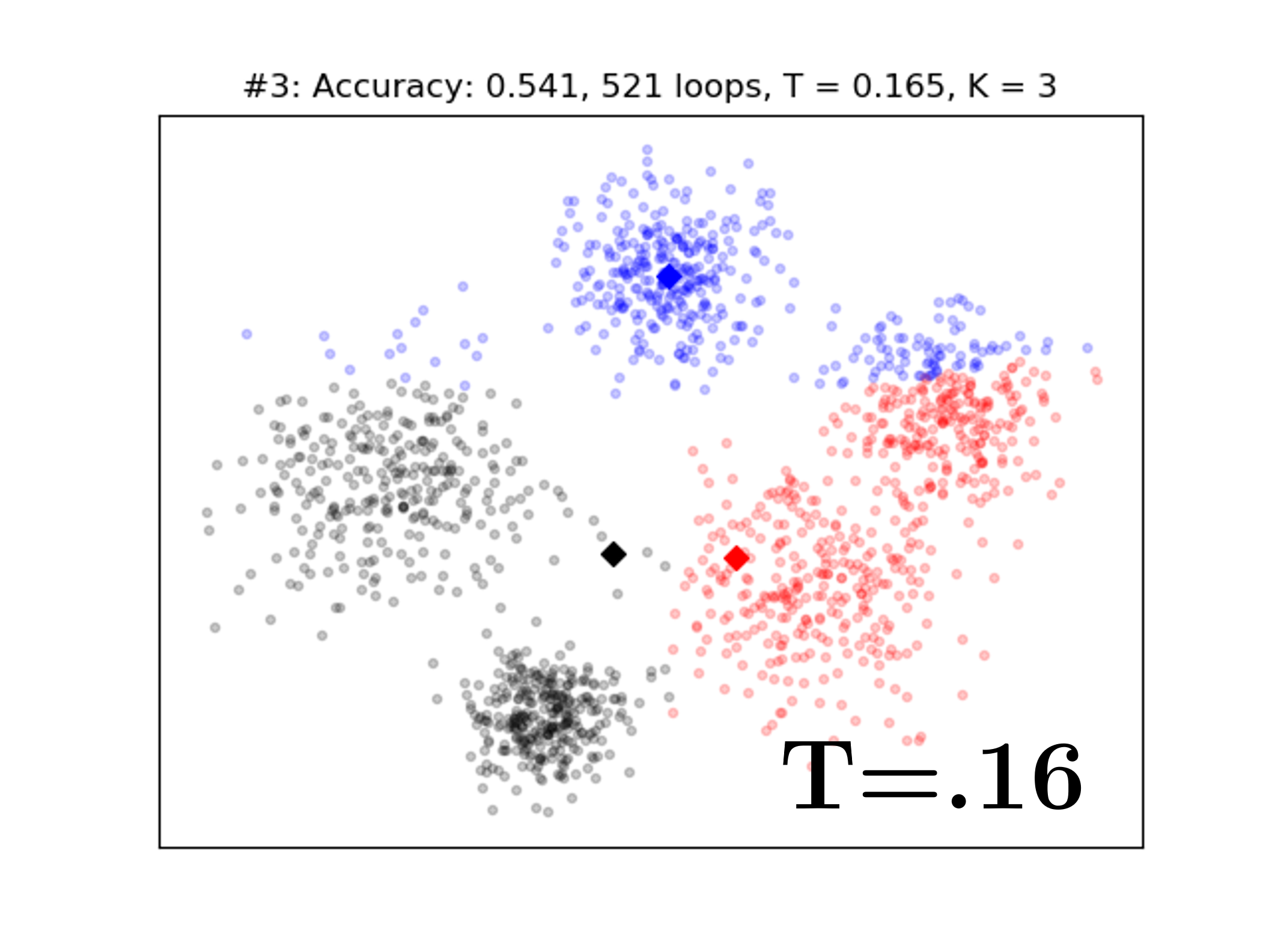}
\includegraphics[trim=80 50 60 55,clip,width=0.24\textwidth]{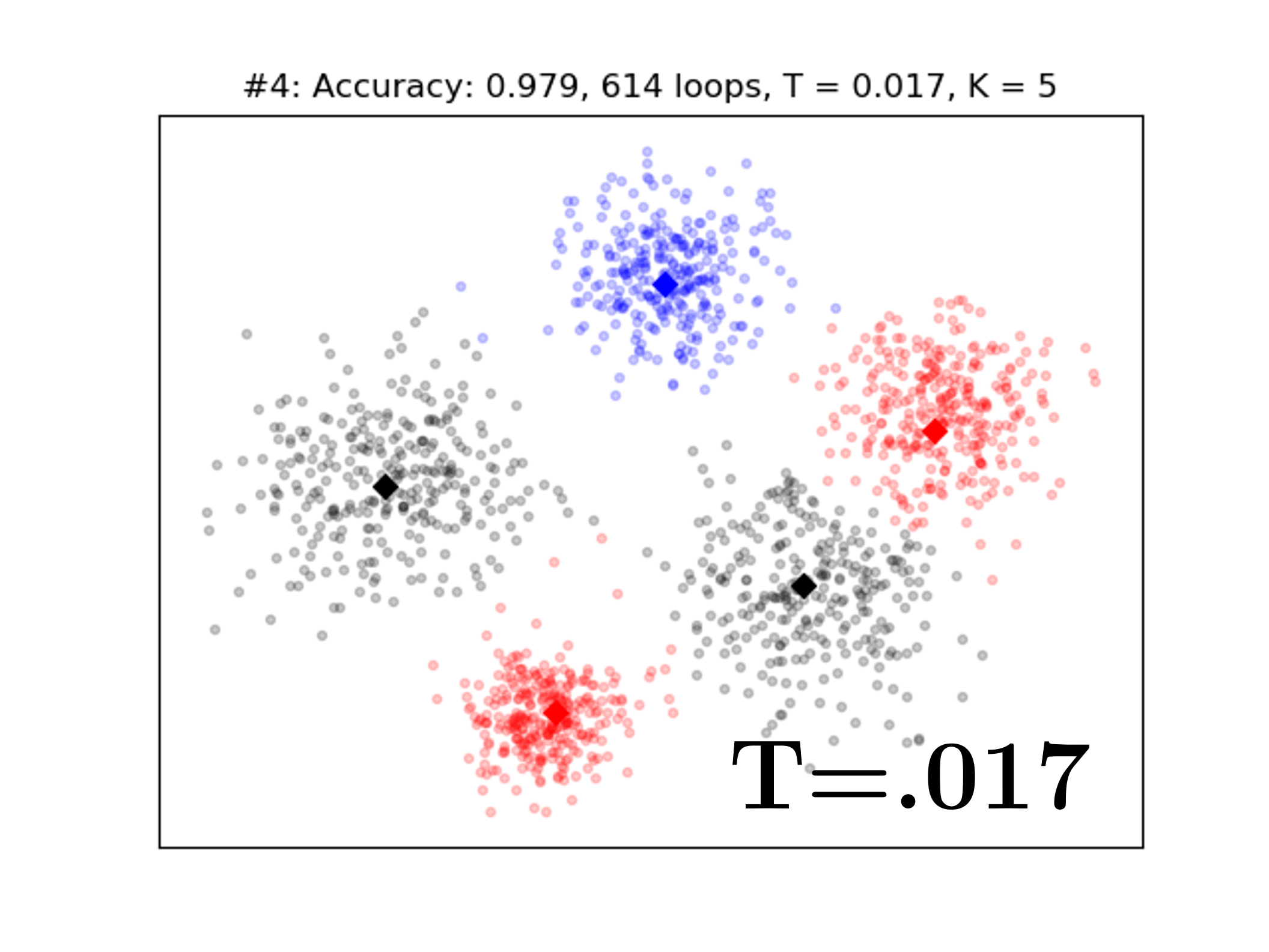}
\includegraphics[trim=80 50 60 55,clip,width=0.24\textwidth]{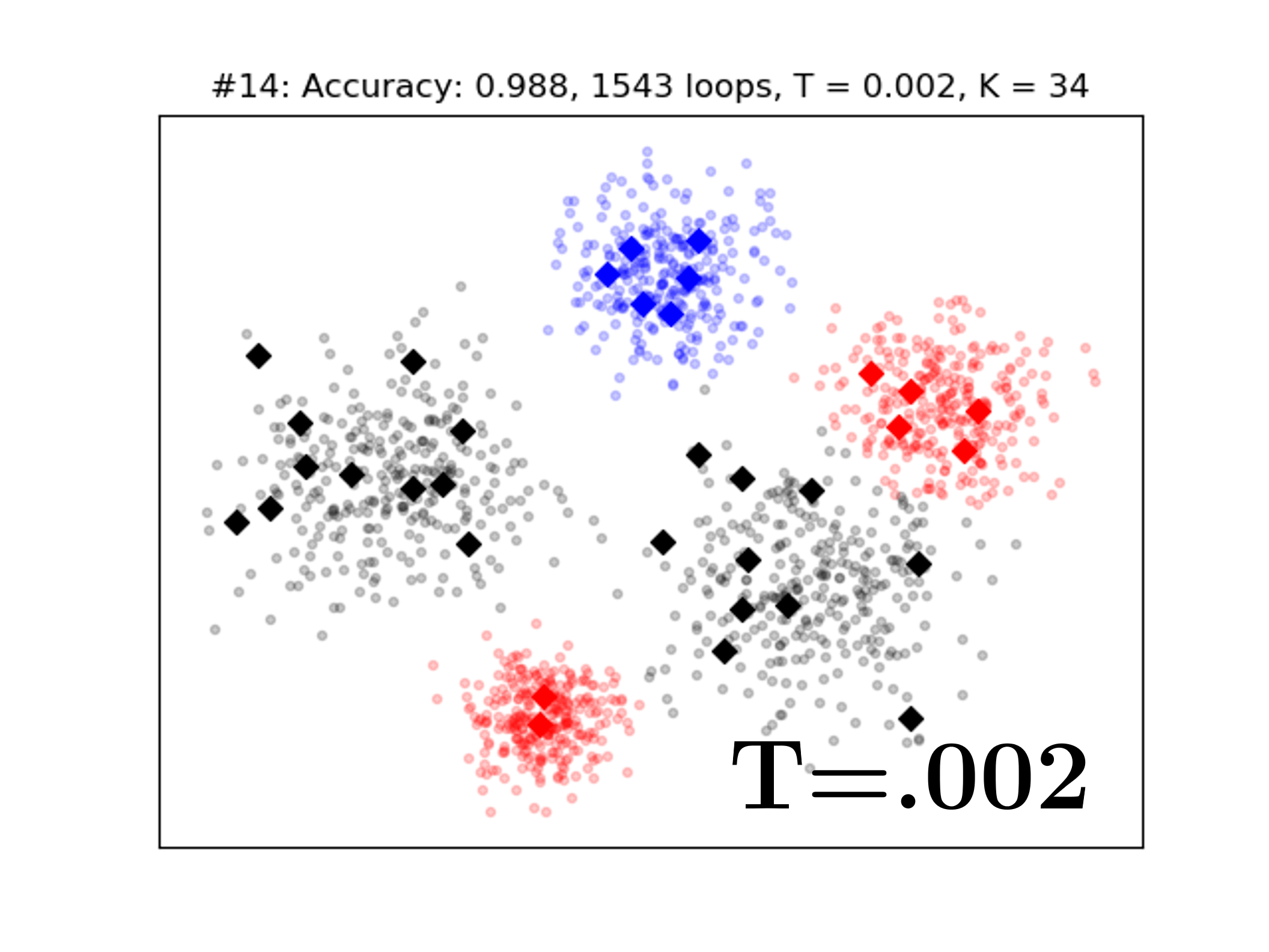}
\caption{Gaussians.}
\label{sfig:R1_blobs}
\end{subfigure}
\begin{subfigure}[b]{0.45\textwidth}
\centering
\includegraphics[trim=70 45 60 55,clip,width=0.24\textwidth]{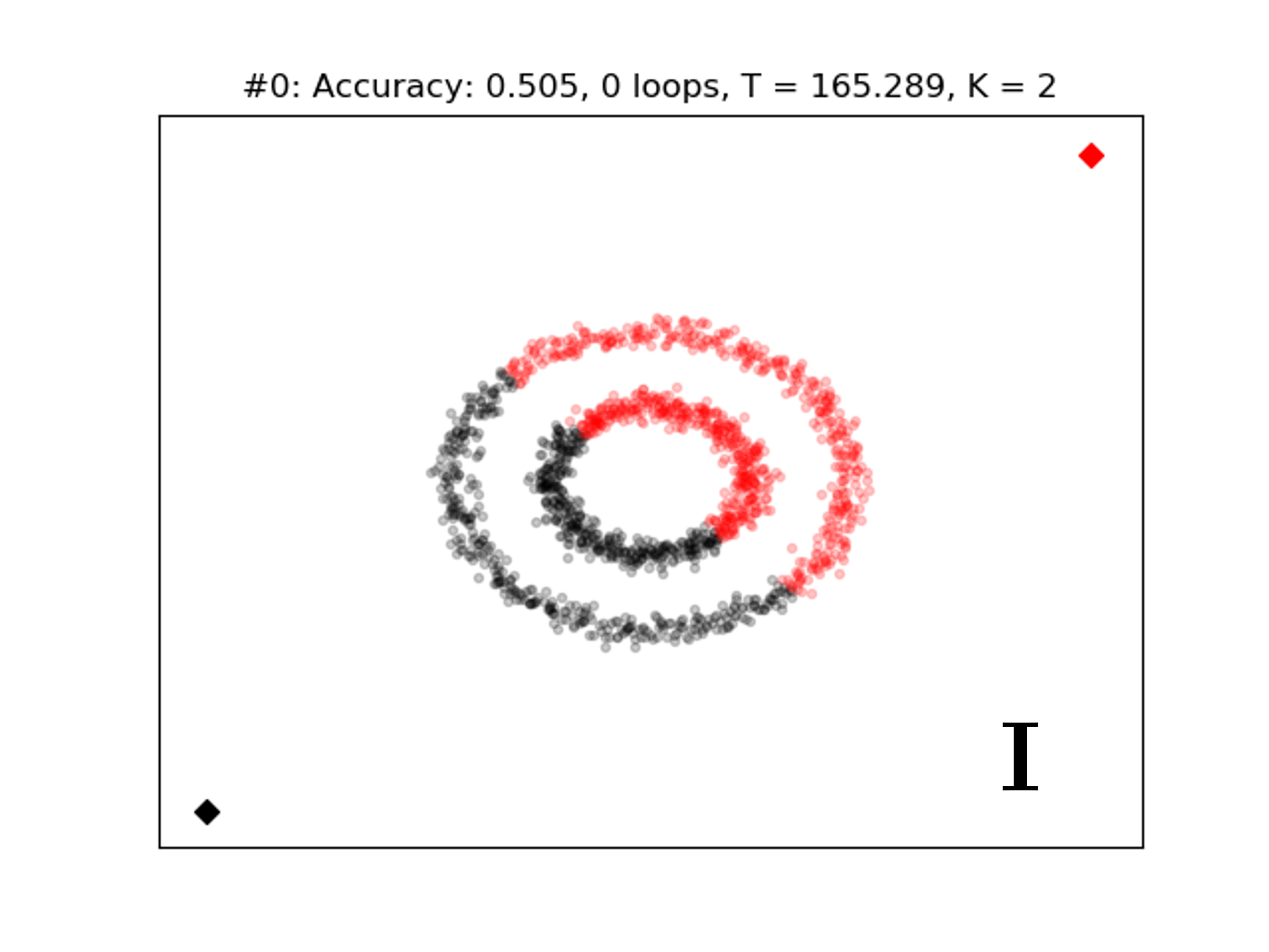}
\includegraphics[trim=70 45 60 55,clip,width=0.24\textwidth]{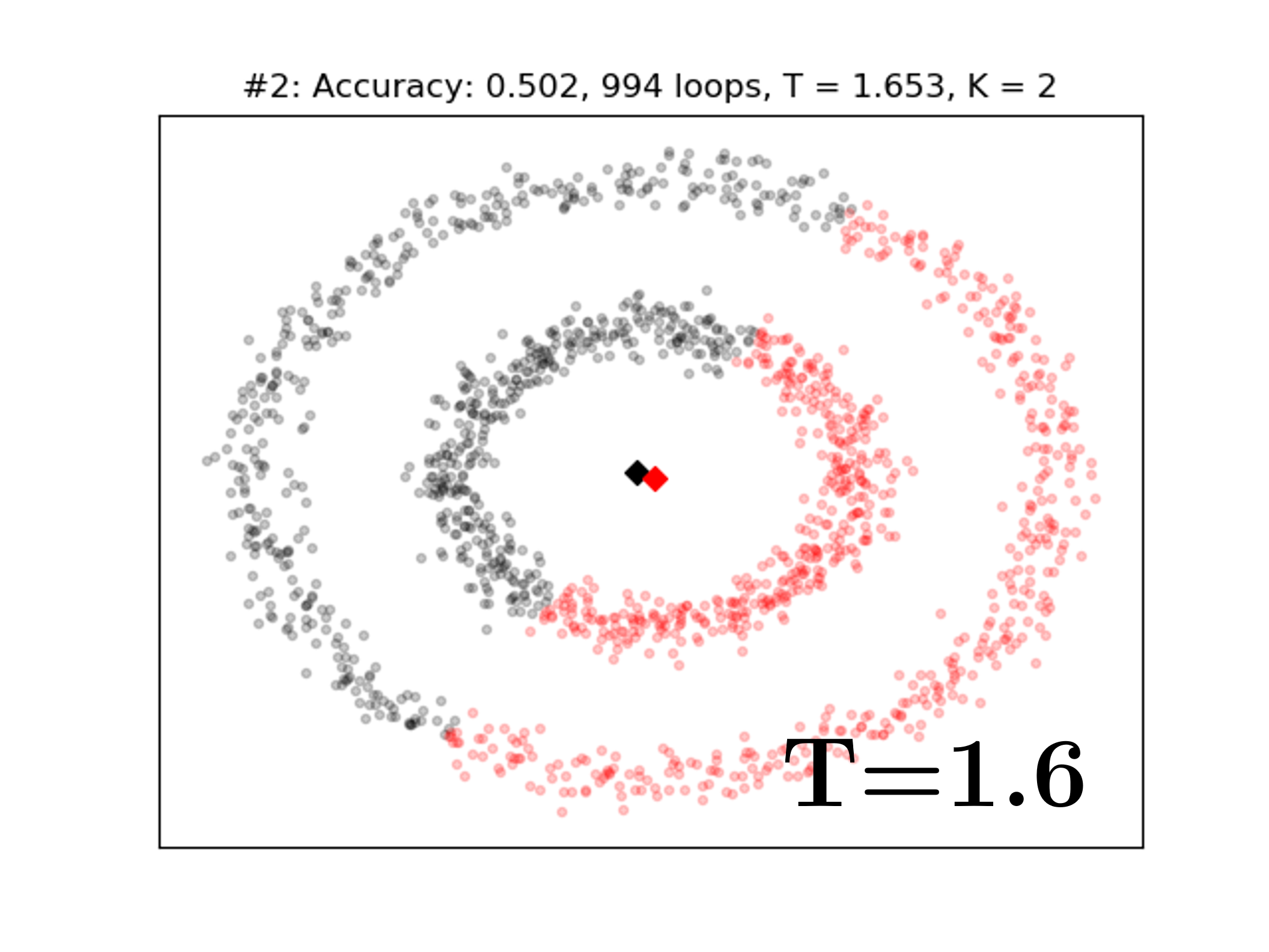}
\includegraphics[trim=70 45 60 55,clip,width=0.24\textwidth]{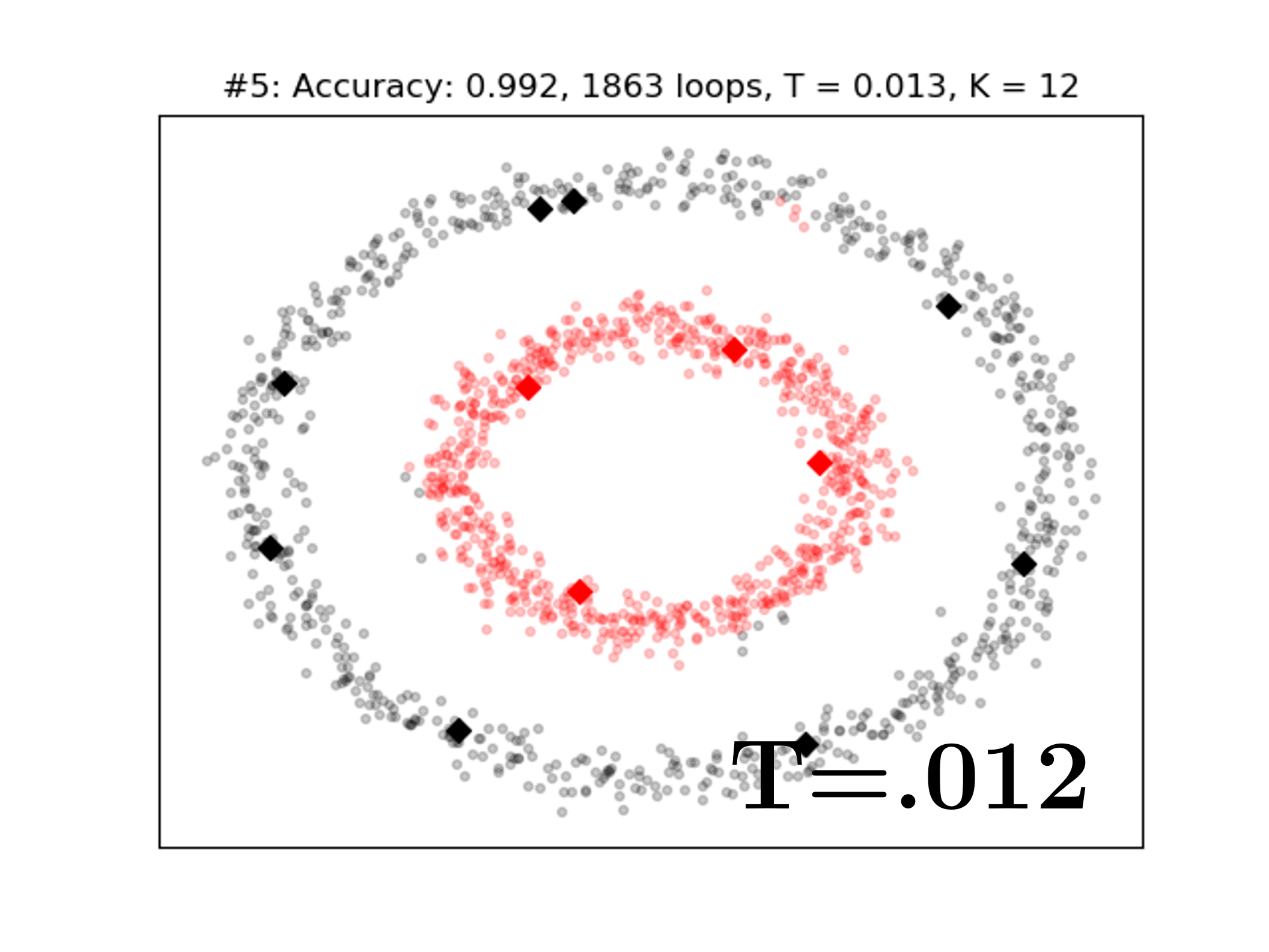}
\includegraphics[trim=70 45 60 55,clip,width=0.24\textwidth]{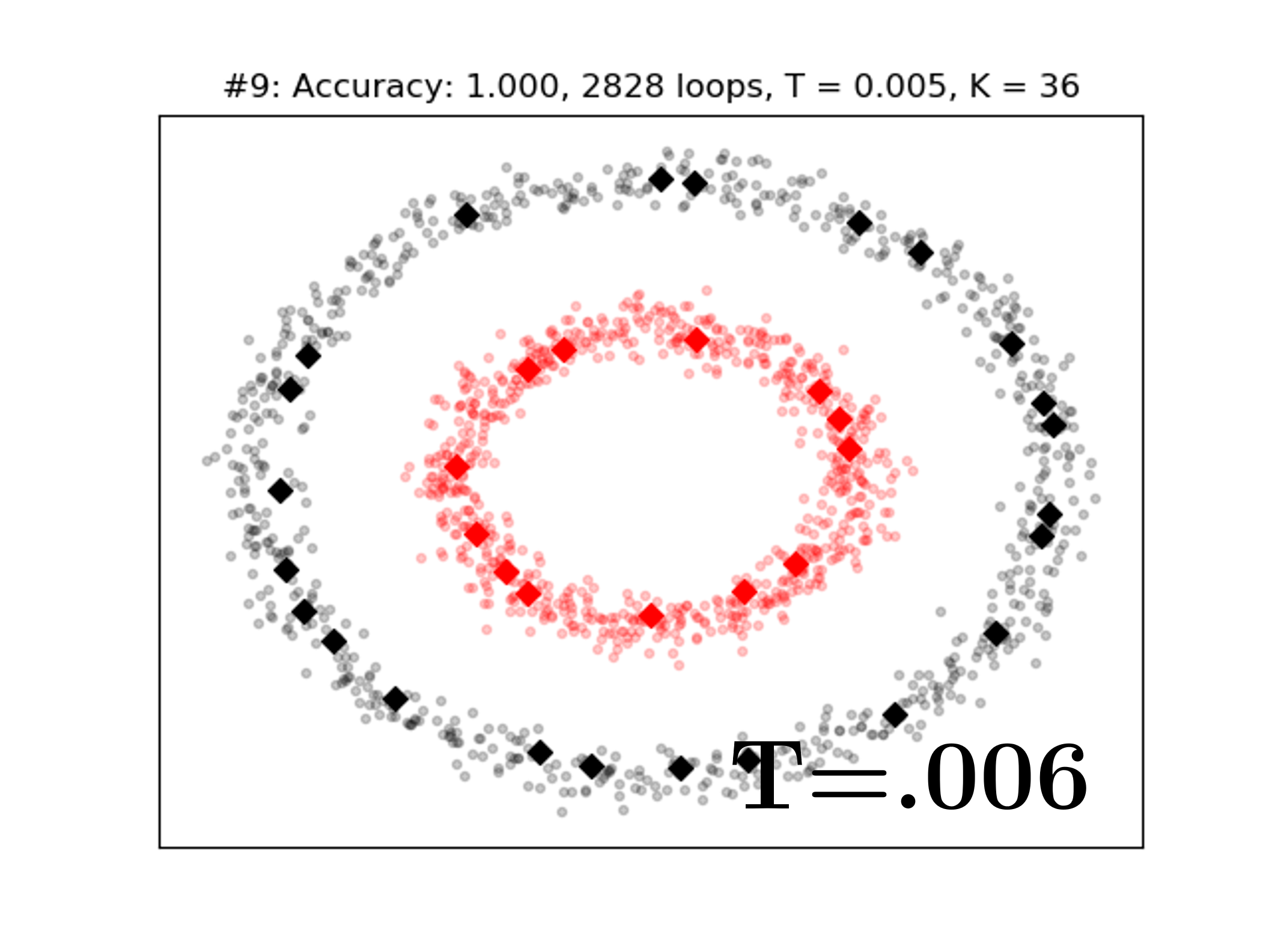}
\caption{Concentric circles.}
\label{sfig:R1_circles}
\end{subfigure}
\caption{Illustration of the evolution of Alg. \ref{alg:ODA} 
	for decreasing temperature $T$ in binary classification in 2D.
	$(b)$ Showcasing robustness with respect to bad initial conditions.}
\label{fig:R1-toy}
\end{figure}

\begin{figure}[h]
\centering
\includegraphics[trim=0 0 0 0,clip,width=0.22\textwidth]{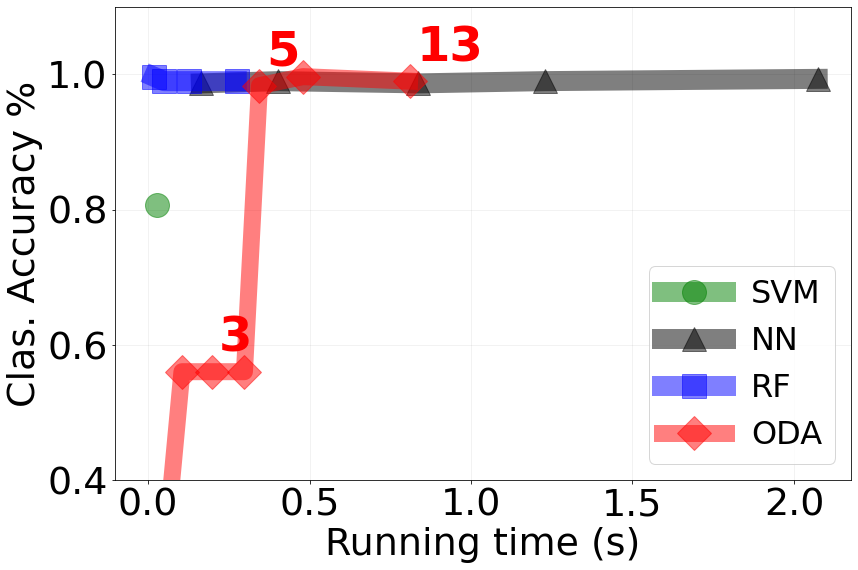}
\includegraphics[trim=0 0 0 0,clip,width=0.25\textwidth]{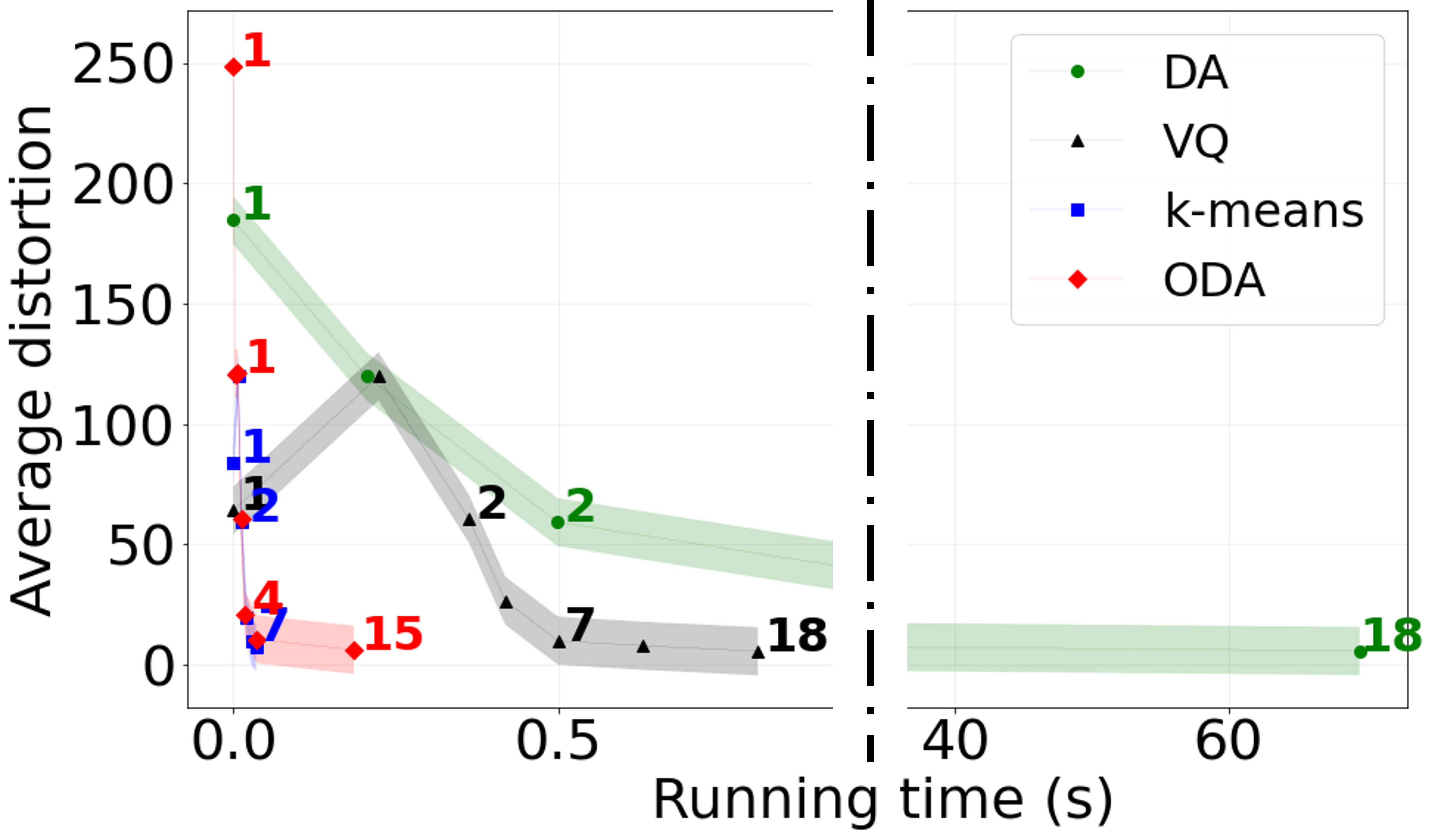}
\caption{Running time of the algorithms in Fig. \ref{sfig:R1_blobs}.}
\label{fig:rtime}
\end{figure}

We first showcase how Alg. \ref{alg:ODA} 
works in a simple, but illustrative, classification problem
in two dimensions (Fig. \ref{fig:R1-toy}).
The underlying class distributions are shaped as
concentric circles and the dataset consist of $1500$ samples.
Since the objective is to give a geometric illustration 
of how the algorithm works in the two-dimensional plane, 
the Euclidean distance is used. 
The algorithm starts at high temperature 
with a single codevector for each class. 
As the temperature coefficient gradually decreases
(Fig. \ref{fig:R1-toy}, from left to right), 
the number of codevectors progressively increases.
The accuracy of the algorithm typically increases as well.
As the temperature goes to zero, the complexity of the model, 
i.e. the number of codevectors,
rapidly increases.
This may, or may not, translate to a corresponding performance boost. 
A single parameter --the temperature $T$-- 
offers online control on this complexity-accuracy trade-off.
Finally, the
robustness of the proposed algorithm with respect to 
the initial configuration is showcased.
Here the codevectors are poorly initialized 
outside the support of the data, which is not assumed known a priori
(e.g. online observations of unknown domain).
In this example the LVQ algorithm has been shown to fail \cite{aLaVigna_LVQconvergence_1990}.
In contrast, the entropy term $H$ in the optimization objective
of Alg. \ref{alg:ODA}, allows for the 
online adaptation to the domain of the dataset 
and helps to prevent poor local minima.  
Finally, the running time of the ODA algorithm for the classification and clustering 
problems of the dataset in Fig. \ref{sfig:R1_blobs} 
is compared in Fig. \ref{fig:rtime} against state-of-the-art algorithms.
All experiments were implemented on a personal computer. 
%
The reader is referred to \cite{mavridis2021online}
for a more detailed assessment of the ODA algorithm (Alg. \ref{alg:ODA}) in both 
classification and clustering, as well as a discussion on the choice of 
the parameters and the limitations of the approach.

%

\subsection{Multi-Resolution Learning}

\begin{figure*}[h]
\centering
\includegraphics[trim=100 35 90 40,clip,width=0.16\textwidth]{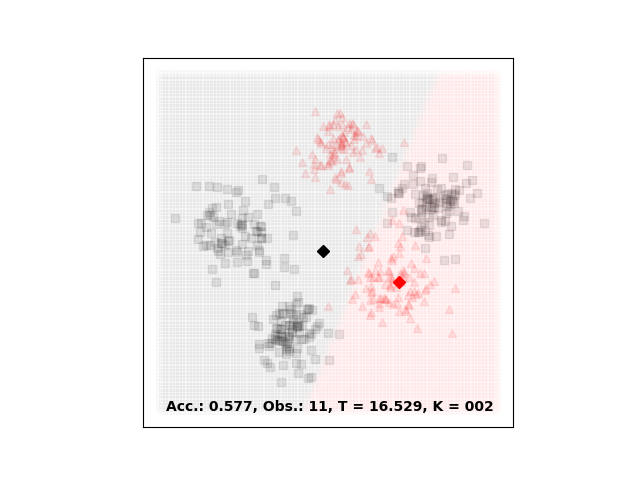}
\includegraphics[trim=100 35 90 40,clip,width=0.16\textwidth]{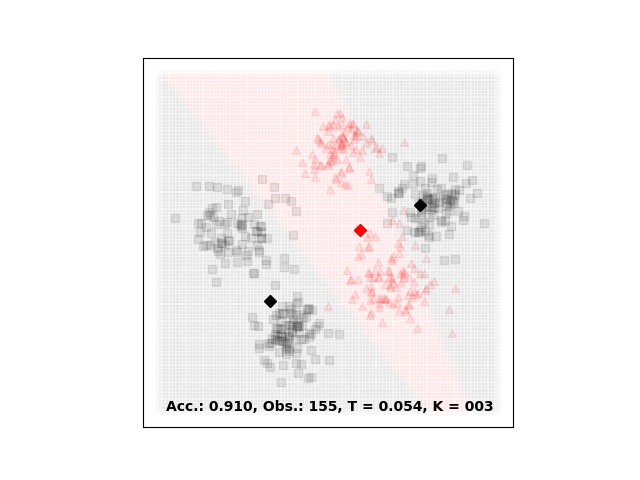}
\includegraphics[trim=100 35 90 40,clip,width=0.16\textwidth]{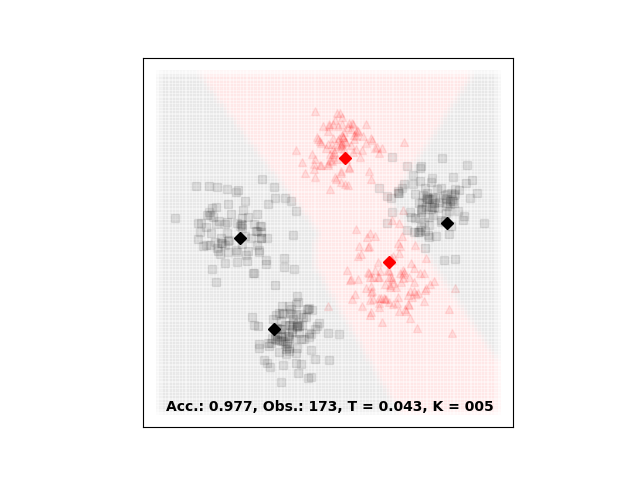}
\includegraphics[trim=100 35 90 40,clip,width=0.16\textwidth]{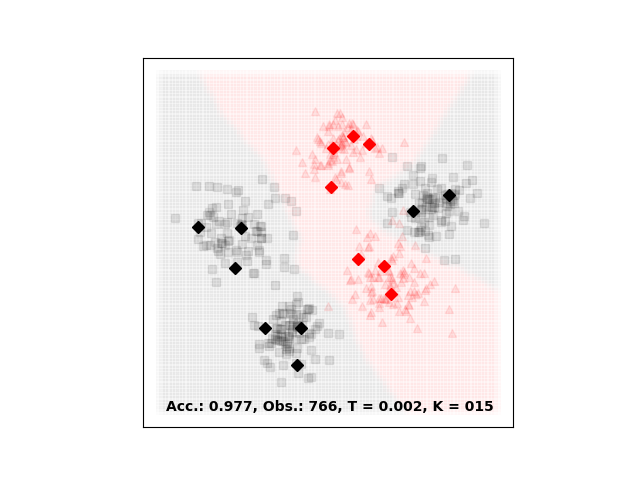}
\includegraphics[trim=100 35 90 40,clip,width=0.16\textwidth]{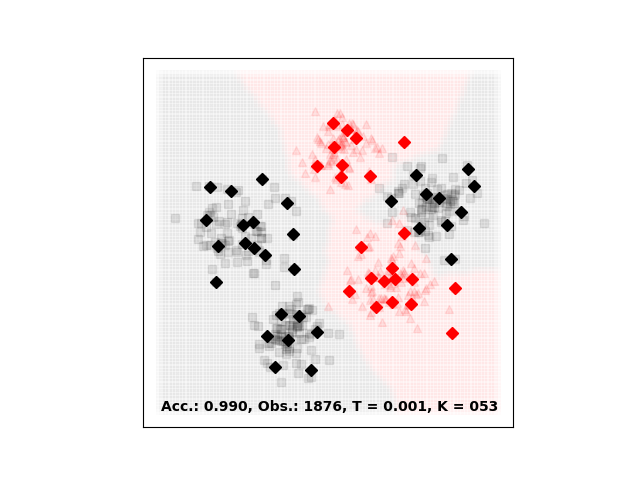}
\includegraphics[trim=100 35 90 40,clip,width=0.16\textwidth]{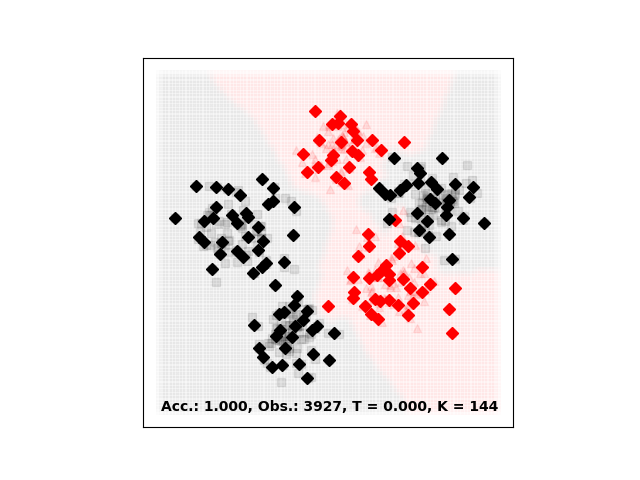}
\caption{Evolution of the ODA algorithm in 2D.}
\label{fig:domain-1}
\end{figure*}

\begin{figure*}[h]
\centering
\includegraphics[trim=100 35 90 40,clip,width=0.16\textwidth]{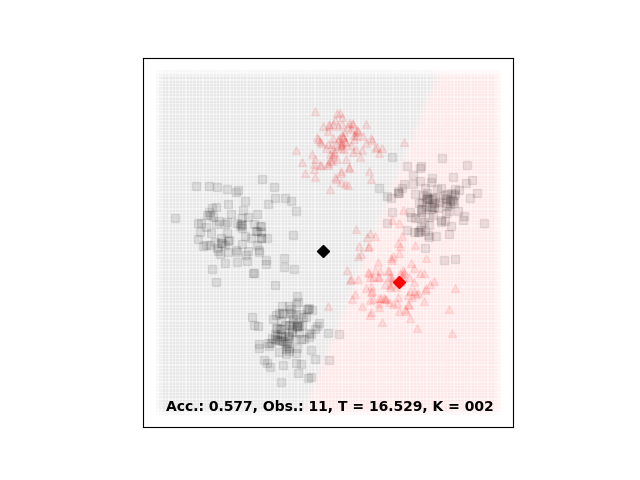}
\includegraphics[trim=100 35 90 40,clip,width=0.16\textwidth]{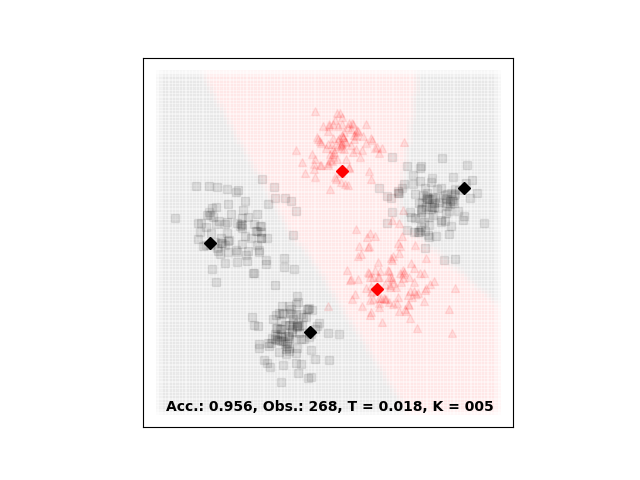}
\includegraphics[trim=100 35 90 40,clip,width=0.16\textwidth]{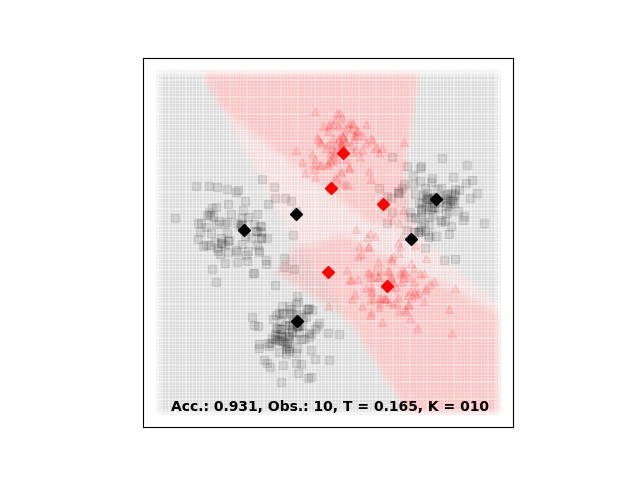}
\includegraphics[trim=100 35 90 40,clip,width=0.16\textwidth]{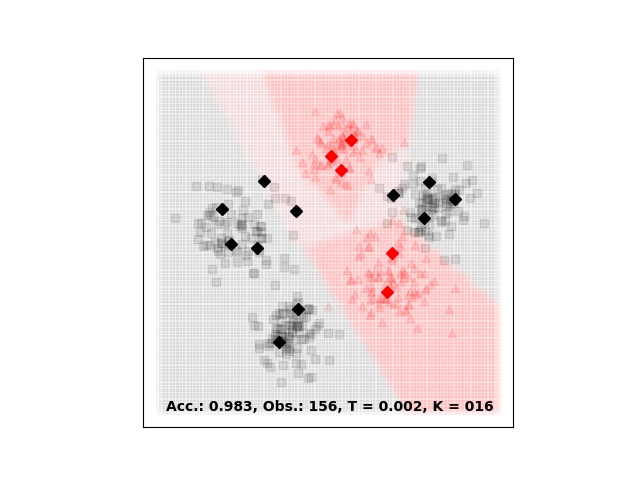}
\includegraphics[trim=100 35 90 40,clip,width=0.16\textwidth]{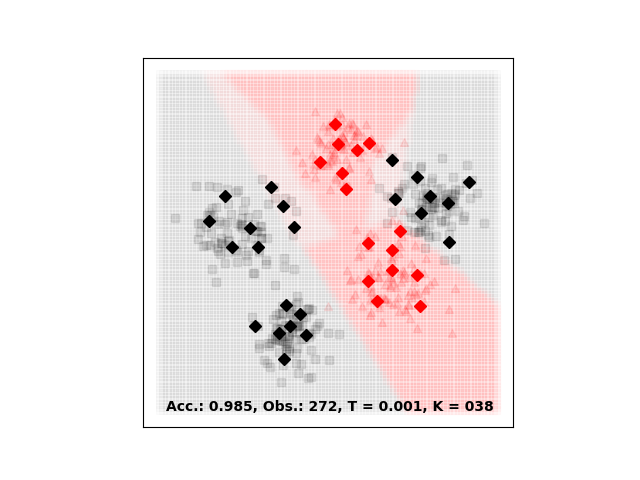}
\includegraphics[trim=100 35 90 40,clip,width=0.16\textwidth]{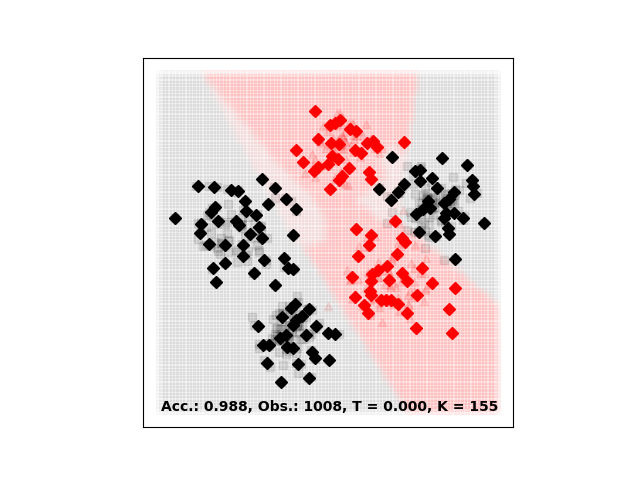}
\caption{Evolution of the Tree-Structured ODA algorithm in a single resolution in 2D.}
\label{fig:domain-11}
\end{figure*}

\begin{figure*}[h]
\centering
\includegraphics[trim=100 35 90 40,clip,width=0.16\textwidth]{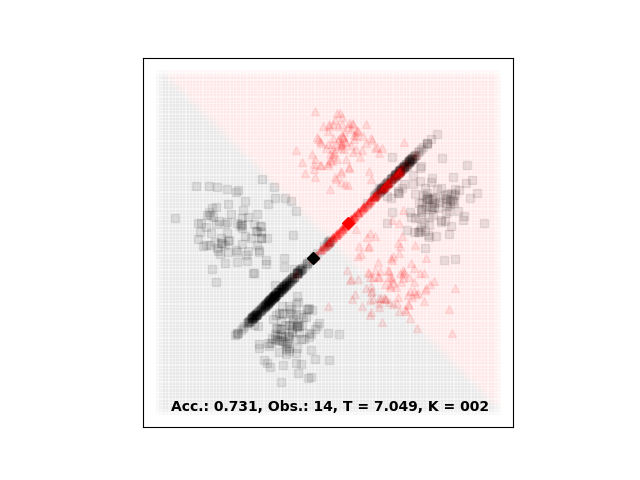}
\includegraphics[trim=100 35 90 40,clip,width=0.16\textwidth]{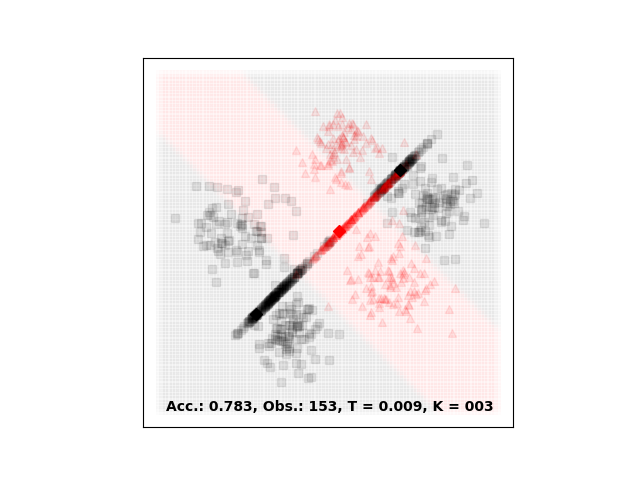}
\includegraphics[trim=100 35 90 40,clip,width=0.16\textwidth]{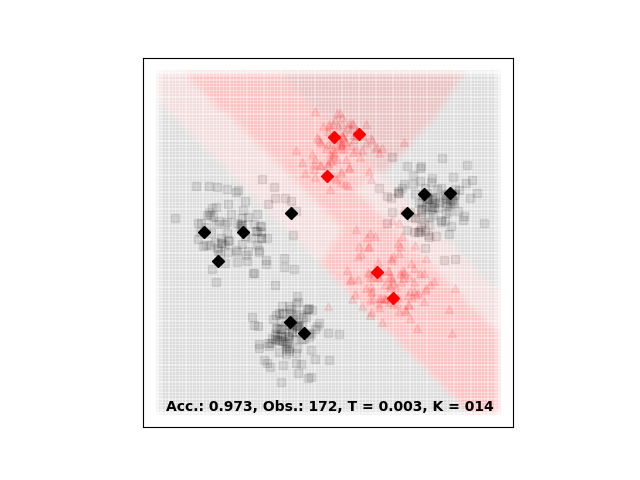}
\includegraphics[trim=100 35 90 40,clip,width=0.16\textwidth]{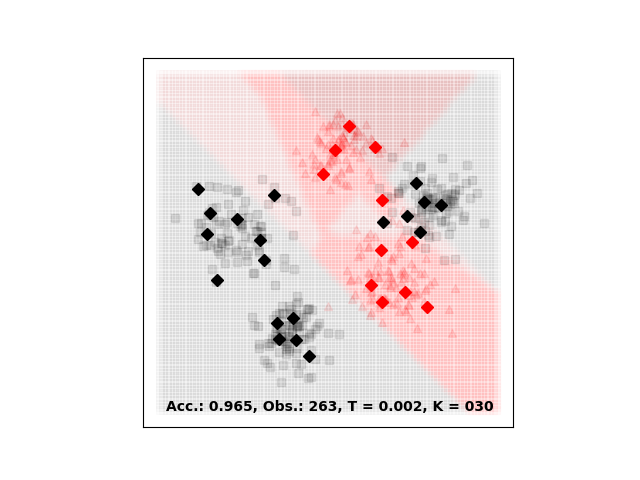}
\includegraphics[trim=100 35 90 40,clip,width=0.16\textwidth]{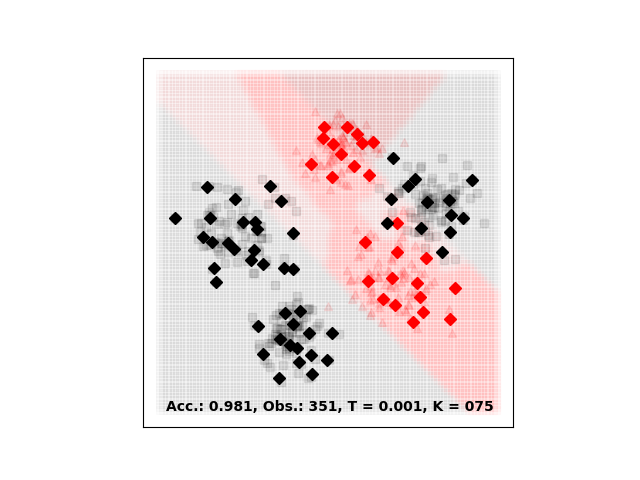}
\includegraphics[trim=100 35 90 40,clip,width=0.16\textwidth]{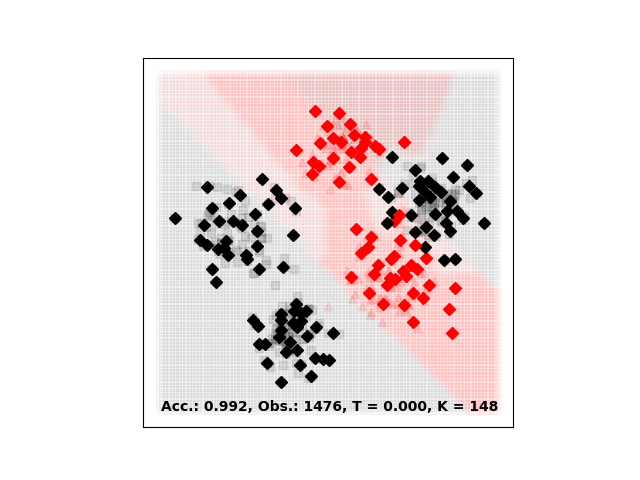}
\caption{Evolution of the Multi-Resolution ODA algorithm in multiple resolutions (1D and 2D).
The low-resolution representations consist of the projections of the data points to the first
principal component of the dataset.}
\label{fig:domain-01}
\end{figure*}

In Fig. \ref{fig:domain-1}, Fig. \ref{fig:domain-11}, and Fig. \ref{fig:domain-01},
we illustrate the evolution of the 
multi-resolution online deterministic annealing algorithm (Alg. \ref{alg:ts-oda}) compared to 
the single-resolution ODA algorithm and the tree-structured ODA algorithm, 
in a 2D classification problem.
The accuracy of the model, the number of observations needed for convergence and the number 
of codevectors (neurons) used at each temperature level, are also shown.

We stress that the computational benefit of using a tree-structured architecture is significant. 
We remind the readers that the complexity of the ODA algorithm is $O(N_c(2\bar K)^2 d)$, 
where $N_c$ is an upper bound on 
the number of data samples observed until convergence, and $\bar K$ the number of effective codevectors
used by the algorithm.
The partition of the space in smaller sets (Voronoi regions) results in easier to solve, localized ODA problems.
In particular, the size of each Voronoi set decreases in a geometric rate with respect to the depth of the tree.
A decrease of similar magnitude is observed on the number $N_c$.
If an ODA algorithm needs $\bar K$ codevectors on the original domain $S$,
a tree-structured ODA algorithm with $\tilde l$ layers 
will require $O(\bar K^{1/\tilde l})$ number of codevectors,
assuming the same number of codevectors in each cell.

Another advantage of using a tree-structured learning module is the localization properties. 
%
The Voronoi regions shrink quickly, 
and allow for the use of local models, which is especially important in high-dimensional spaces. 
Unlike most learning models, it is possible to locate the area of the data space that 
presents the highest error rate and selectively split it by using local ODA.
This process can be iterated until the desired error rate (or average distortion) is achieved.
When using a training dataset for classification, it is often possible to force 
accuracy of up to $100\%$ 
on the training dataset. 
This is similar to an over-fitted classification and regression tree (CART)
\cite{breiman2001random}.

However, over-fitting on the training dataset often adversely affects the 
generalization properties of the model, the performance on the 
testing dataset, and the robustness against adversarial attacks. 
Therefore, the progressive process of ODA becomes important in establishing a robust way to 
control the trade-off between performance and complexity, before you reach that limit.
Finally, an important question in tree-structured learning models is 
the question of which cell to split next.
An exhaustive search in the entire tree to find the node that presents the largest error rate 
is possible but is often not desired due to the large computational overhead. 
This is automatically answered by the 
multi-resolution ODA algorithm (Alg. \ref{alg:ts-oda}) as it asynchronously updates all cells 
depending on the sequence of the online observations. 
As a result, the regions of the data space that are more densely populated with data samples 
are trained first, which results in a higher percentage of performance increase per cell split.
We stress that this property makes the proposed algorithm completely dataset-agnostic, 
in the sense that it does not require the knowledge of 
a training dataset a priori, but instead operates completely online, i.e., using one observation at a time 
to update its knowledge base.

\begin{figure*}[ht]
\centering
\begin{subfigure}[b]{0.9\textwidth}
\centering
\includegraphics[trim=0 0 0 0,clip,width=0.06\textwidth]{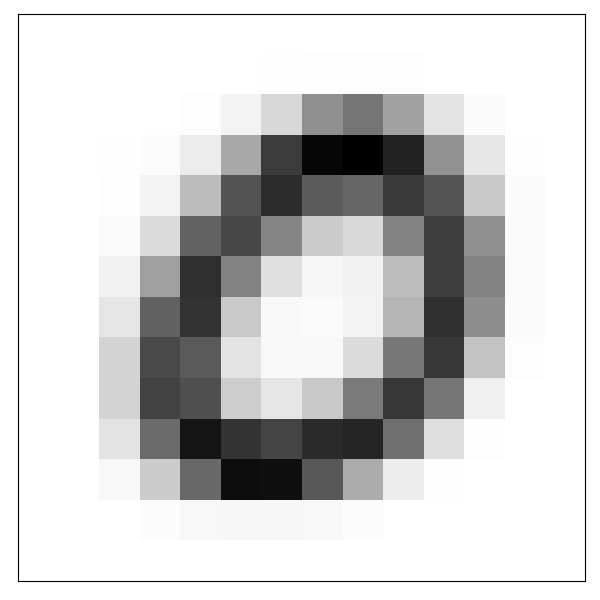}
\includegraphics[trim=0 0 0 0,clip,width=0.06\textwidth]{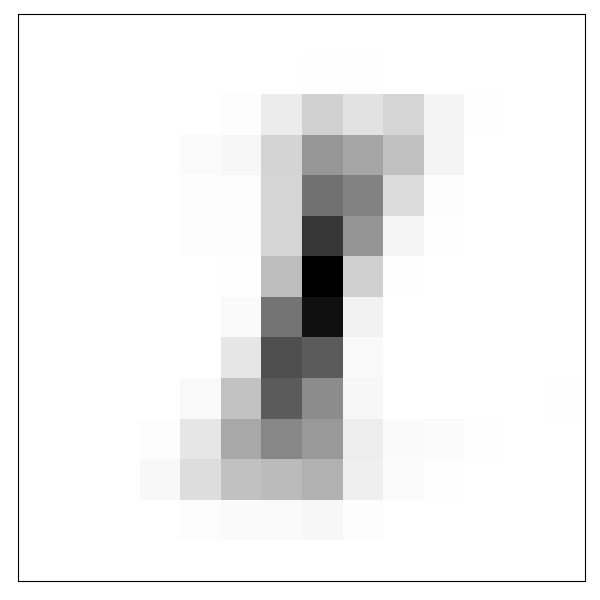}
\includegraphics[trim=0 0 0 0,clip,width=0.06\textwidth]{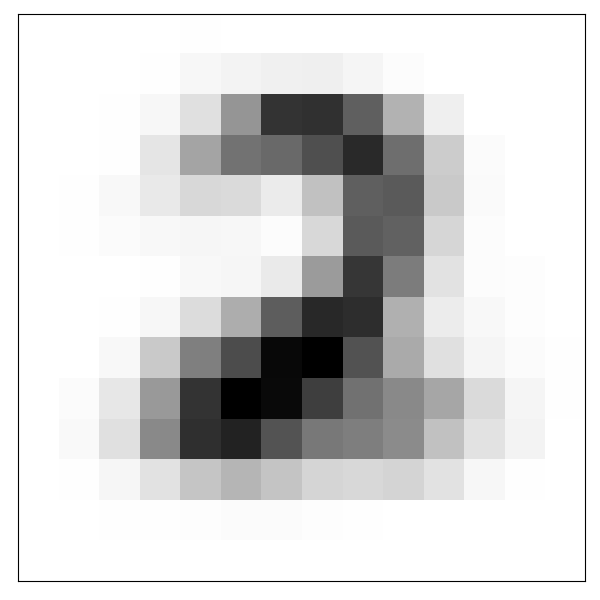}
\includegraphics[trim=0 0 0 0,clip,width=0.06\textwidth]{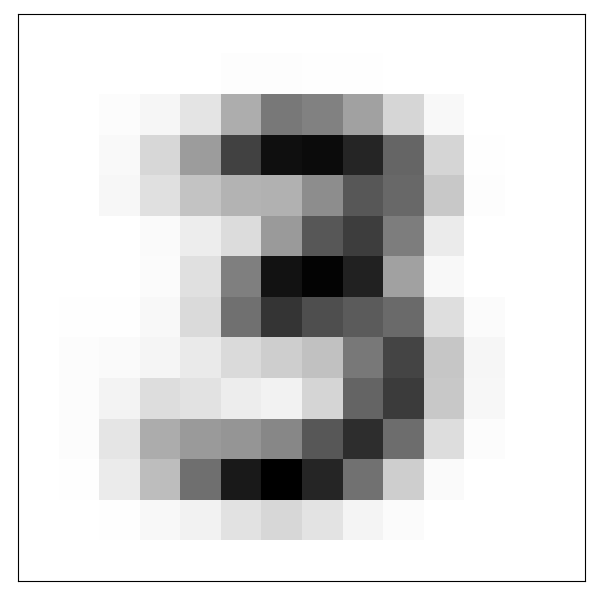}
\includegraphics[trim=0 0 0 0,clip,width=0.06\textwidth]{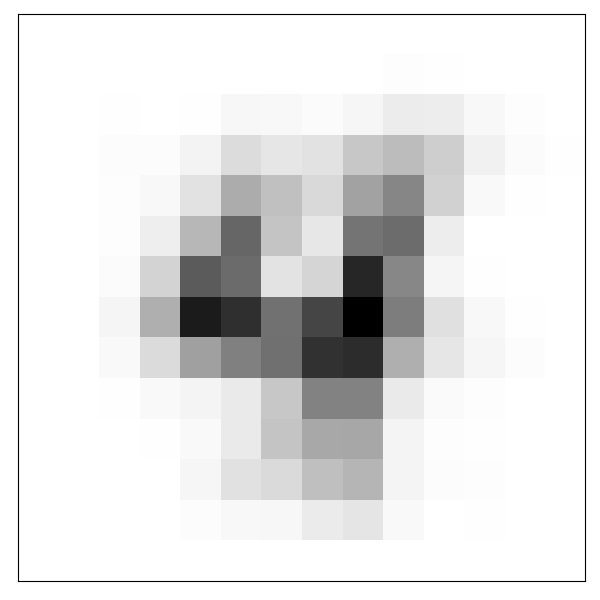}
\includegraphics[trim=0 0 0 0,clip,width=0.06\textwidth]{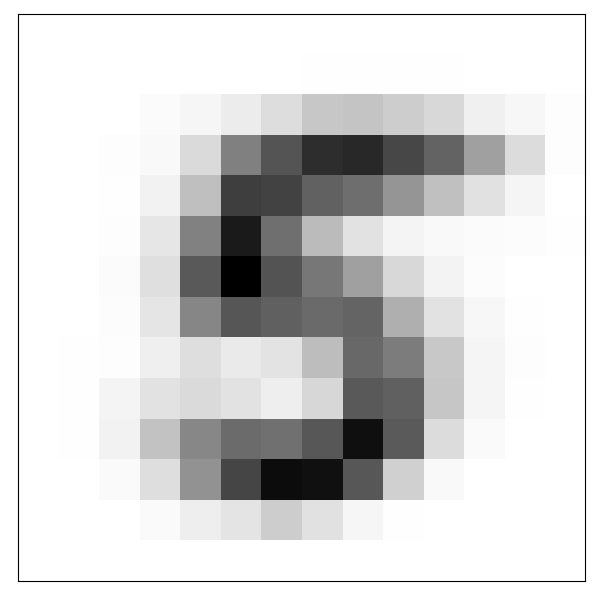}
\includegraphics[trim=0 0 0 0,clip,width=0.06\textwidth]{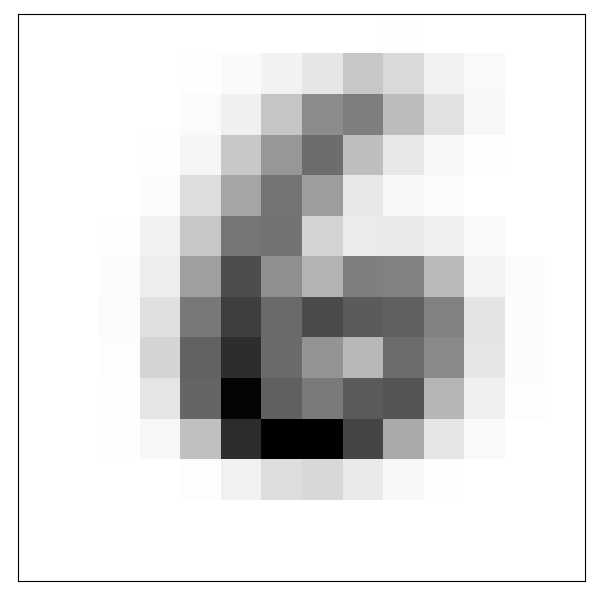}
\includegraphics[trim=0 0 0 0,clip,width=0.06\textwidth]{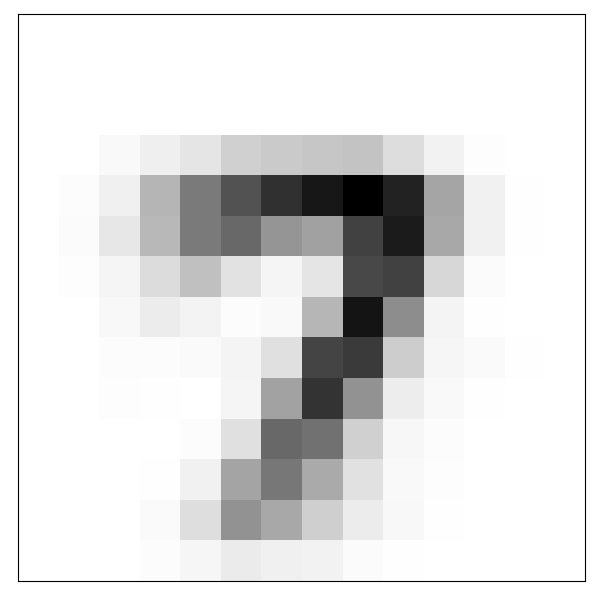}
\includegraphics[trim=0 0 0 0,clip,width=0.06\textwidth]{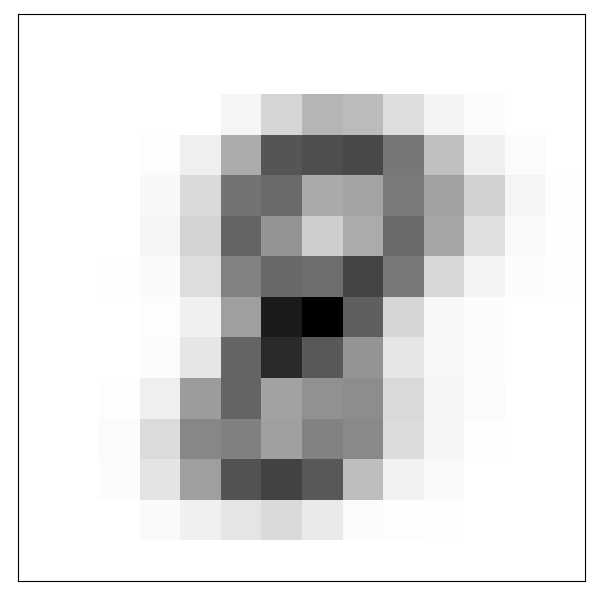}
\includegraphics[trim=0 0 0 0,clip,width=0.06\textwidth]{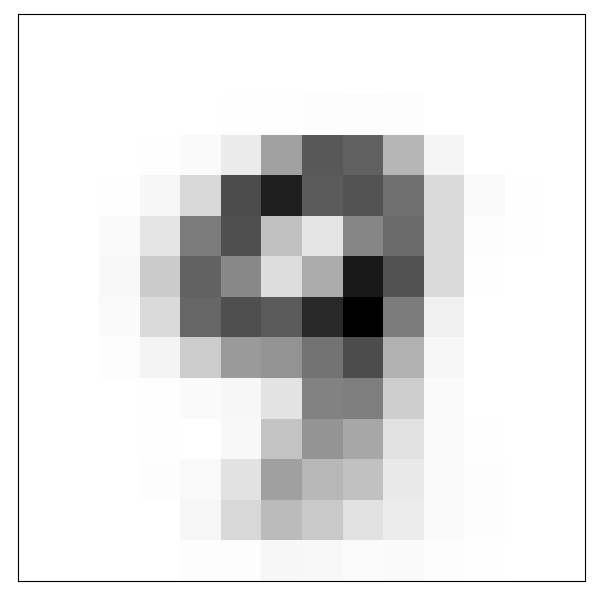}
\caption{First layer ($l=1$).}
\end{subfigure}
\begin{subfigure}[b]{0.15\textwidth}
\centering
\includegraphics[trim=0 0 0 0,clip,width=0.25\textwidth]{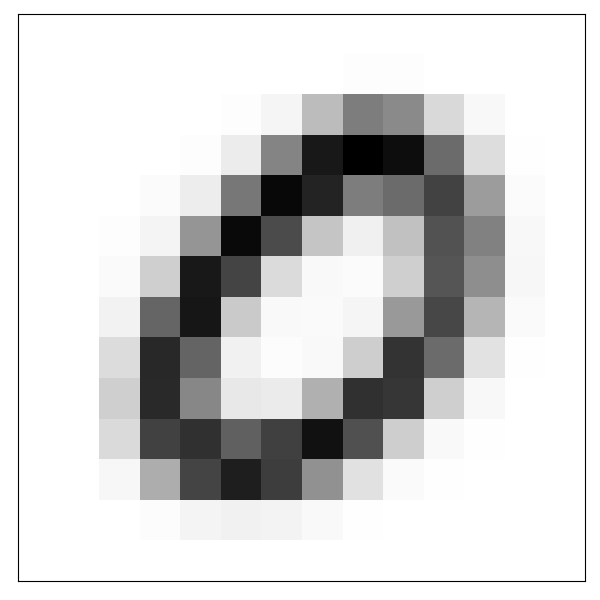}
\includegraphics[trim=0 0 0 0,clip,width=0.25\textwidth]{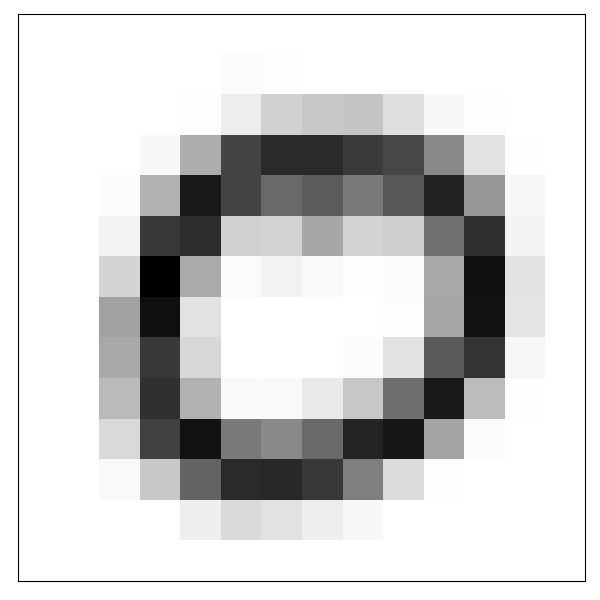}
\includegraphics[trim=0 0 0 0,clip,width=0.25\textwidth]{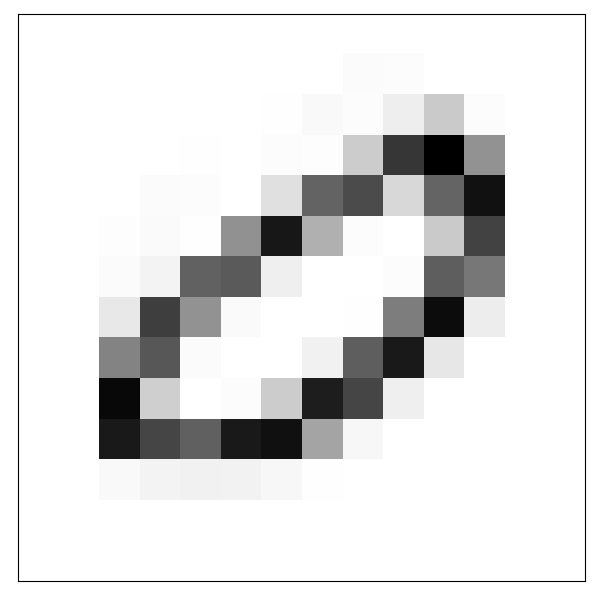}
\includegraphics[trim=0 0 0 0,clip,width=0.25\textwidth]{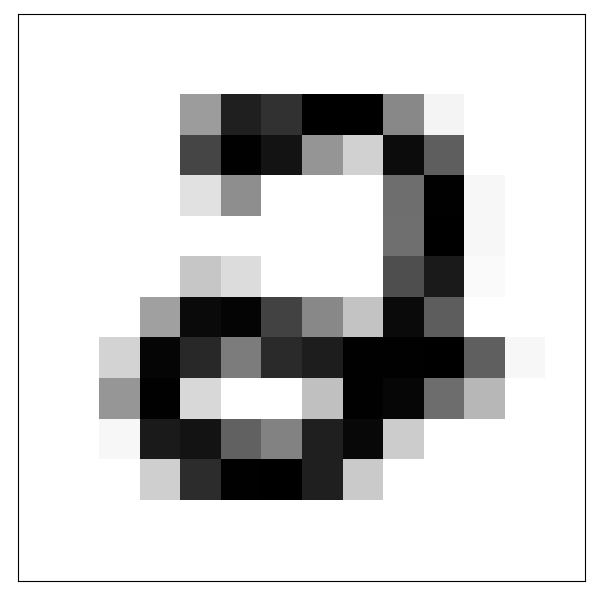}
\includegraphics[trim=0 0 0 0,clip,width=0.25\textwidth]{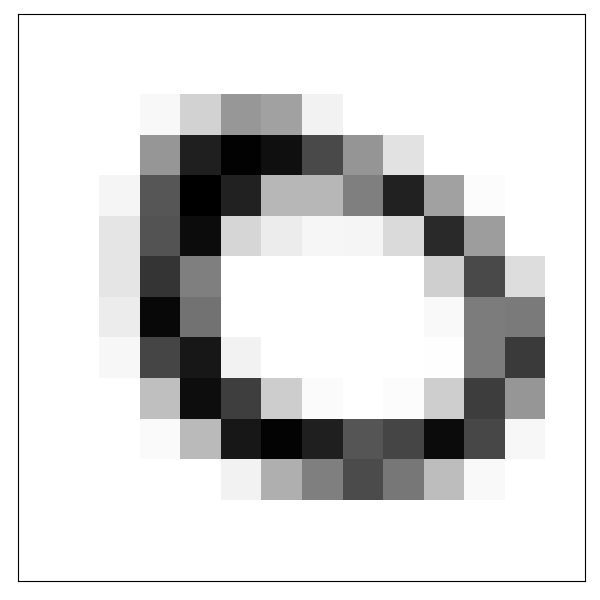}
\includegraphics[trim=0 0 0 0,clip,width=0.25\textwidth]{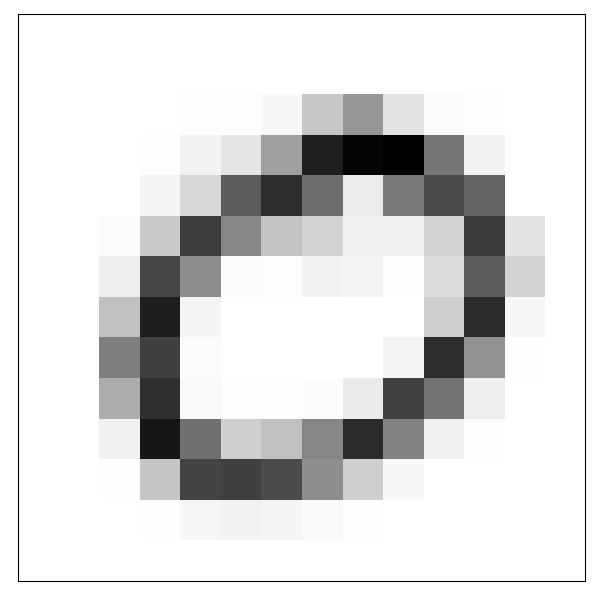}
\includegraphics[trim=0 0 0 0,clip,width=0.25\textwidth]{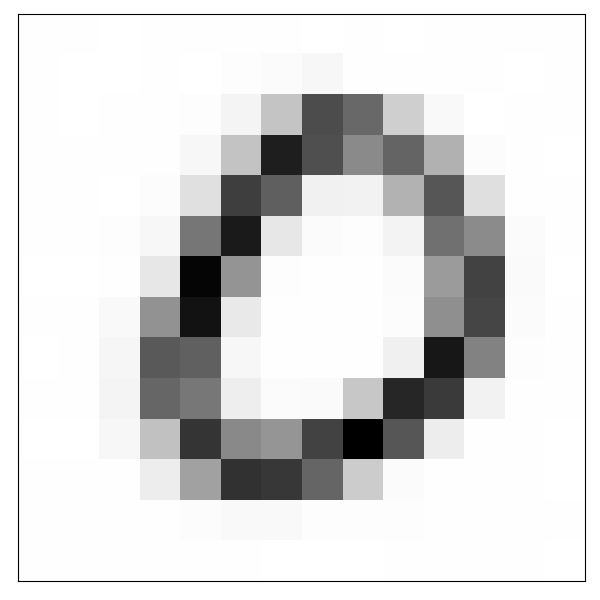}
\includegraphics[trim=0 0 0 0,clip,width=0.25\textwidth]{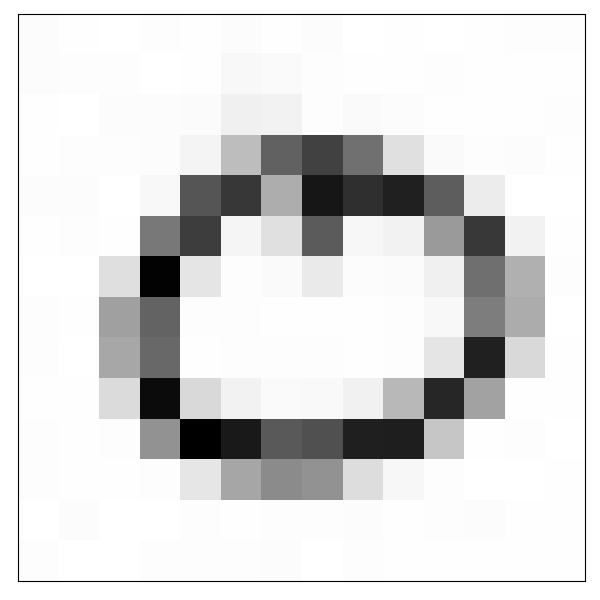}
\includegraphics[trim=0 0 0 0,clip,width=0.25\textwidth]{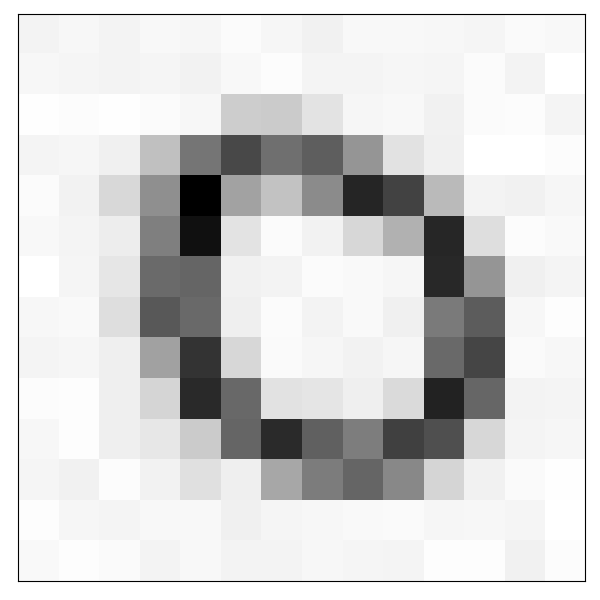}
\includegraphics[trim=0 0 0 0,clip,width=0.25\textwidth]{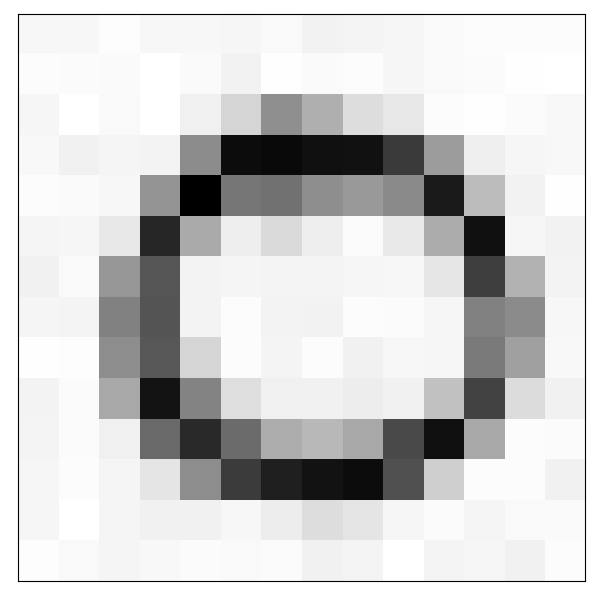}
\caption{$l=2$ (parent node: 0).}
\end{subfigure}
\begin{subfigure}[b]{0.15\textwidth}
\centering
\includegraphics[trim=0 0 0 0,clip,width=0.25\textwidth]{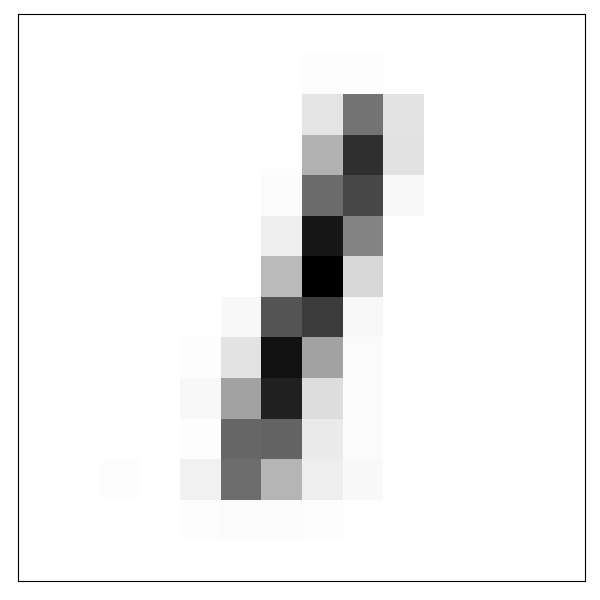}
\includegraphics[trim=0 0 0 0,clip,width=0.25\textwidth]{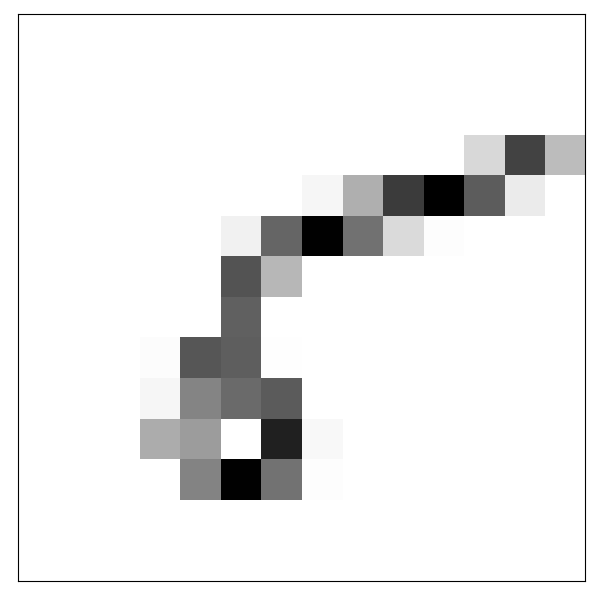}
\includegraphics[trim=0 0 0 0,clip,width=0.25\textwidth]{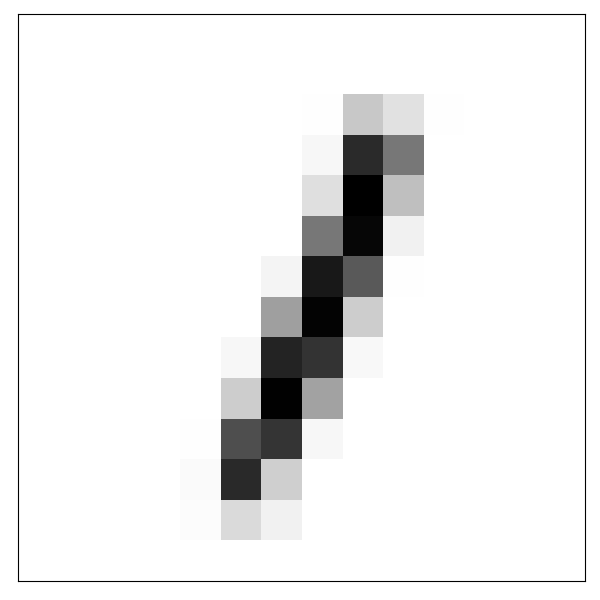}
\includegraphics[trim=0 0 0 0,clip,width=0.25\textwidth]{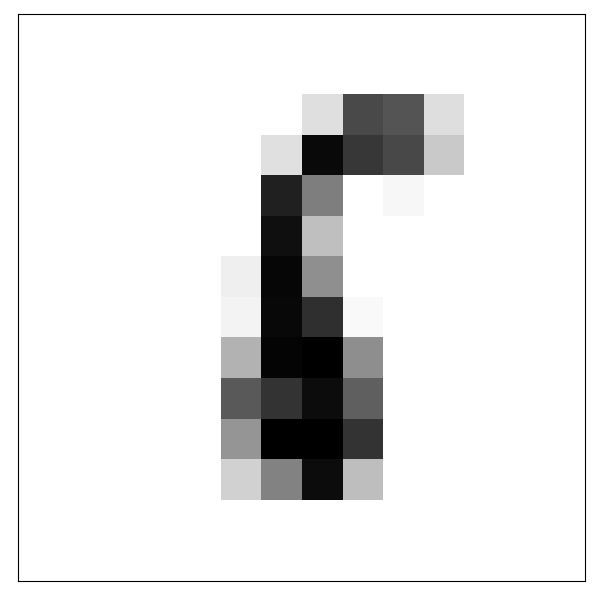}
\includegraphics[trim=0 0 0 0,clip,width=0.25\textwidth]{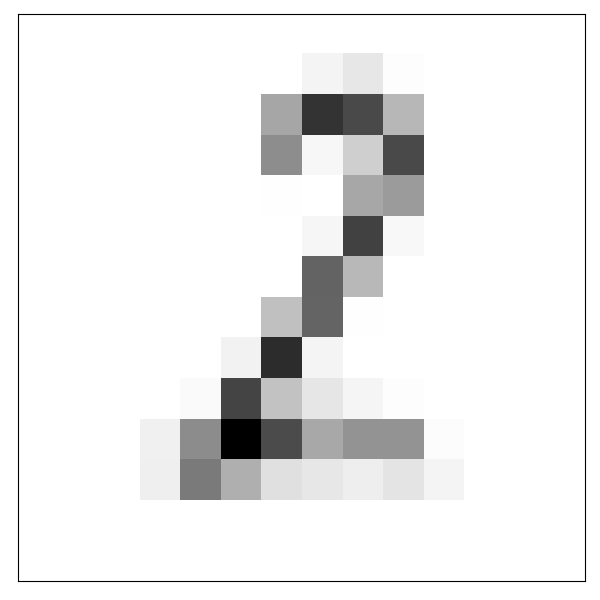}
\includegraphics[trim=0 0 0 0,clip,width=0.25\textwidth]{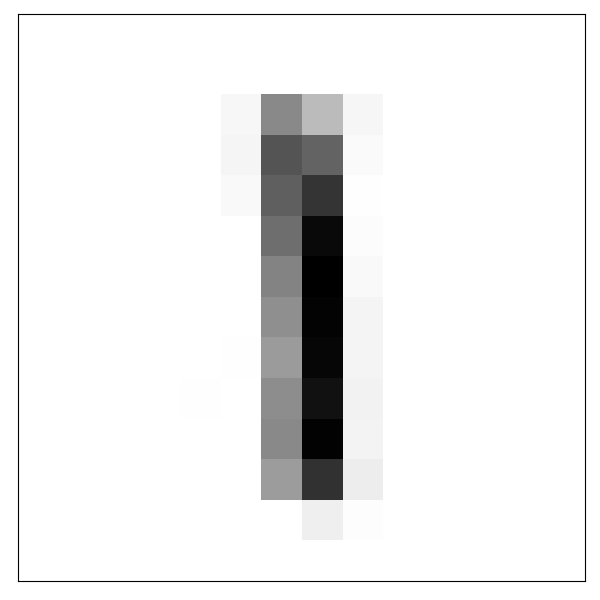}
\includegraphics[trim=0 0 0 0,clip,width=0.25\textwidth]{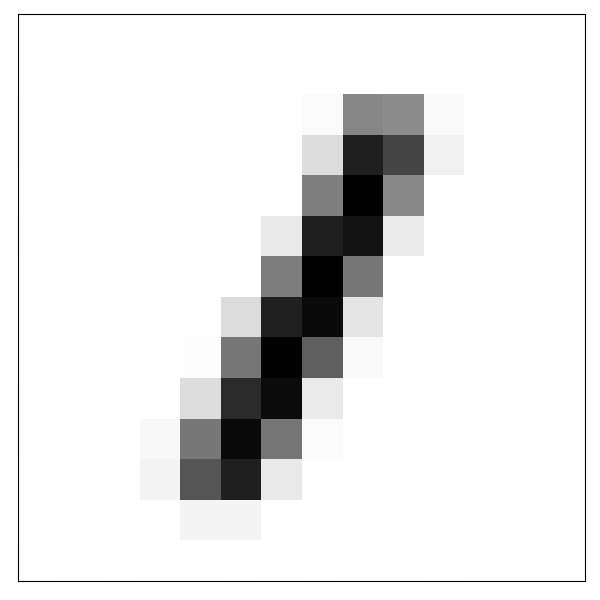}
\includegraphics[trim=0 0 0 0,clip,width=0.25\textwidth]{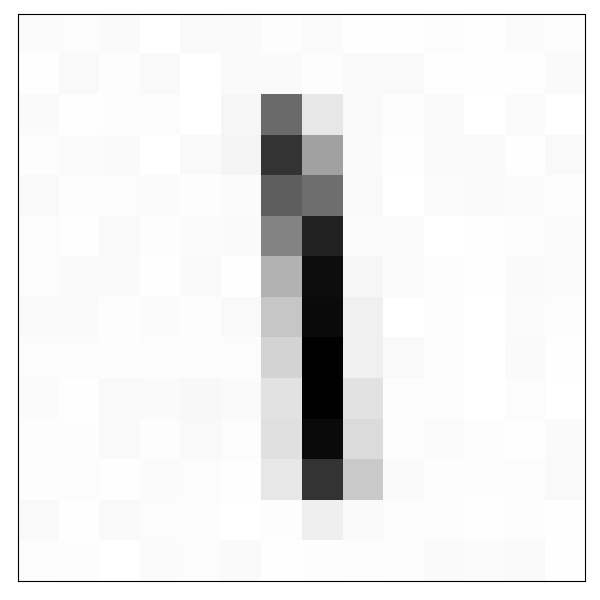}
\includegraphics[trim=0 0 0 0,clip,width=0.25\textwidth]{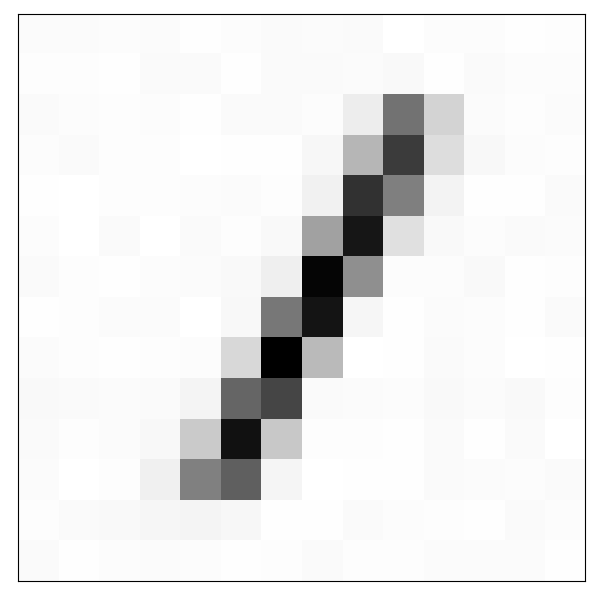}
\includegraphics[trim=0 0 0 0,clip,width=0.25\textwidth]{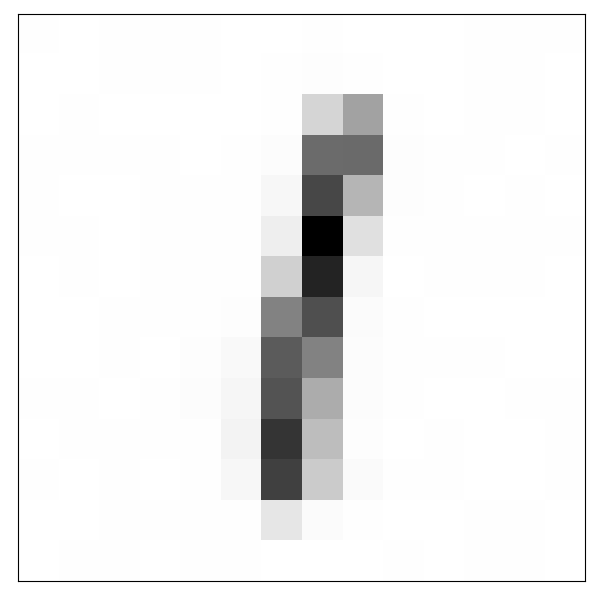}
\caption{$l=2$ (parent node: 1).}
\end{subfigure}
\begin{subfigure}[b]{0.15\textwidth}
\centering
\includegraphics[trim=0 0 0 0,clip,width=0.25\textwidth]{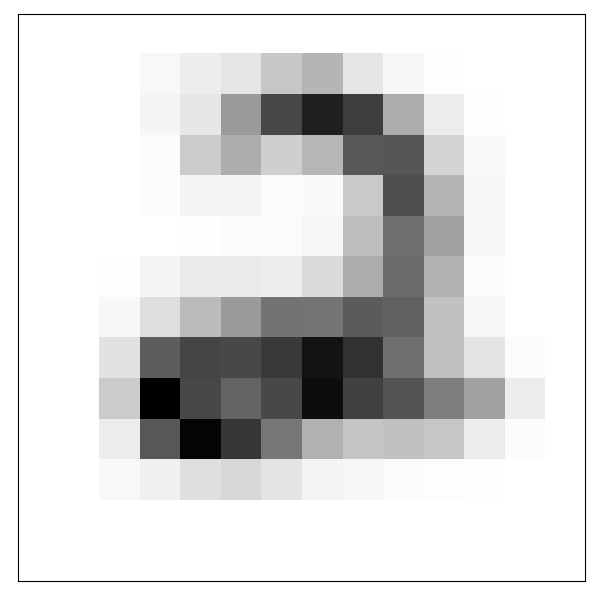}
\includegraphics[trim=0 0 0 0,clip,width=0.25\textwidth]{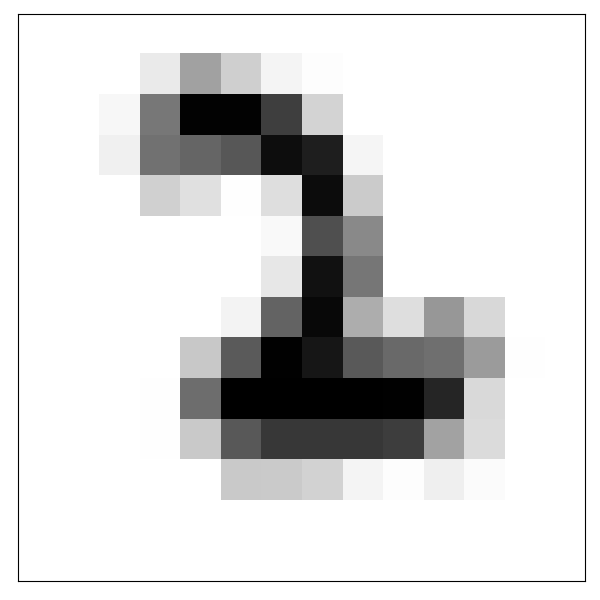}
\includegraphics[trim=0 0 0 0,clip,width=0.25\textwidth]{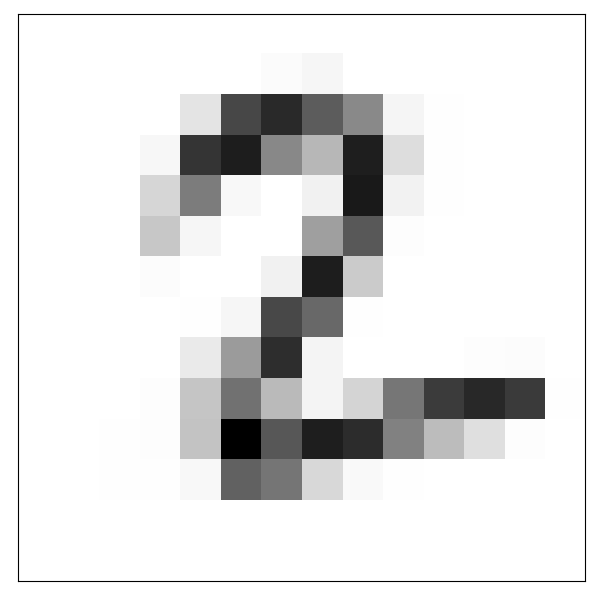}
\includegraphics[trim=0 0 0 0,clip,width=0.25\textwidth]{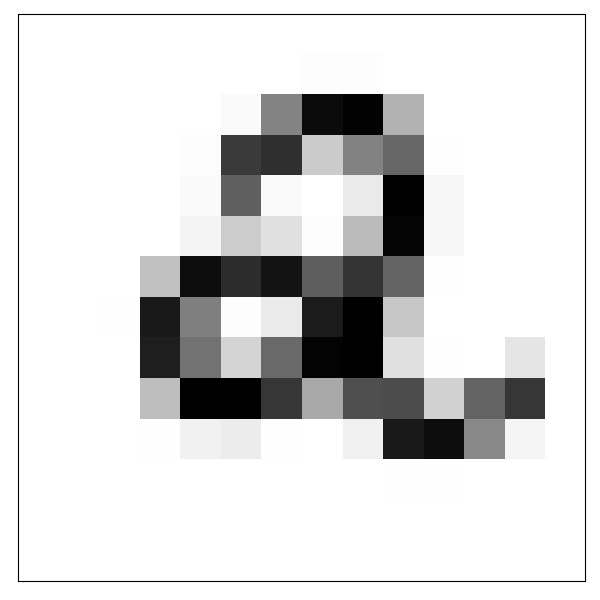}
\includegraphics[trim=0 0 0 0,clip,width=0.25\textwidth]{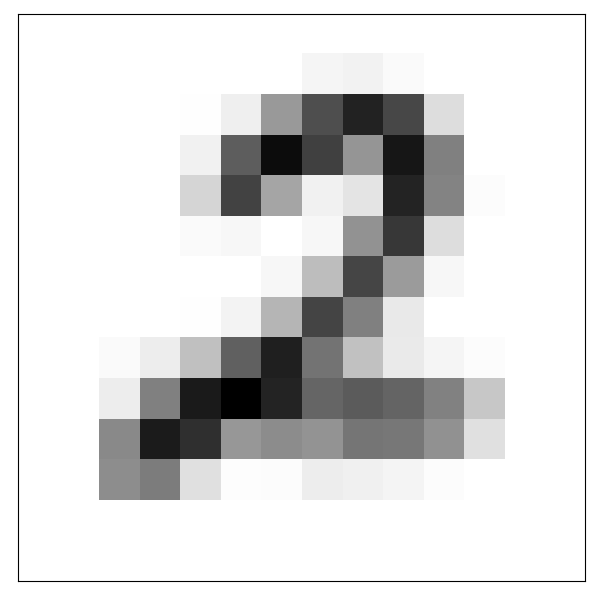}
\includegraphics[trim=0 0 0 0,clip,width=0.25\textwidth]{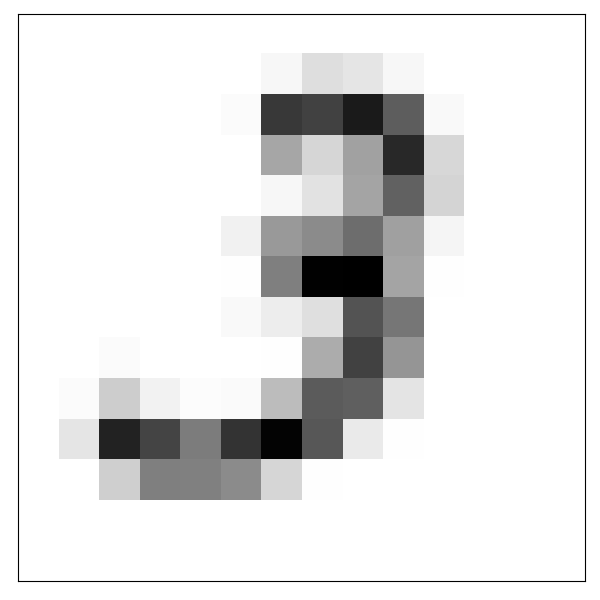}
\includegraphics[trim=0 0 0 0,clip,width=0.25\textwidth]{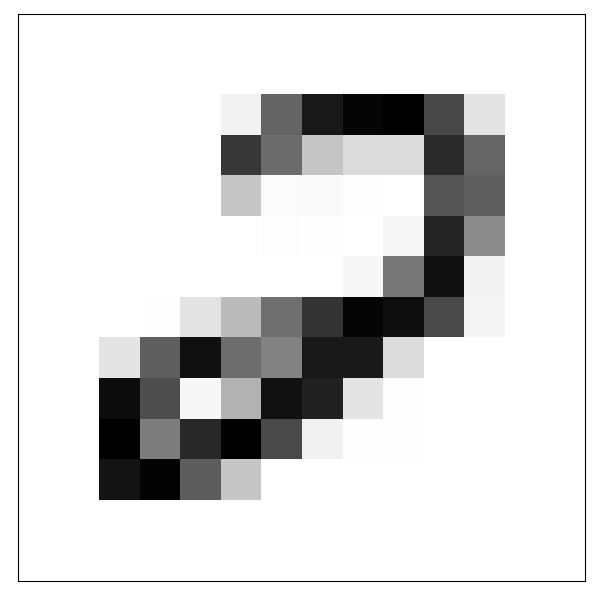}
\includegraphics[trim=0 0 0 0,clip,width=0.25\textwidth]{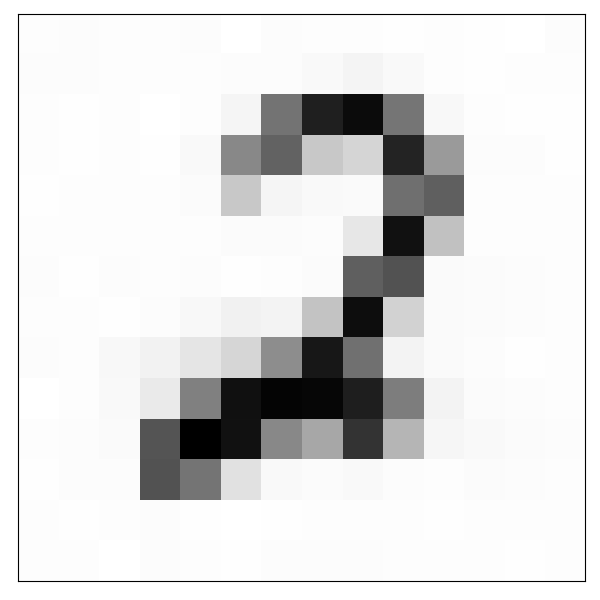}
\includegraphics[trim=0 0 0 0,clip,width=0.25\textwidth]{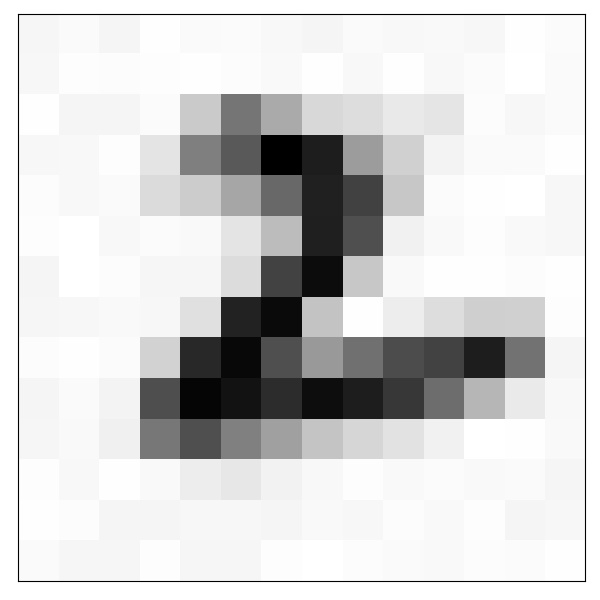}
\includegraphics[trim=0 0 0 0,clip,width=0.25\textwidth]{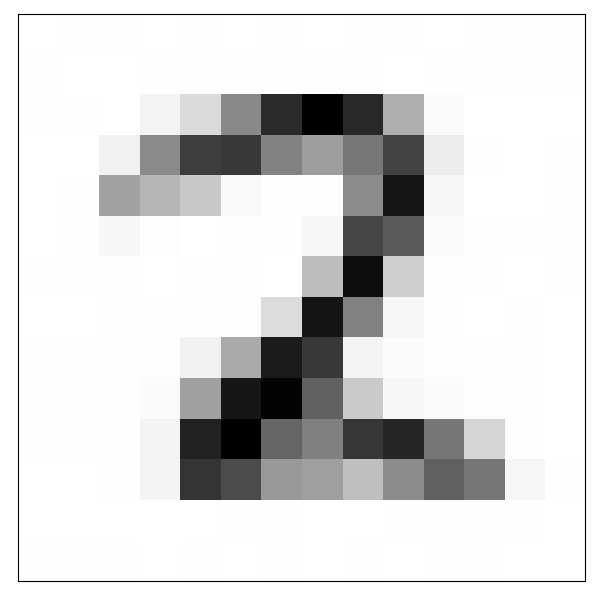}
\caption{$l=2$ (parent node: 2).}
\end{subfigure}
\begin{subfigure}[b]{0.15\textwidth}
\centering
\includegraphics[trim=0 0 0 0,clip,width=0.25\textwidth]{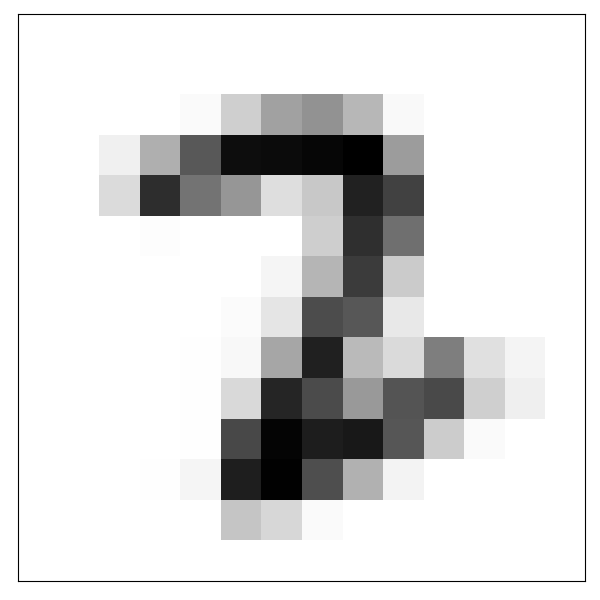}
\includegraphics[trim=0 0 0 0,clip,width=0.25\textwidth]{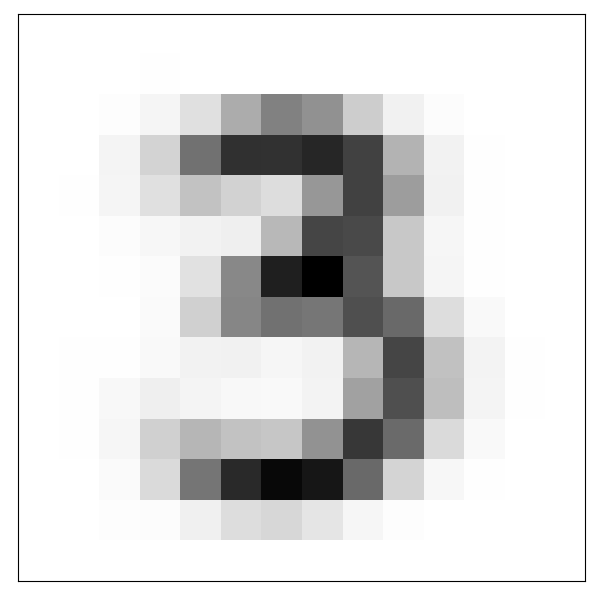}
\includegraphics[trim=0 0 0 0,clip,width=0.25\textwidth]{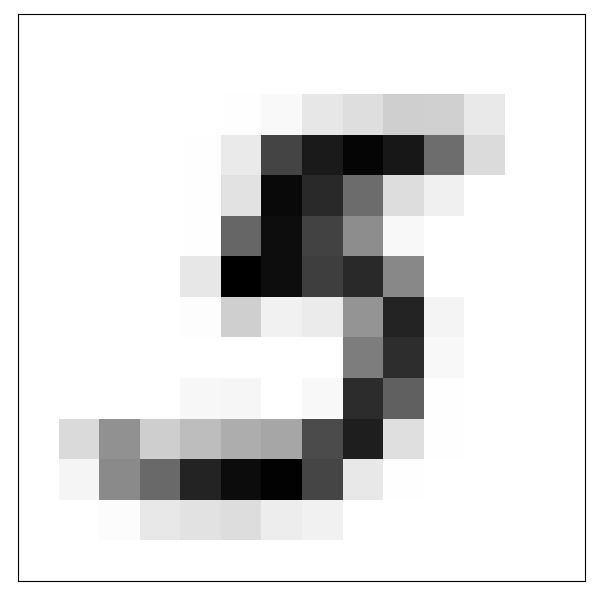}
\includegraphics[trim=0 0 0 0,clip,width=0.25\textwidth]{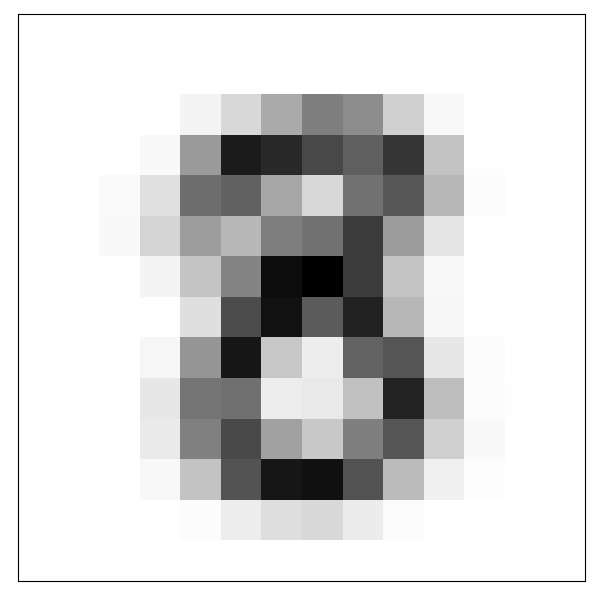}
\includegraphics[trim=0 0 0 0,clip,width=0.25\textwidth]{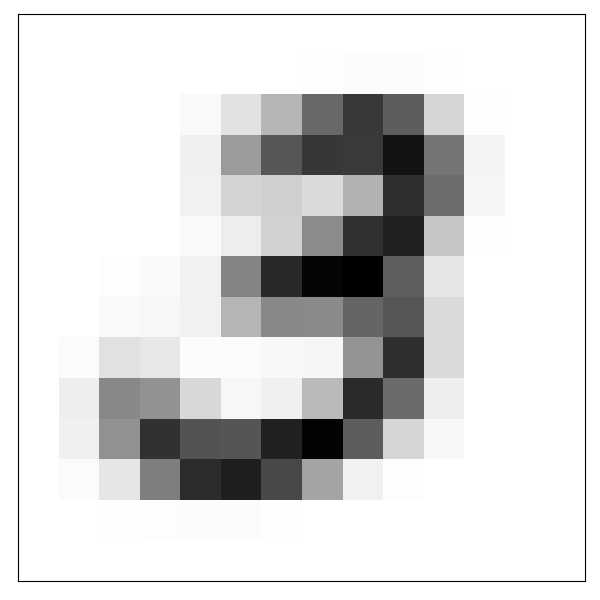}
\includegraphics[trim=0 0 0 0,clip,width=0.25\textwidth]{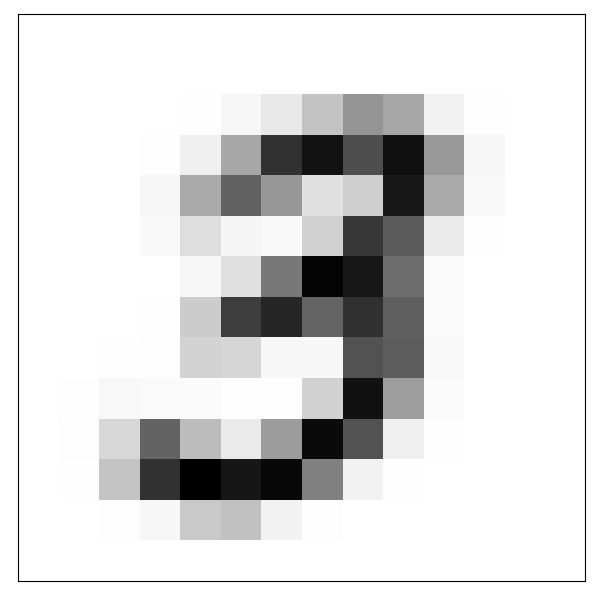}
\includegraphics[trim=0 0 0 0,clip,width=0.25\textwidth]{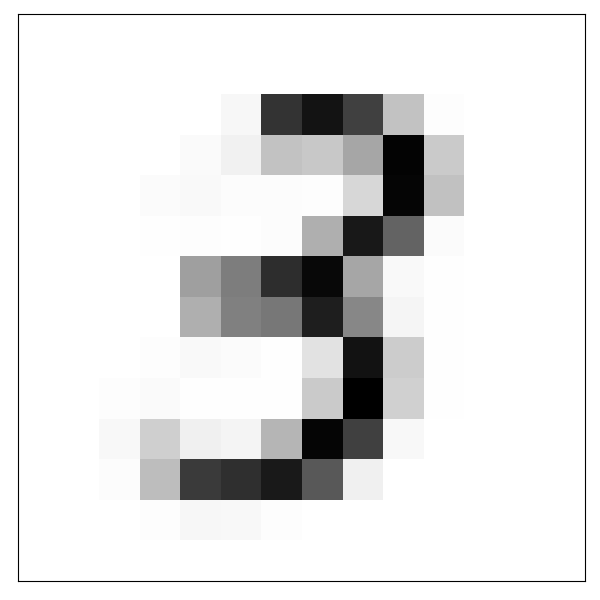}
\includegraphics[trim=0 0 0 0,clip,width=0.25\textwidth]{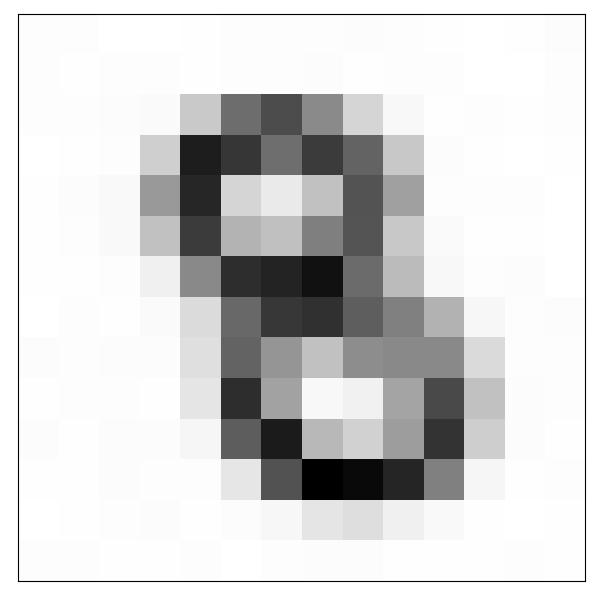}
\includegraphics[trim=0 0 0 0,clip,width=0.25\textwidth]{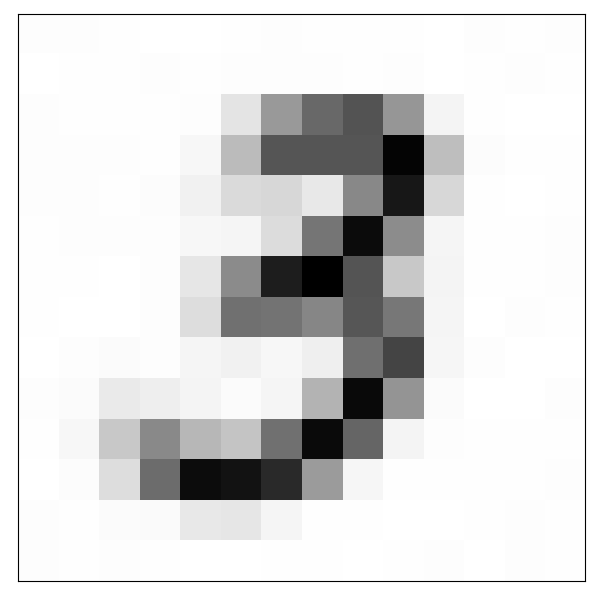}
\includegraphics[trim=0 0 0 0,clip,width=0.25\textwidth]{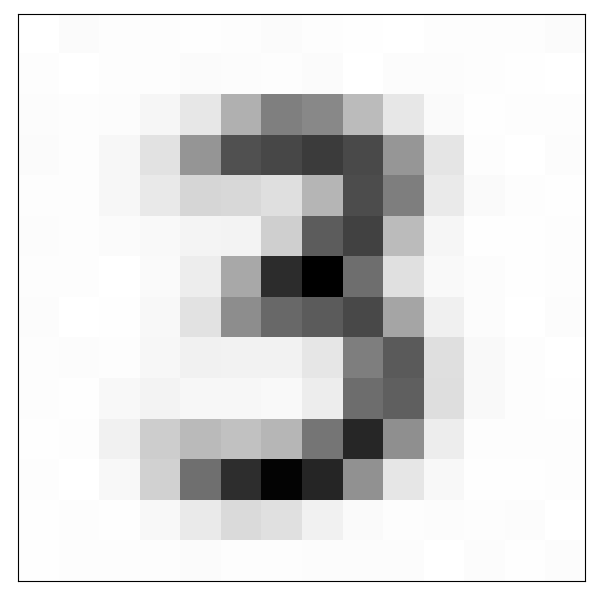}
\caption{$l=2$ (parent node: 3).}
\end{subfigure}
\begin{subfigure}[b]{0.15\textwidth}
\centering
\includegraphics[trim=0 0 0 0,clip,width=0.25\textwidth]{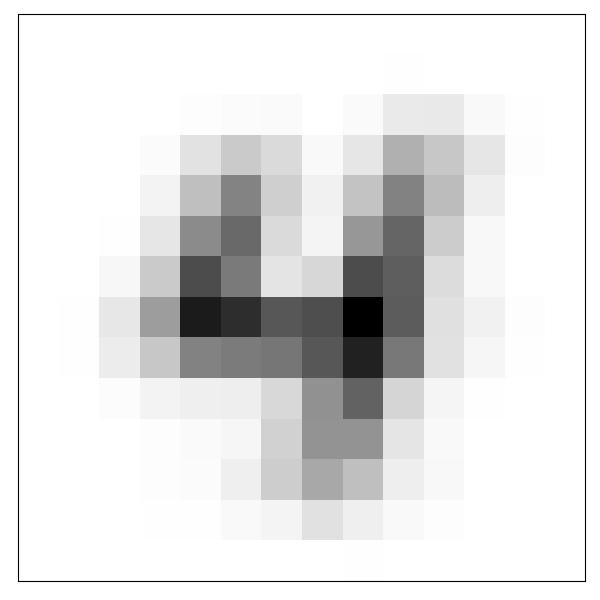}
\includegraphics[trim=0 0 0 0,clip,width=0.25\textwidth]{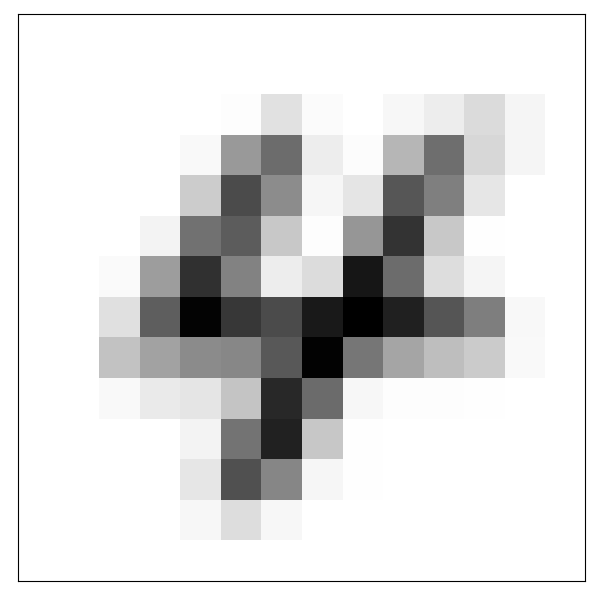}
\includegraphics[trim=0 0 0 0,clip,width=0.25\textwidth]{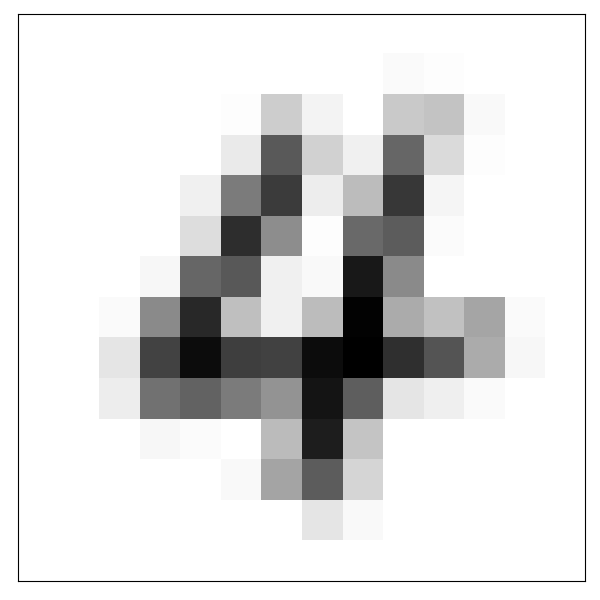}
\includegraphics[trim=0 0 0 0,clip,width=0.25\textwidth]{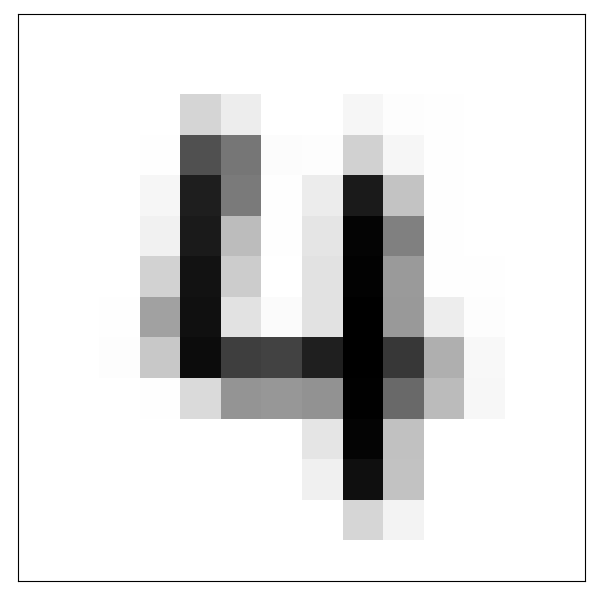}
\includegraphics[trim=0 0 0 0,clip,width=0.25\textwidth]{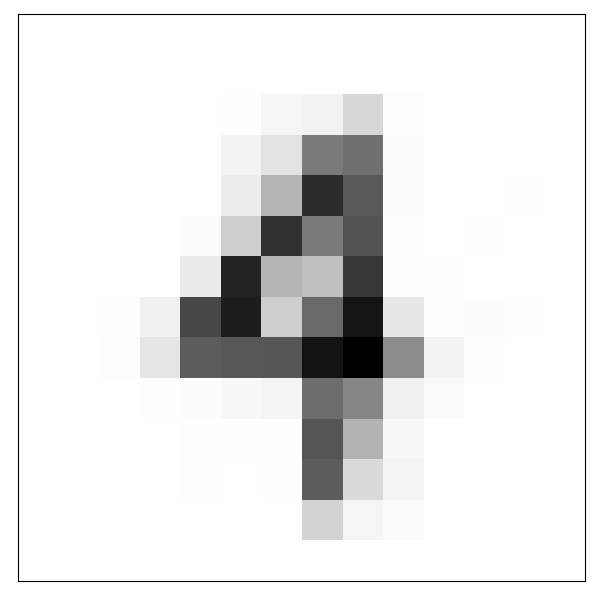}
\includegraphics[trim=0 0 0 0,clip,width=0.25\textwidth]{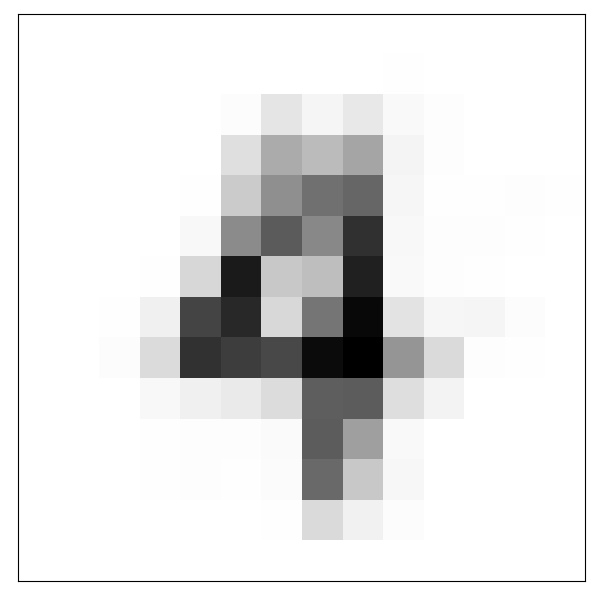}
\includegraphics[trim=0 0 0 0,clip,width=0.25\textwidth]{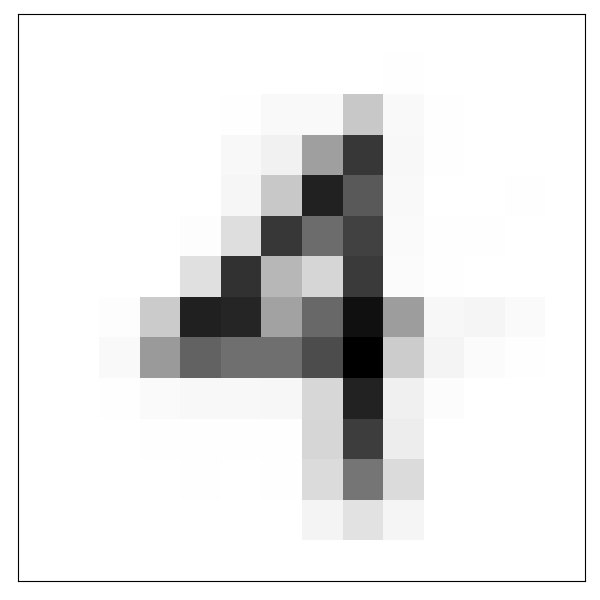}
\includegraphics[trim=0 0 0 0,clip,width=0.25\textwidth]{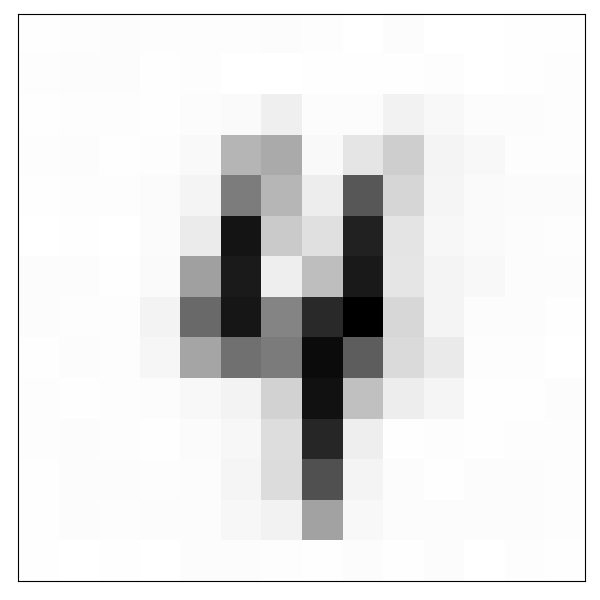}
\includegraphics[trim=0 0 0 0,clip,width=0.25\textwidth]{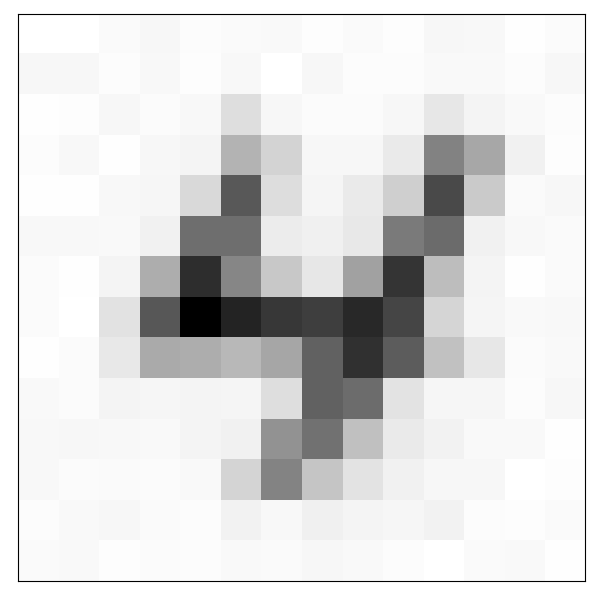}
\includegraphics[trim=0 0 0 0,clip,width=0.25\textwidth]{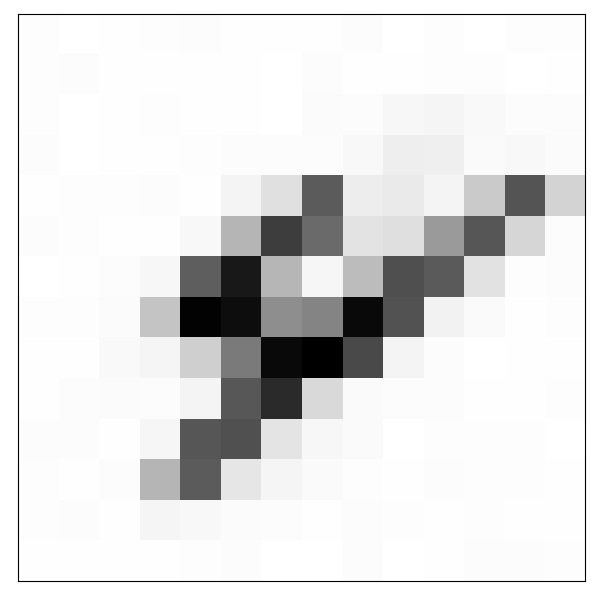}
\caption{$l=2$ (parent node: 4).}
\end{subfigure}
\begin{subfigure}[b]{0.15\textwidth}
\centering
\includegraphics[trim=0 0 0 0,clip,width=0.25\textwidth]{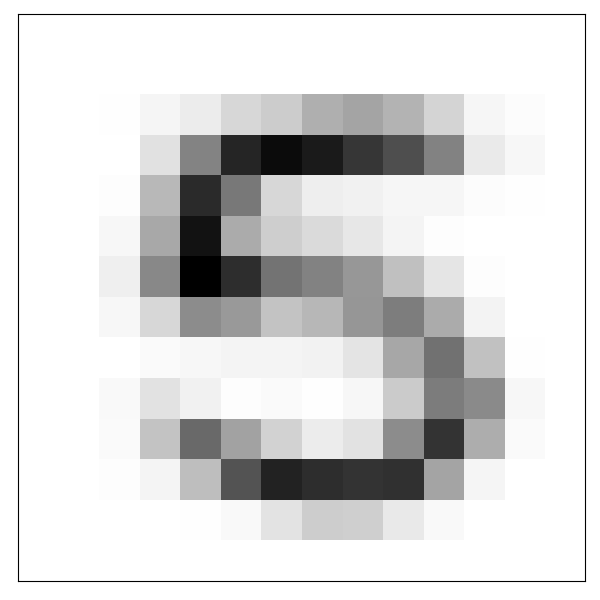}
\includegraphics[trim=0 0 0 0,clip,width=0.25\textwidth]{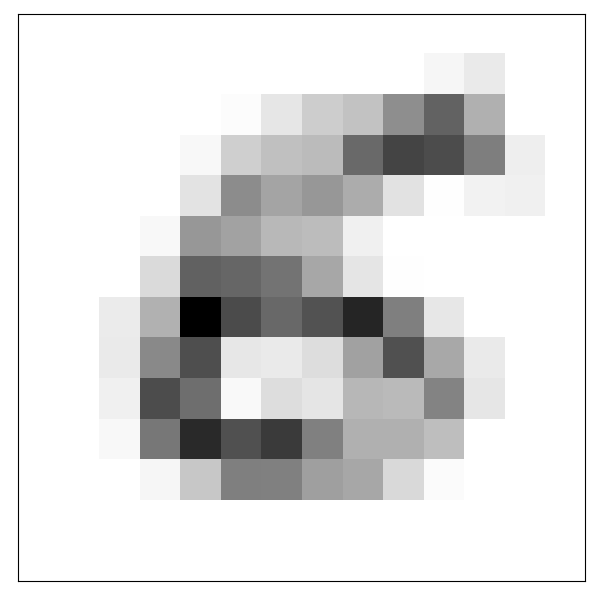}
\includegraphics[trim=0 0 0 0,clip,width=0.25\textwidth]{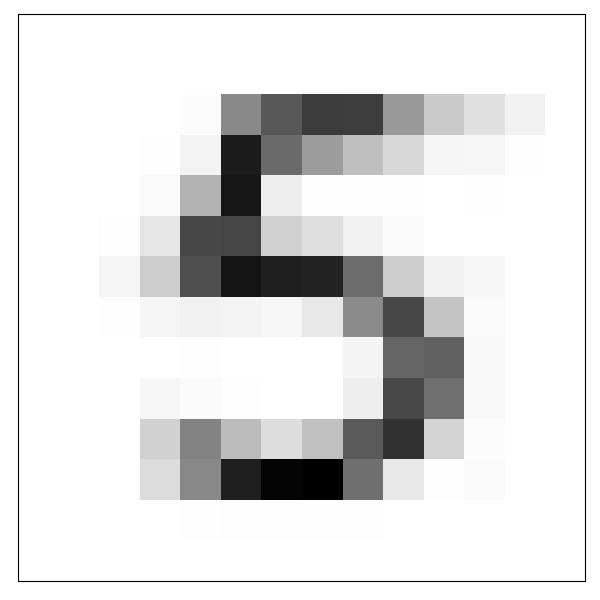}
\includegraphics[trim=0 0 0 0,clip,width=0.25\textwidth]{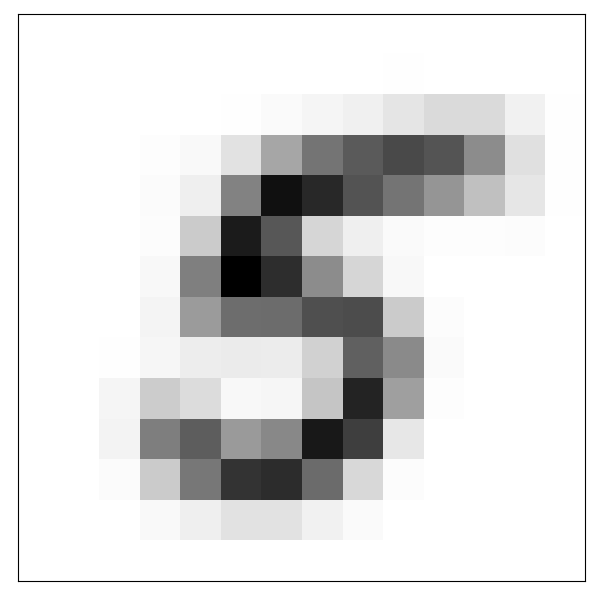}
\includegraphics[trim=0 0 0 0,clip,width=0.25\textwidth]{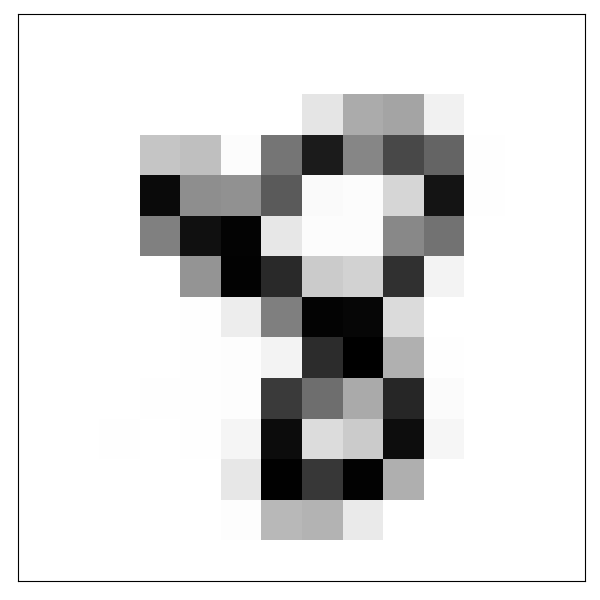}
\includegraphics[trim=0 0 0 0,clip,width=0.25\textwidth]{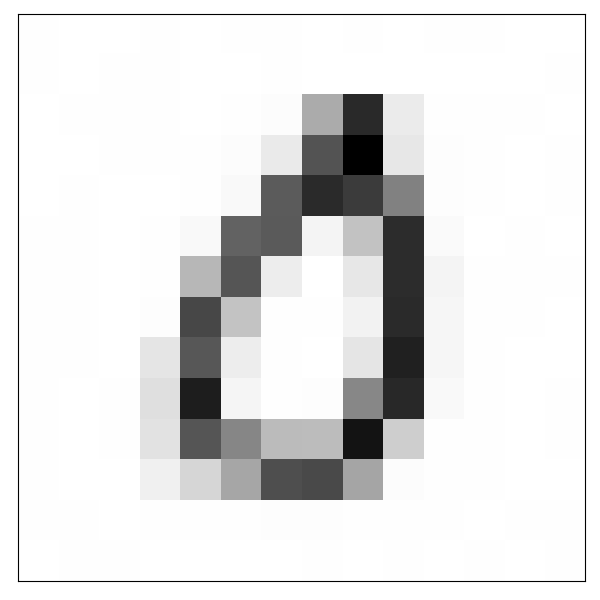}
\includegraphics[trim=0 0 0 0,clip,width=0.25\textwidth]{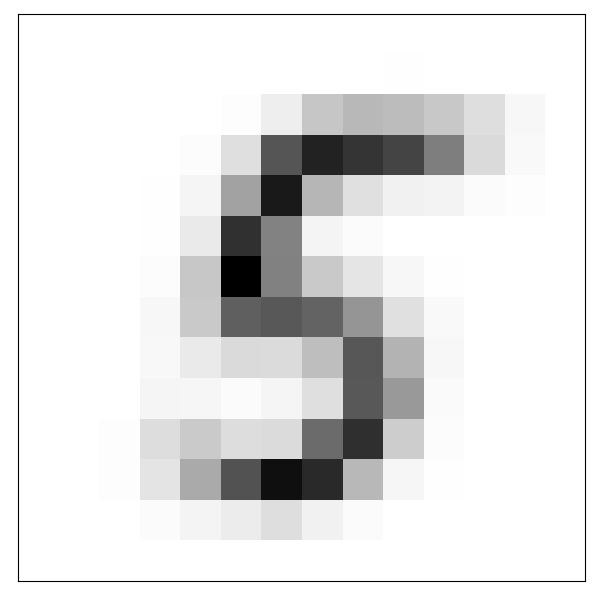}
\includegraphics[trim=0 0 0 0,clip,width=0.25\textwidth]{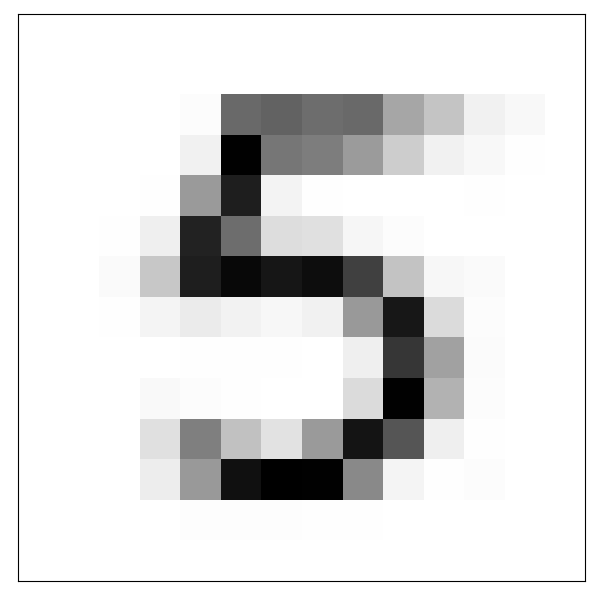}
\includegraphics[trim=0 0 0 0,clip,width=0.25\textwidth]{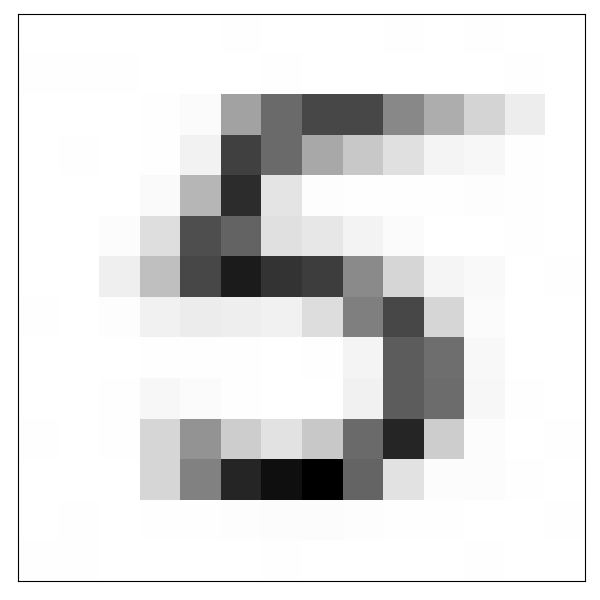}
\includegraphics[trim=0 0 0 0,clip,width=0.25\textwidth]{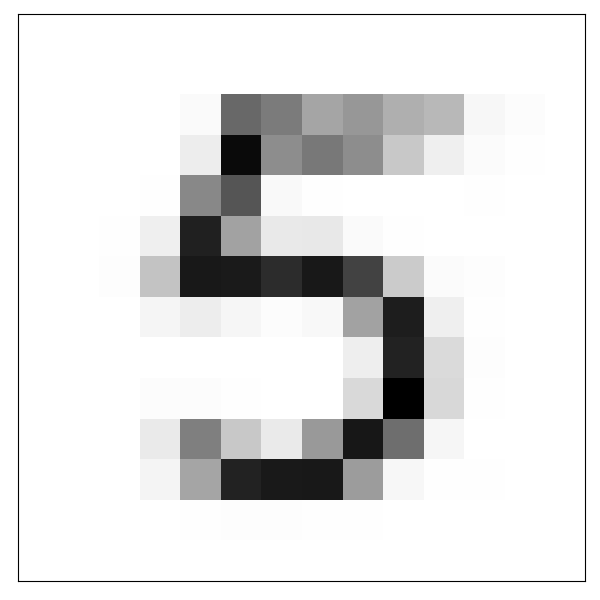}
\caption{$l=2$ (parent node: 5).}
\end{subfigure}
\begin{subfigure}[b]{0.15\textwidth}
\centering
\includegraphics[trim=0 0 0 0,clip,width=0.25\textwidth]{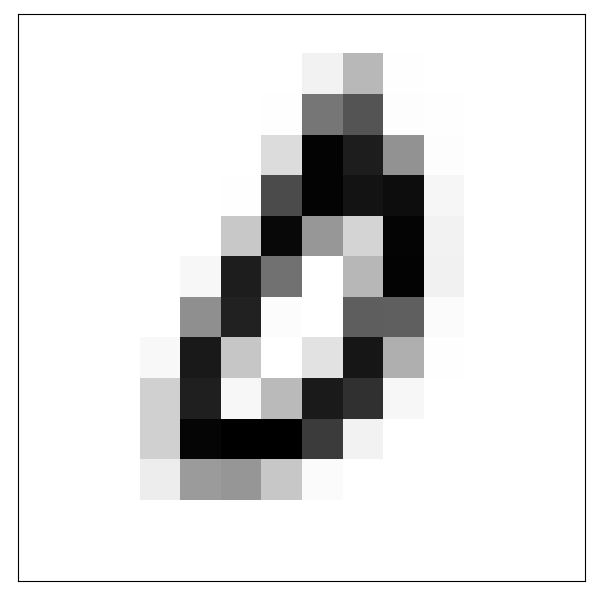}
\includegraphics[trim=0 0 0 0,clip,width=0.25\textwidth]{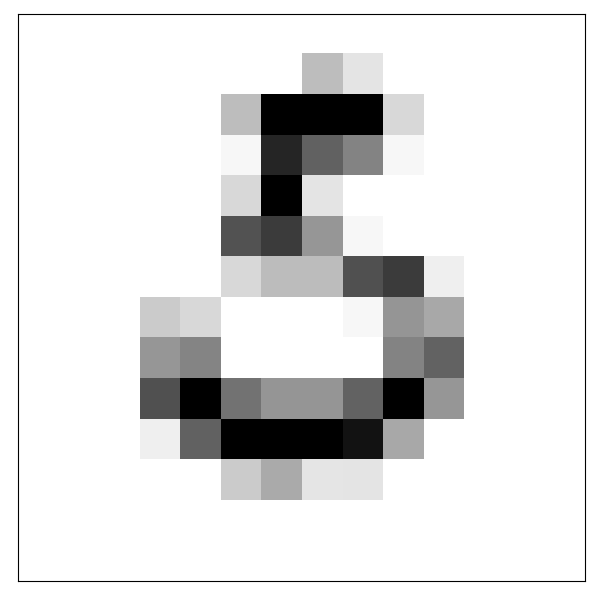}
\includegraphics[trim=0 0 0 0,clip,width=0.25\textwidth]{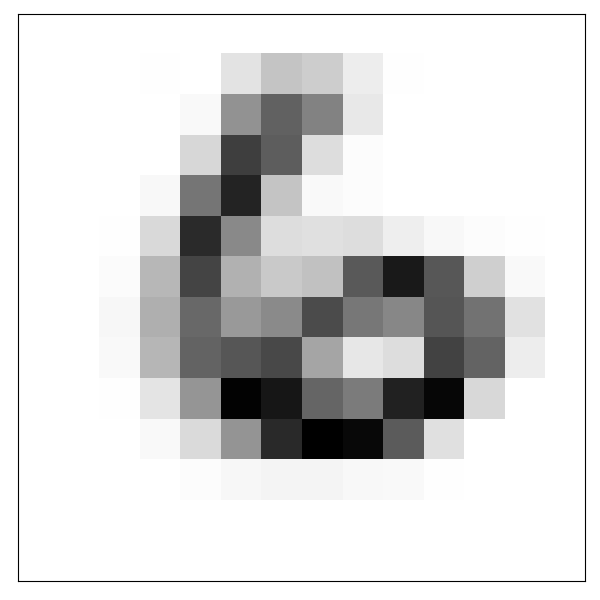}
\includegraphics[trim=0 0 0 0,clip,width=0.25\textwidth]{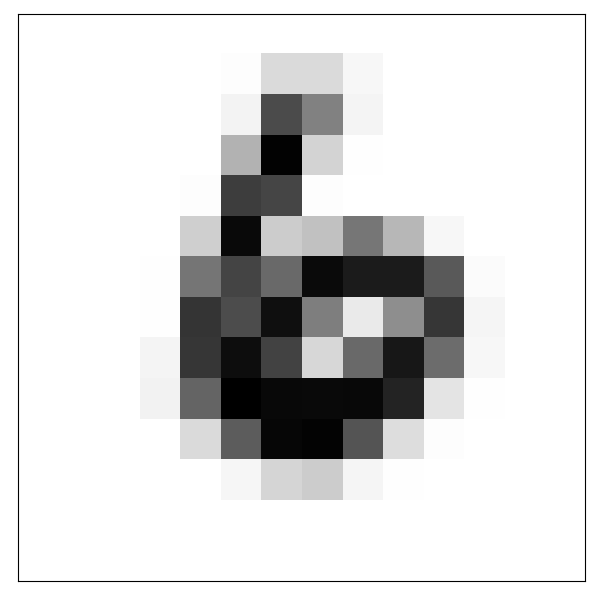}
\includegraphics[trim=0 0 0 0,clip,width=0.25\textwidth]{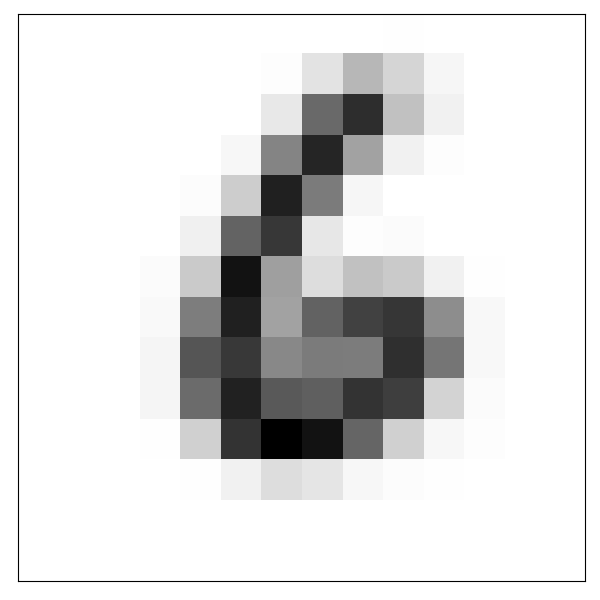}
\includegraphics[trim=0 0 0 0,clip,width=0.25\textwidth]{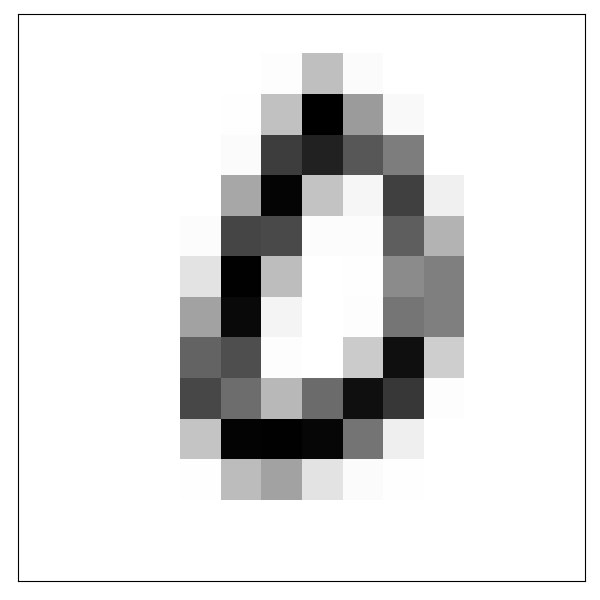}
\includegraphics[trim=0 0 0 0,clip,width=0.25\textwidth]{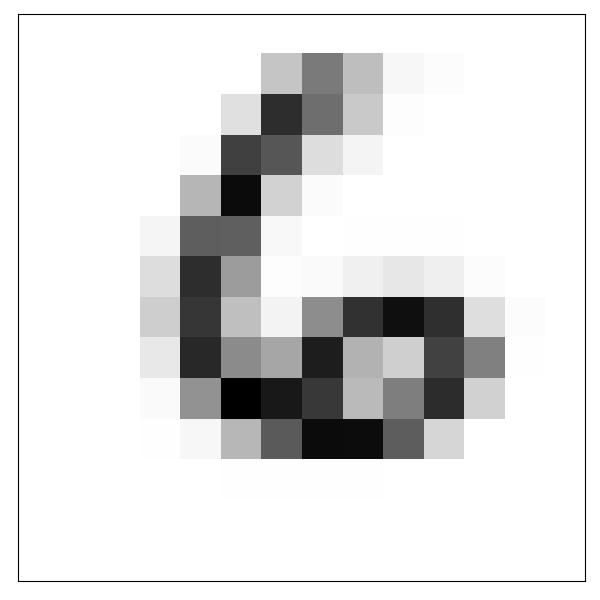}
\includegraphics[trim=0 0 0 0,clip,width=0.25\textwidth]{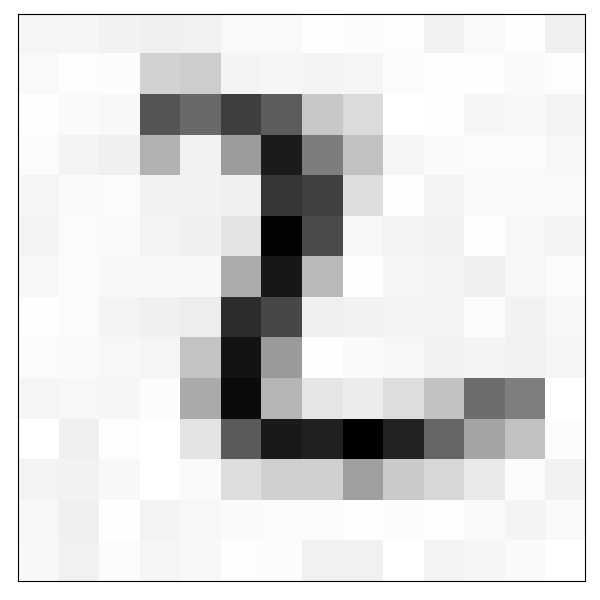}
\includegraphics[trim=0 0 0 0,clip,width=0.25\textwidth]{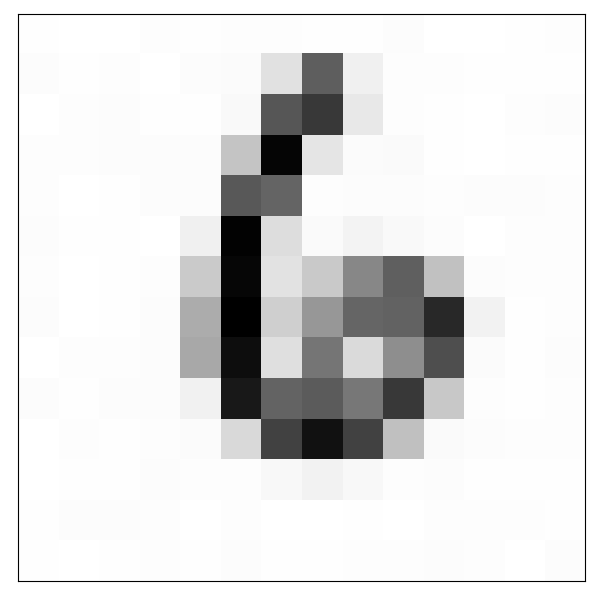}
\includegraphics[trim=0 0 0 0,clip,width=0.25\textwidth]{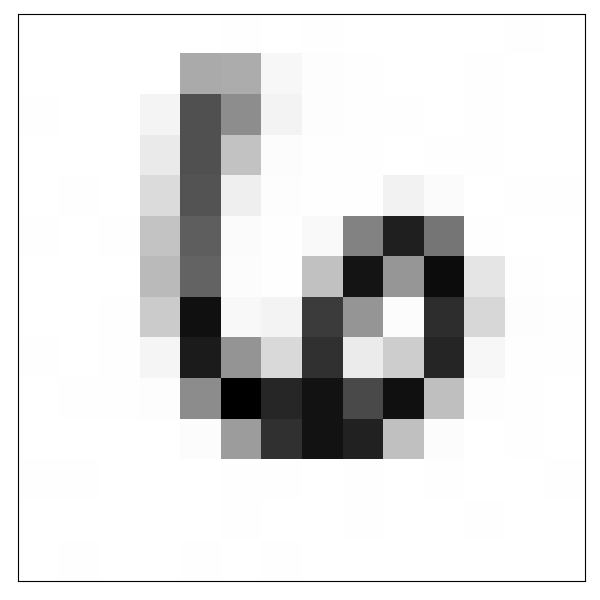}
\caption{$l=2$ (parent node: 6).}
\end{subfigure}
\begin{subfigure}[b]{0.15\textwidth}
\centering
\includegraphics[trim=0 0 0 0,clip,width=0.25\textwidth]{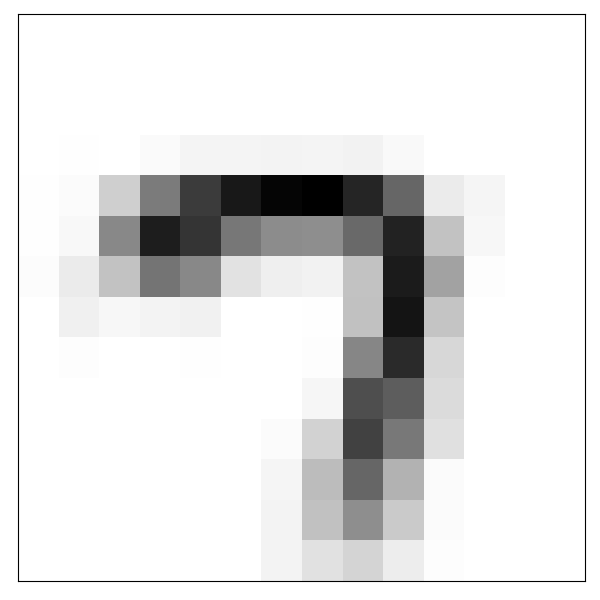}
\includegraphics[trim=0 0 0 0,clip,width=0.25\textwidth]{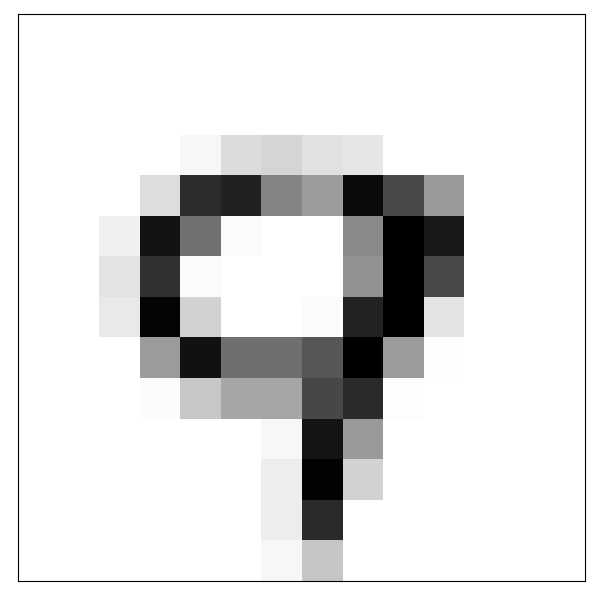}
\includegraphics[trim=0 0 0 0,clip,width=0.25\textwidth]{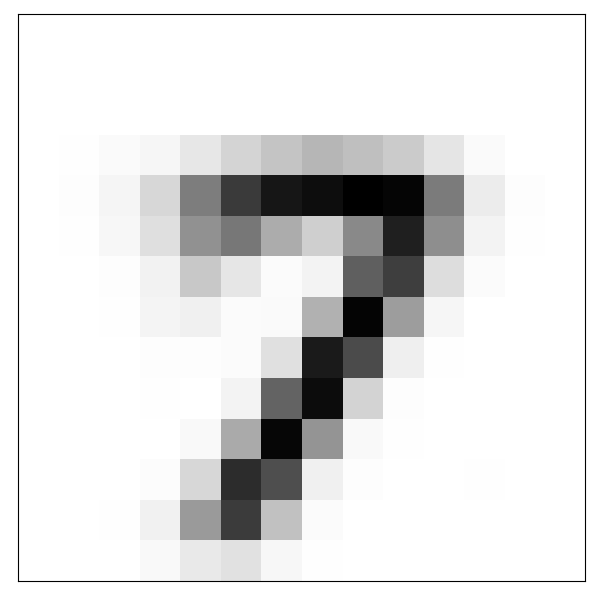}
\includegraphics[trim=0 0 0 0,clip,width=0.25\textwidth]{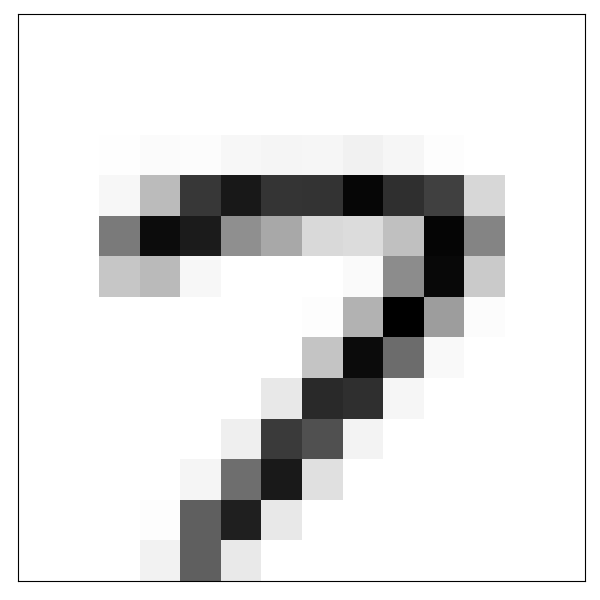}
\includegraphics[trim=0 0 0 0,clip,width=0.25\textwidth]{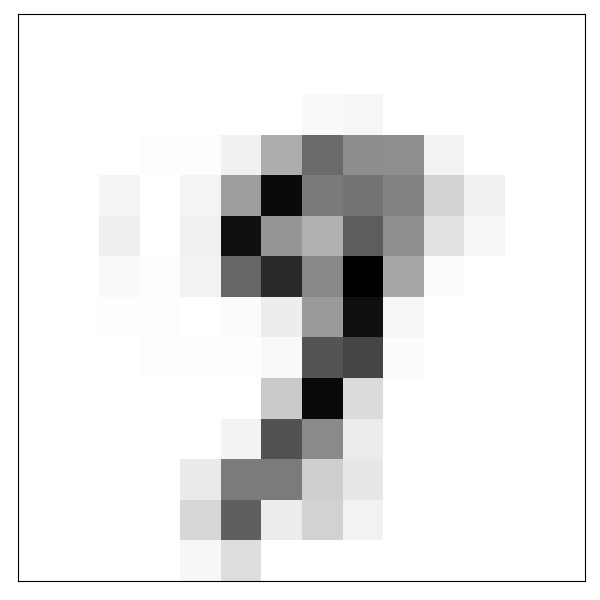}
\includegraphics[trim=0 0 0 0,clip,width=0.25\textwidth]{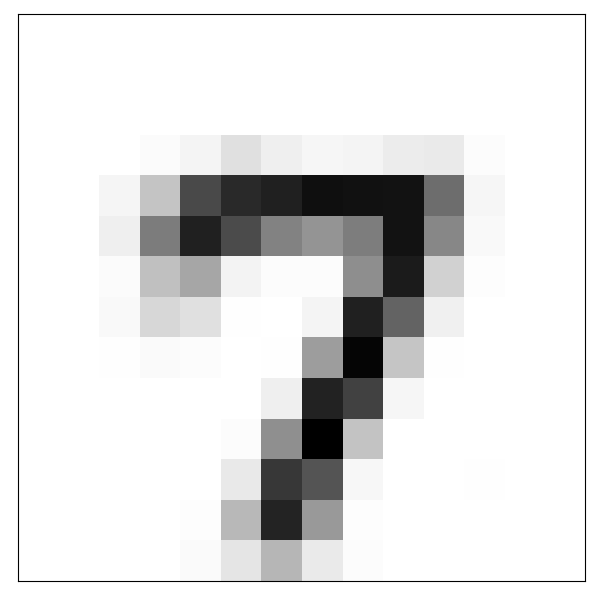}
\includegraphics[trim=0 0 0 0,clip,width=0.25\textwidth]{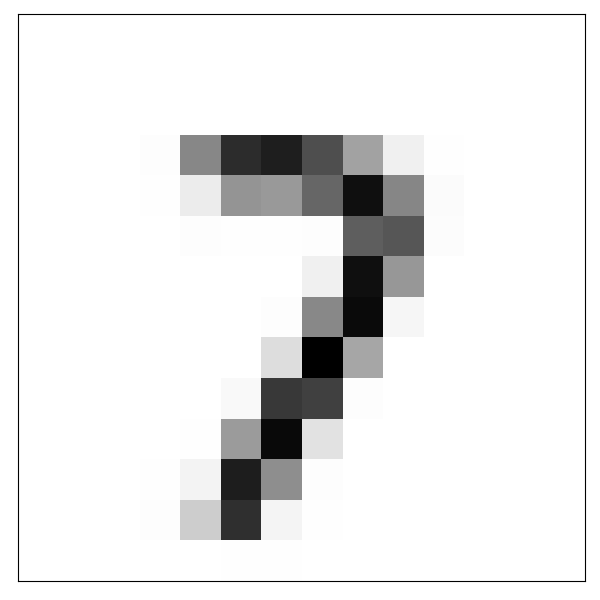}
\includegraphics[trim=0 0 0 0,clip,width=0.25\textwidth]{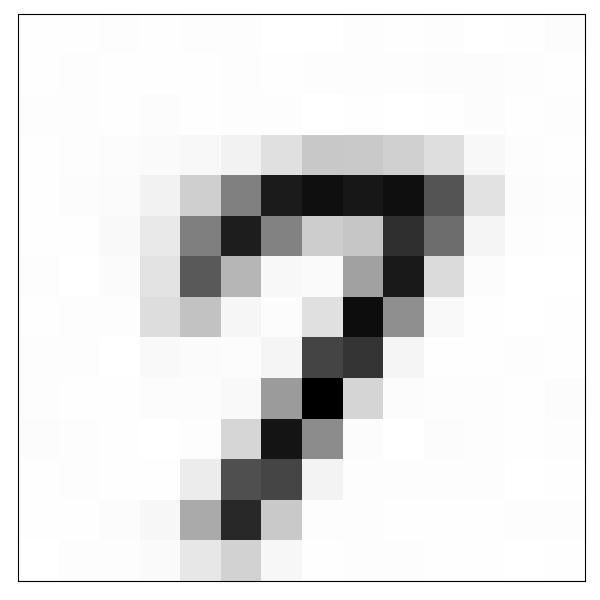}
\includegraphics[trim=0 0 0 0,clip,width=0.25\textwidth]{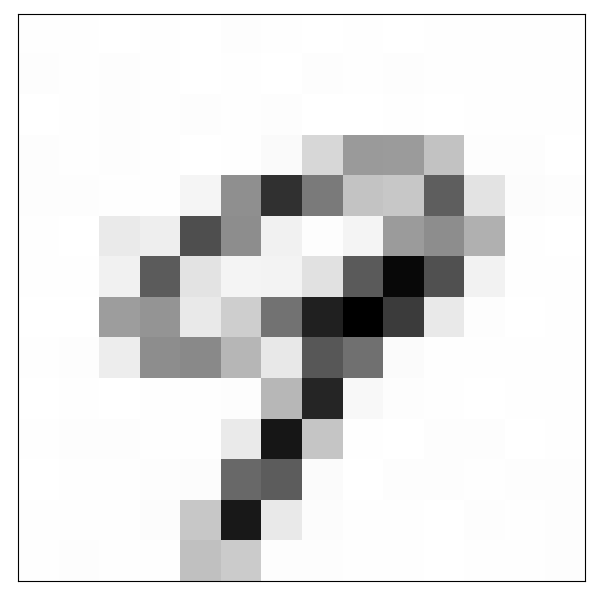}
\includegraphics[trim=0 0 0 0,clip,width=0.25\textwidth]{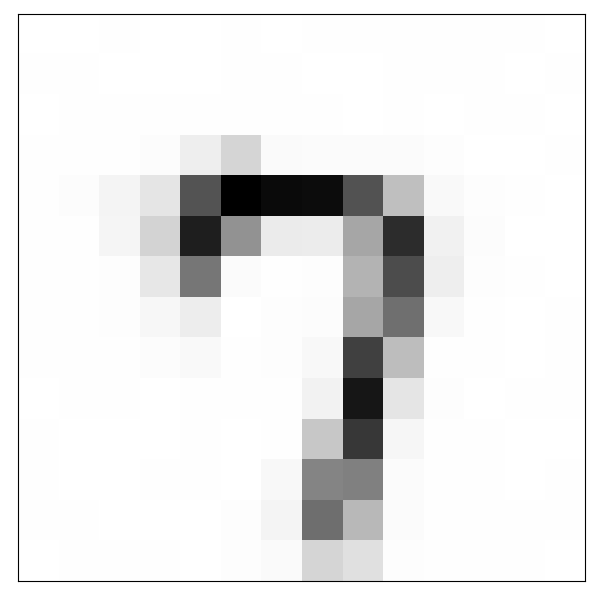}
\caption{$l=2$ (parent node: 7).}
\end{subfigure}
\begin{subfigure}[b]{0.15\textwidth}
\centering
\includegraphics[trim=0 0 0 0,clip,width=0.25\textwidth]{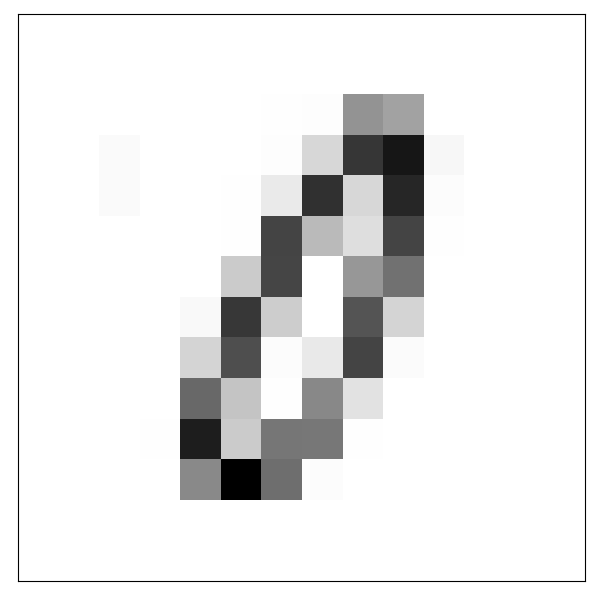}
\includegraphics[trim=0 0 0 0,clip,width=0.25\textwidth]{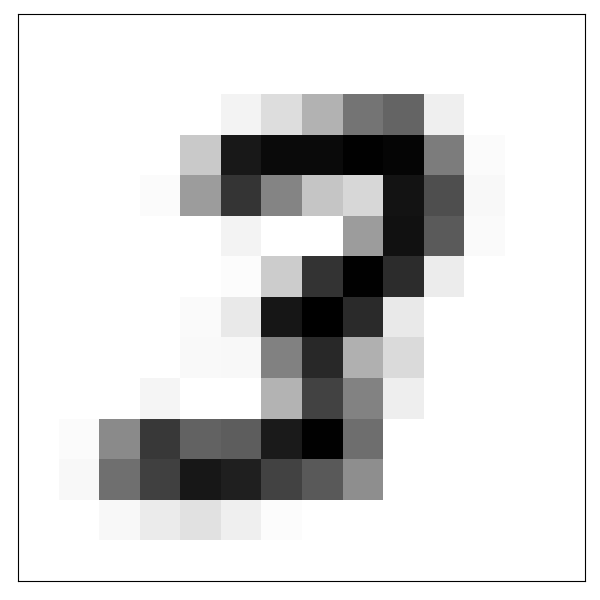}
\includegraphics[trim=0 0 0 0,clip,width=0.25\textwidth]{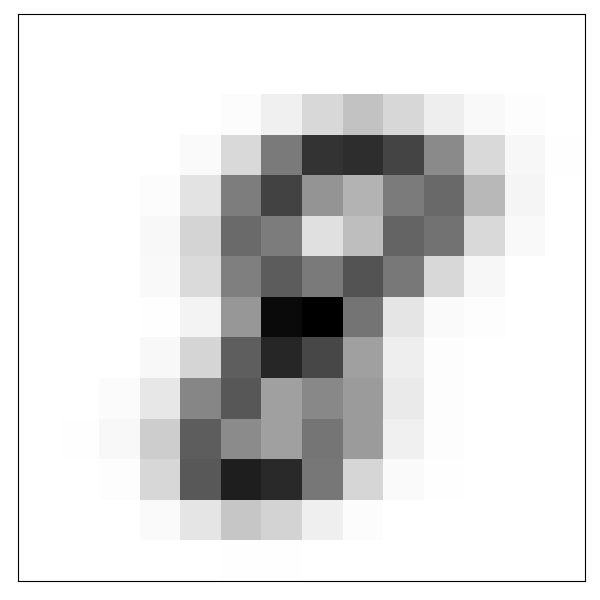}
\includegraphics[trim=0 0 0 0,clip,width=0.25\textwidth]{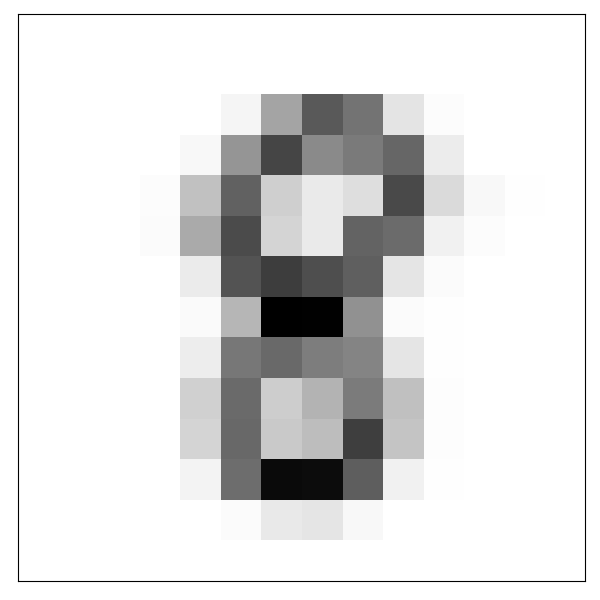}
\includegraphics[trim=0 0 0 0,clip,width=0.25\textwidth]{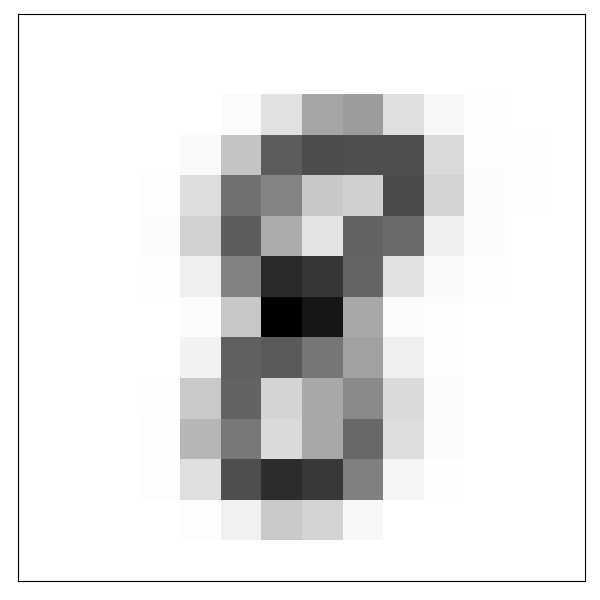}
\includegraphics[trim=0 0 0 0,clip,width=0.25\textwidth]{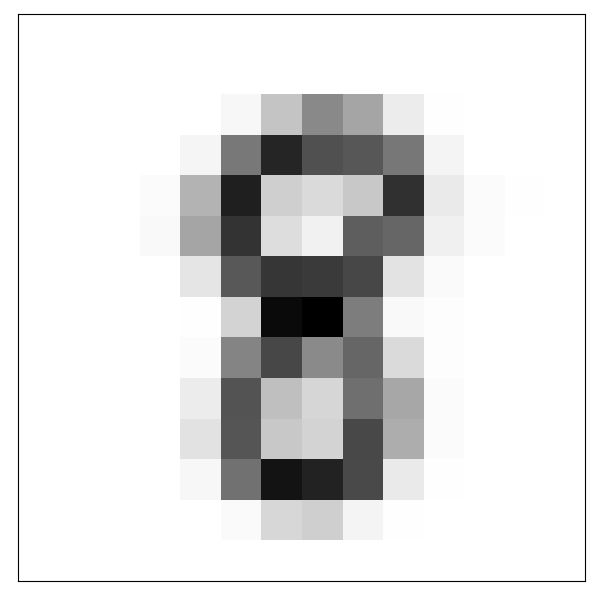}
\includegraphics[trim=0 0 0 0,clip,width=0.25\textwidth]{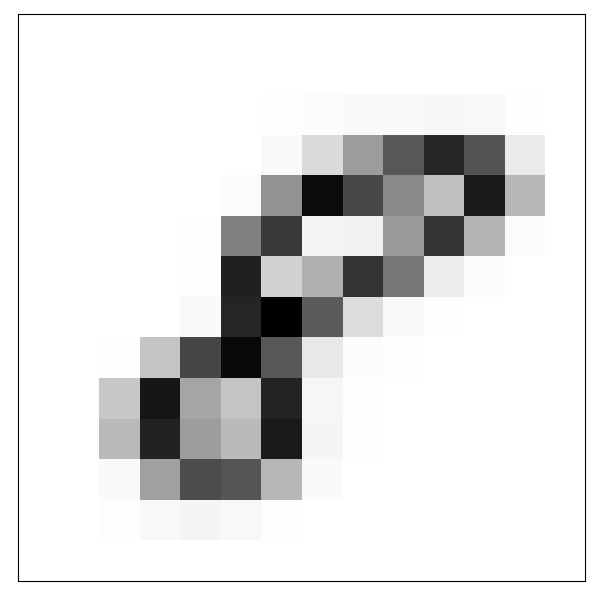}
\includegraphics[trim=0 0 0 0,clip,width=0.25\textwidth]{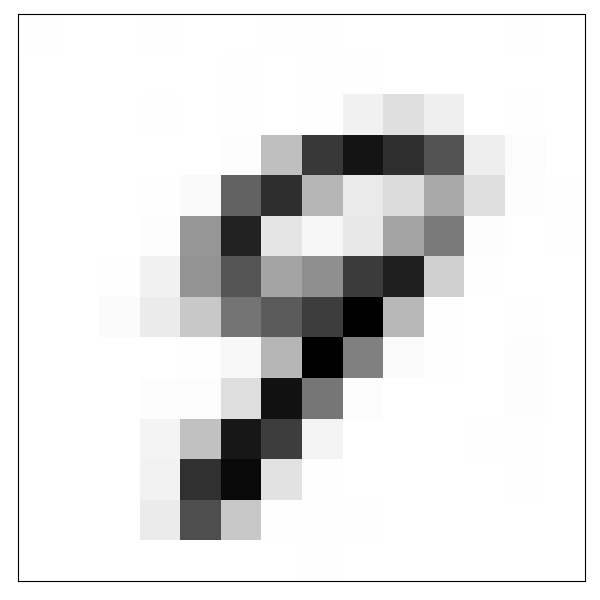}
\includegraphics[trim=0 0 0 0,clip,width=0.25\textwidth]{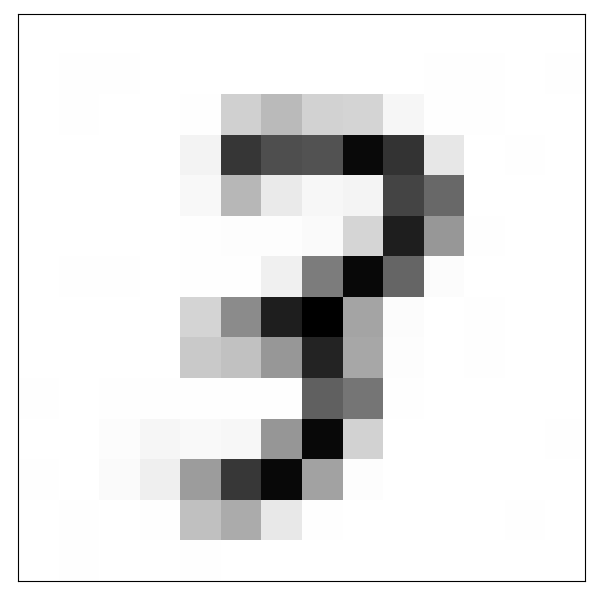}
\includegraphics[trim=0 0 0 0,clip,width=0.25\textwidth]{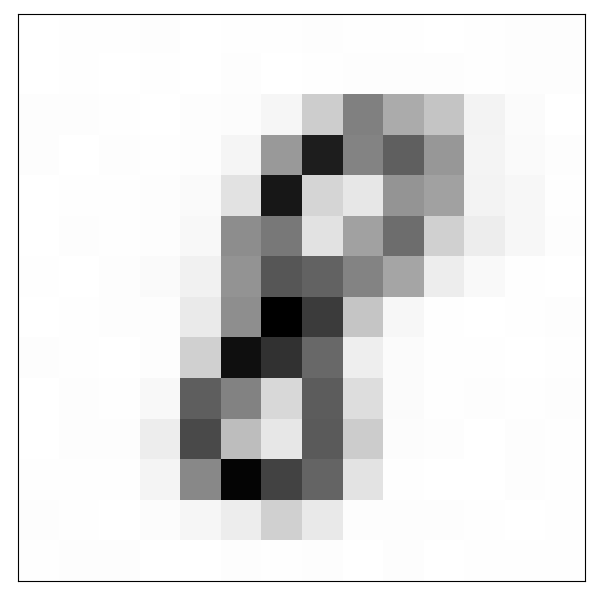}
\caption{$l=2$ (parent node: 8).}
\end{subfigure}
\begin{subfigure}[b]{0.15\textwidth}
\centering
\includegraphics[trim=0 0 0 0,clip,width=0.25\textwidth]{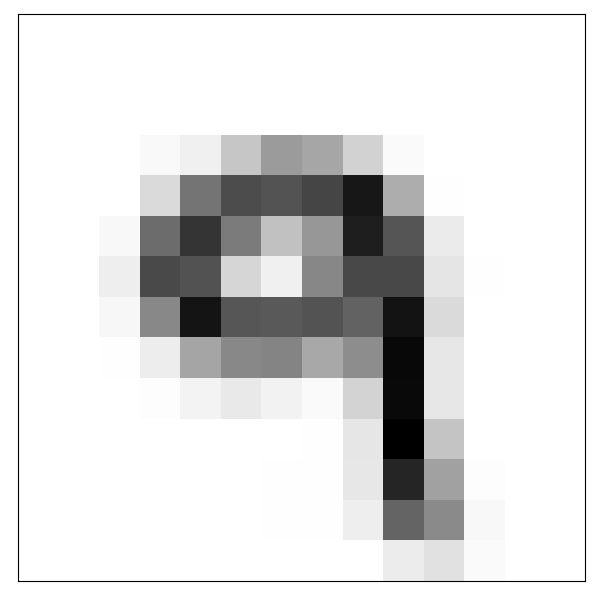}
\includegraphics[trim=0 0 0 0,clip,width=0.25\textwidth]{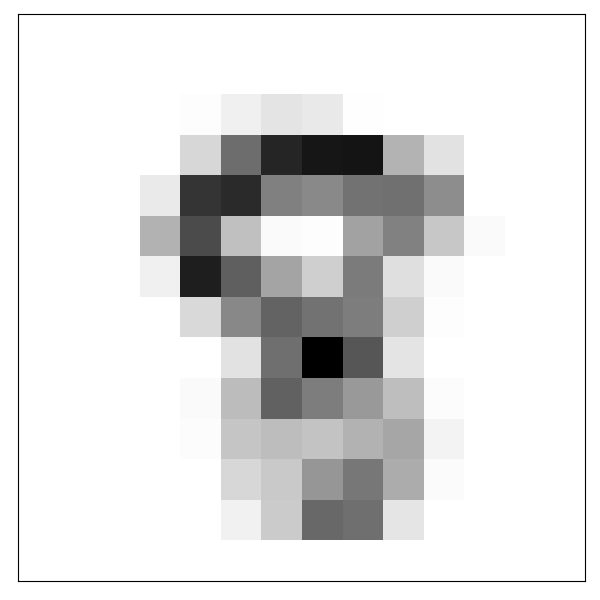}
\includegraphics[trim=0 0 0 0,clip,width=0.25\textwidth]{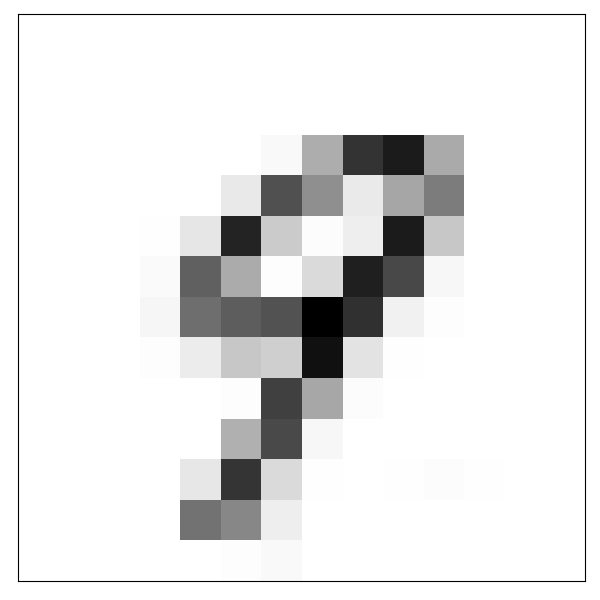}
\includegraphics[trim=0 0 0 0,clip,width=0.25\textwidth]{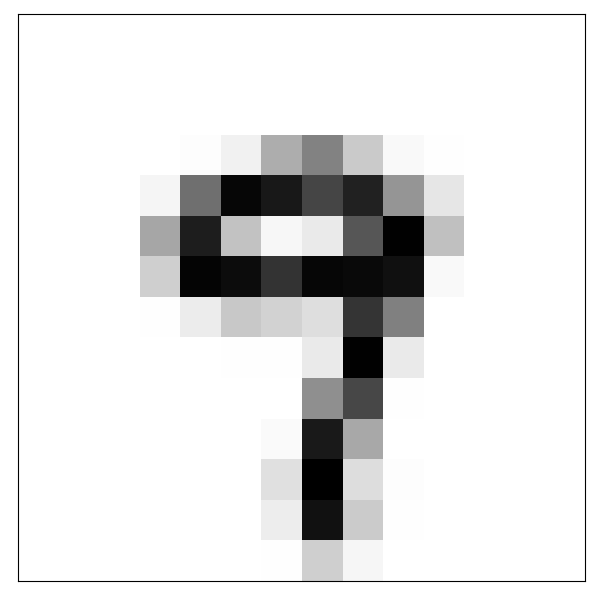}
\includegraphics[trim=0 0 0 0,clip,width=0.25\textwidth]{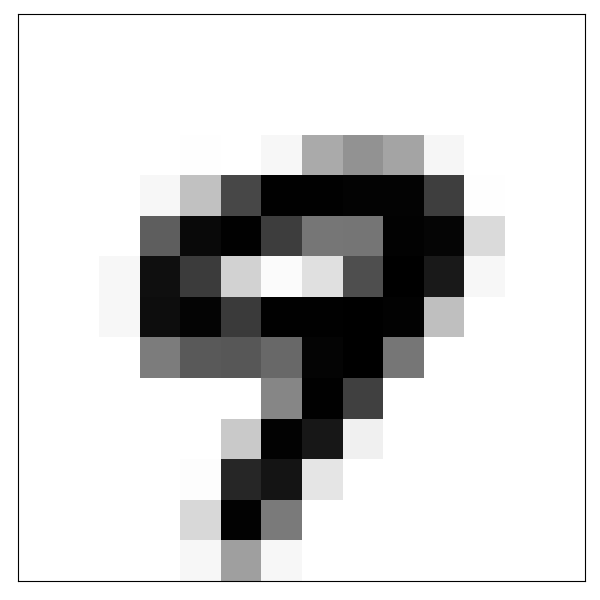}
\includegraphics[trim=0 0 0 0,clip,width=0.25\textwidth]{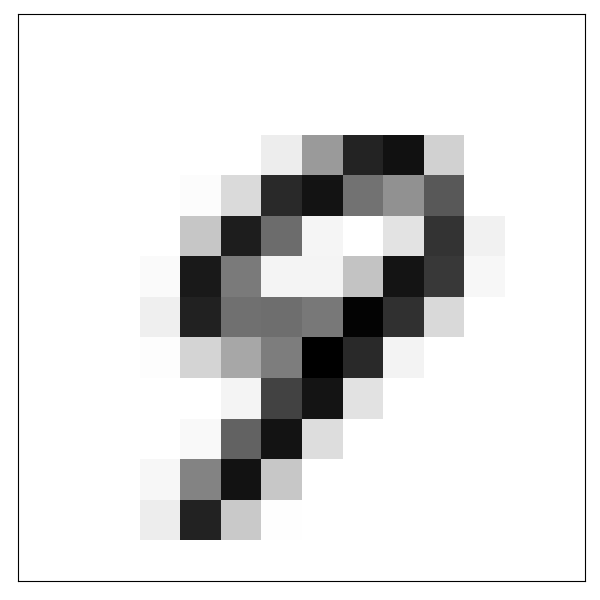}
\includegraphics[trim=0 0 0 0,clip,width=0.25\textwidth]{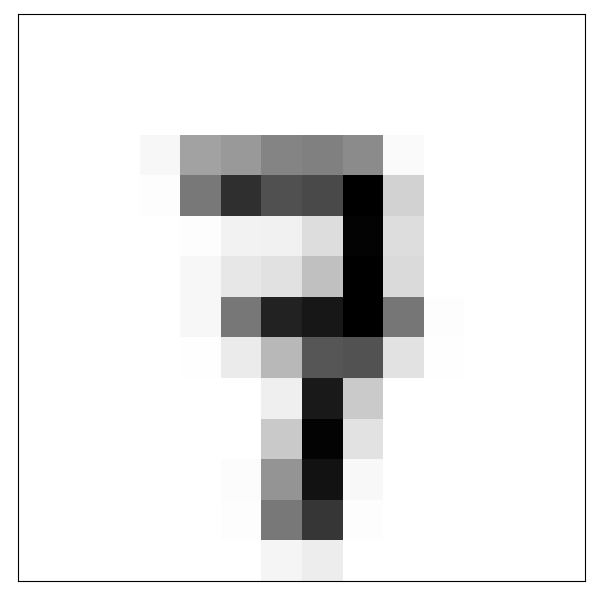}
\includegraphics[trim=0 0 0 0,clip,width=0.25\textwidth]{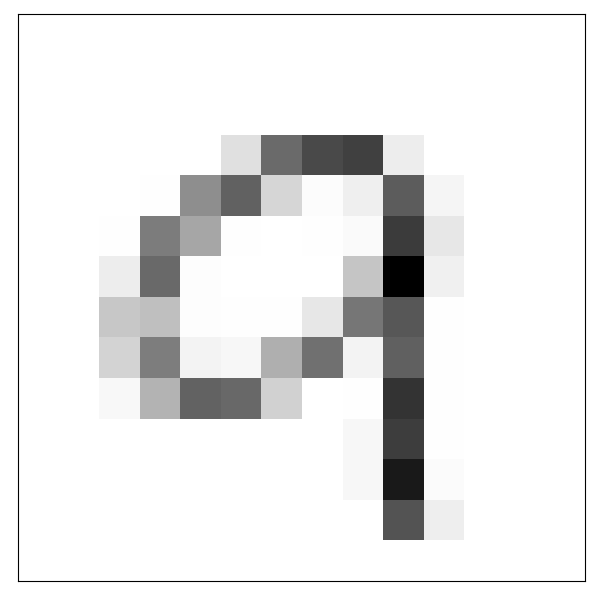}
\includegraphics[trim=0 0 0 0,clip,width=0.25\textwidth]{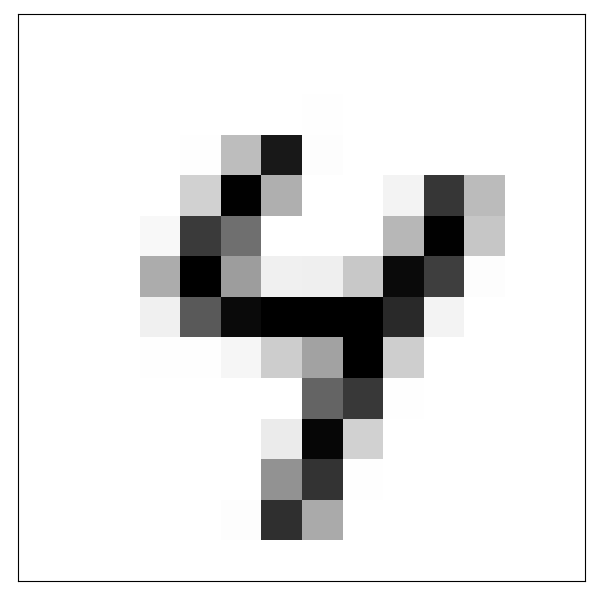}
\includegraphics[trim=0 0 0 0,clip,width=0.25\textwidth]{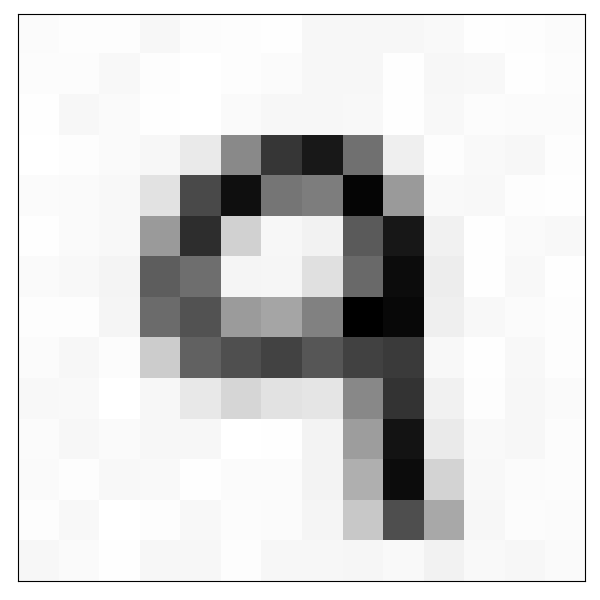}
\caption{$l=2$ (parent node: 9).}
\end{subfigure}
\caption{Learned representations of the first two layers a multi-resolution Online Deterministic Annealing algorithm trained on the MNIST dataset.}
\label{fig:mnist}
\end{figure*}

Finally, we evaluate the performance of the algorithm in image recognition, which 
is considered to be central to human perception and cognition.
To illustrate the interpretability of the representations learned by the model, 
we show in Fig. \ref{fig:mnist} the weights of the neurons (codevectors) of the first two layers of 
a multi-resolution ODA architecture trained on the MNIST dataset \cite{lecun1998gradient} of hand-written digits.
All $10$ neurons of the first layer are shown. For the second layer, 10 of the neurons-children of each first-layer neuron 
are randomly selected and shown.
For the first two layers, a wavelet representation of images of $7\times 7$ and $14\times 14$ pixels was used.
The testing performance for the first layer of $10$ neurons is $82\%$, and it goes up to $89\%$ using both layers
which add up to $212$ neurons.
This is only using the low-resolution wavelet representation of $14\times 14$ pixels.
Notice how different deformations of the hand-written digits are recognized by the algorithm 
without explicitly having to specify these transformations a priori. 
By studying the relationship between different codevectors that belong to the same class, 
it is possible to reconstruct class-invariant transformations and conserved quantities that 
will enable better feature extraction.
We continue the learning process using two more layers based on 
the scattering transform provided by the `Kymatio' software \cite{andreux2019kymatio} 
using a spatial resolution $J=2$ and `morlet' complex directional wavelets of 
four angles for the wavelet transform.
We stop the process when the percentage increment in accuracy is below a certain threshold
which signifies that further increase in the model complexity will likely not have a corresponding effect 
in the performance, and may cause over-fitting.
The final architecture consists of $1,960$ neurons and achieves accuracy $97.21\%$, in a
testing dataset of $10,000$ samples. 
As discussed above, by following this process, a knowledge representation has been built 
such that, if further performance boost 
is required, we can locate the region of the data space that should be given more emphasis.
This property, in conjunction with the fact that the neurons live in the data space itself (Fig. \ref{fig:mnist}),
make the proposed learning architecture inherently interpretable.

%


\subsection{Source Code and Reproducibility}

Code and Reproducibility: The source code is publicly available 
at \url{https://github.com/MavridisChristos/OnlineDeterministicAnnealing}.

\section{Conclusion}
\label{Sec:Conclusion}

We used principles form mathematics, system theory, and optimization to investigate the 
structure of a data-agnostic learning architecture that resembles the ``one learning algorithm'' 
believed to exist in the visual and auditory cortices of the human brain.
Our approach consists of a closed-loop system with three main components:
(i) a multi-resolution analysis pre-processor, 
(ii) a group-invariant feature extractor, and 
(iii) a progressive knowledge-based learning module, 
along with
multi-resolution feedback loops that are used for learning.
The design of (i) and (ii), 
is based on the well-established theory of wavelet-based multi-resolution 
analysis and group convolution operators. 
%
Regarding (iii), we introduced a multi-resolution extension of the Online Deterministic Annealing algorithm, 
a novel learning algorithm 
that constructs progressively growing knowledge representations.
The resulting architecture shares similar properties with 
Deep Convolutional Networks and the
Scattering Convolutional Networks
but also provides a more general system-theoretic framework with 
some significant properties.
First, it is dataset-agnostic, a term that in this case has a double meaning: 
it is not tailored to a specific learning task, e.g., face recognition only, 
and does not require the a priori knowledge of an entire dataset.
In contrast, online (or real-time) observations are used to train the algorithm.
Second, it is hierarchical as a result of the multi-resolution pre-processing with 
the wavelet transform and the tree-structure of the learning algorithm.
Third, it is memory-based, since a knowledge base, i.e., the set of codevectors
representing the data space, is being built through training.
In addition, it is interpretable, a term that again has a double meaning. In the case of a 
dataset that can be visualized by humans, the fact that the codevectors are representatives
of the data space is enough to convey the information of what has been learned. 
Even in the general case when this is not possible, however, the proposed approach offers a 
structured knowledge base that can be used to localize the error-prone regions of the feature
space, which leads to understanding when and why the learned model fails. 
This information is valuable to the designers of a learning model and can be used to 
incrementally improve its performance without re-training the entire model from scratch.
One more property of the proposed approach is robustness with respect to initial conditions, 
perturbations, and adversarial attacks.
This is an aftermath of the annealing optimization used to train the learning module, 
and the established properties of the vector quantization variants.
Finally, it is progressive in the sense that the knowledge base is 
progressively growing, starting from very few codevectors, and adding more as needed.
This interplay between compression and classification does not only help with the complexity
of the model, but also with the generalization properties and with the sensitivity of the model
with respect to input noise. It also provides one of the first approaches to adaptively 
grow a neural network architecture, 
in contrast to ad-hoc experimentation on the number of neurons and number of layers to be used.


\bibliographystyle{IEEEtran} %
\bibliography{bib_learning.bib,bib_mavridis}

\begin{IEEEbiography}[{\includegraphics[width=1in,height=1.25in,clip,keepaspectratio]
{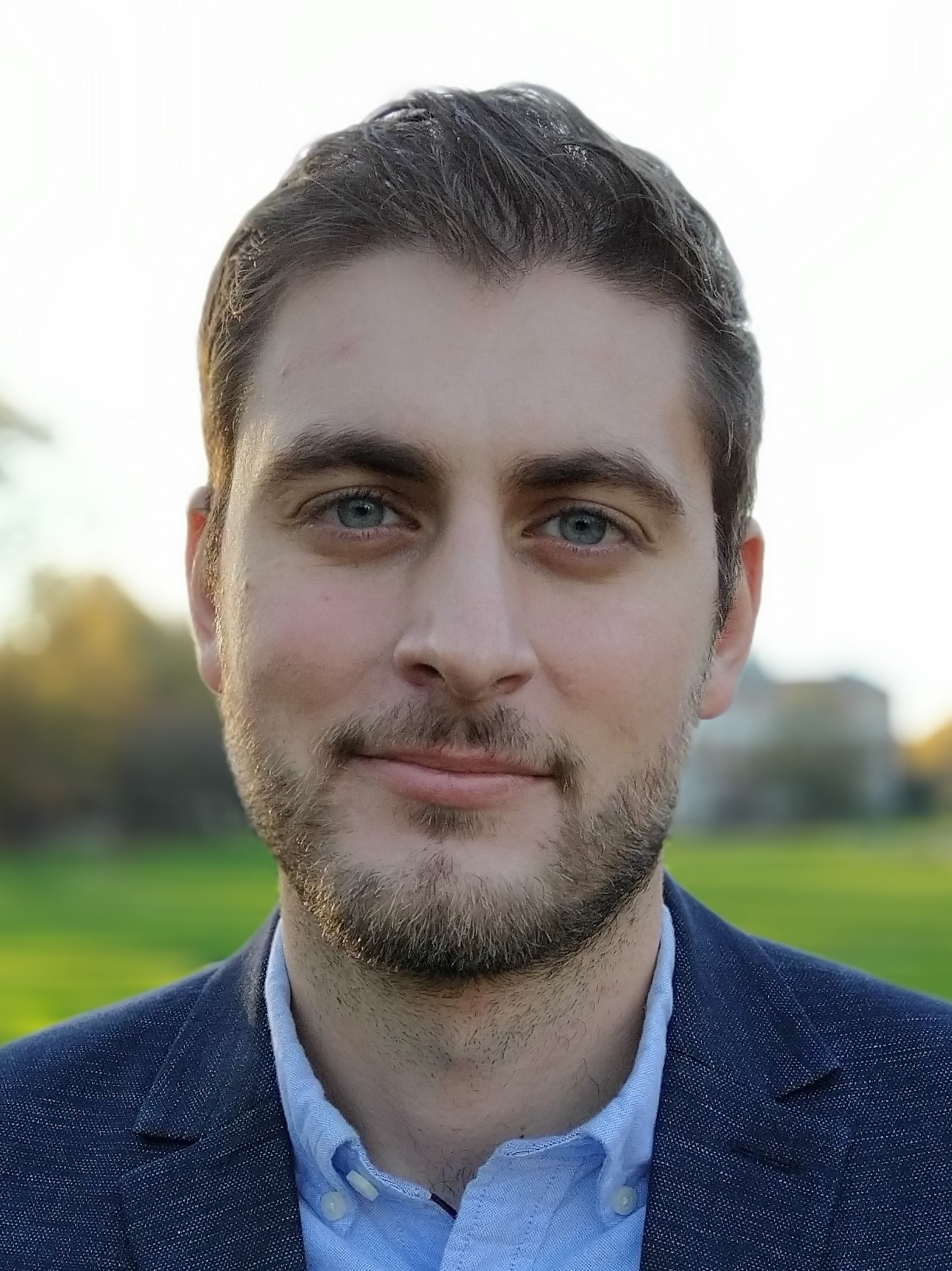}}]{Christos N. Mavridis} (M'20) 
received the Diploma degree in electrical and computer engineering from the National Technical University of Athens, Greece, in 2017,
and the M.S. and  Ph.D. degrees in electrical and computer engineering at the University of Maryland, College Park, MD, USA, in 2021. 
His research interests include learning theory, stochastic optimization, systems and control theory, multi-agent systems, and robotics. 

He has worked as a researcher at the Department of Electrical and Computer Engineering at the University of Maryland, College Park, MD, USA, and as a research intern for the Math and Algorithms Research Group at Nokia Bell Labs, NJ, USA, and the System Sciences Lab at Xerox Palo Alto Research Center (PARC), CA, USA. 

Dr. Mavridis is an IEEE member, and a member of the Institute for Systems Research (ISR) and the Autonomy, Robotics and Cognition (ARC) Lab. He received the Ann G. Wylie Dissertation Fellowship in 2021, and the A. James Clark School of Engineering Distinguished Graduate Fellowship, Outstanding Graduate Research Assistant Award, and Future Faculty Fellowship, in 2017, 2020, and 2021, respectively. He has been a finalist in the Qualcomm Innovation Fellowship US, San Diego, CA, 2018, and he has received the Best Student Paper Award (1st place) in the IEEE International Conference on Intelligent Transportation Systems (ITSC), 2021.
\end{IEEEbiography}

\begin{IEEEbiography}[{\includegraphics[width=1in,height=1.25in,clip,keepaspectratio]
{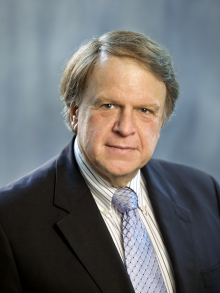}}]{John S. Baras} (LF'13) 
received the Diploma degree in electrical and mechanical engineering from the National Technical University of Athens, Athens, Greece, in 1970, and the M.S. and Ph.D. degrees in applied mathematics from Harvard University, Cambridge, MA, USA, in 1971 and 1973, respectively.

He is a Distinguished University Professor and holds the Lockheed Martin Chair in Systems Engineering, with the Department of Electrical and Computer Engineering and the Institute for Systems Research (ISR), at the University of Maryland College Park. From 1985 to 1991, he was the Founding Director of the ISR. Since 1992, he has been the Director of the Maryland Center for Hybrid Networks (HYNET), which he co-founded. His research interests include systems and control, optimization, communication networks, applied mathematics, machine learning, artificial intelligence, signal processing, robotics, computing systems, security, trust, systems biology, healthcare systems, model-based systems engineering.

Dr. Baras is a Fellow of IEEE (Life), SIAM, AAAS, NAI, IFAC, AMS, AIAA, Member of the National Academy of Inventors and a Foreign Member of the Royal Swedish Academy of Engineering Sciences. Major honors include the 1980 George Axelby Award from the IEEE Control Systems Society, the 2006 Leonard Abraham Prize from the IEEE Communications Society, the 2017 IEEE Simon Ramo Medal, the 2017 AACC Richard E. Bellman Control Heritage Award, the 2018 AIAA Aerospace Communications Award. In 2016 he was inducted in the A. J. Clark School of Engineering Innovation Hall of Fame. In 2018 he was awarded a Doctorate Honoris Causa by his alma mater the National Technical University of Athens, Greece.   
\end{IEEEbiography}

\end{document}